\documentclass[Afour,sageh,times]{sagej}
\usepackage{amsmath,amsfonts}
\usepackage{mathtools} 
\usepackage{algorithmic}
\usepackage{algorithm}
\usepackage{array}
\usepackage{ulem}
\usepackage{empheq}
\usepackage{textcomp}
\usepackage{stfloats}
\usepackage{url}
\usepackage{verbatim}
\usepackage{graphicx}
\usepackage{xcolor}
\usepackage{enumerate}
\usepackage{multirow}

\usepackage{moreverb,url}

\usepackage[colorlinks,bookmarksopen,bookmarksnumbered,citecolor=red,urlcolor=red]{hyperref}
\usepackage{bm}

\newcounter{eqtag}
\newcounter{eqtagb}
\newcounter{eqtagd}
\newcommand{\mytag}{\refstepcounter{eqtag}\tag{B\theeqtag}}
\newcommand{\mytagb}{\refstepcounter{eqtagb}\tag{C\theeqtagb}}
\newcommand{\mytagd}{\refstepcounter{eqtagd}\tag{D\theeqtagd}}

\newcommand\BibTeX{{\rmfamily B\kern-.05em \textsc{i\kern-.025em b}\kern-.08em
T\kern-.1667em\lower.7ex\hbox{E}\kern-.125emX}}

\begin{document}

\runninghead{Mathew et al.}

\title{Analytical Derivatives for Efficient Mechanical Simulations of Hybrid Soft Rigid Robots}

\author{Anup Teejo Mathew\affilnum{1,2},
		Frederic Boyer\affilnum{3},
            Vincent Lebastard\affilnum{3},
		Federico Renda\affilnum{1,2}}

\affiliation{\affilnum{1}Department of Mechanical and Nuclear Engineering, Khalifa University of Science and Technology, Abu Dhabi, UAE\\
\affilnum{2}LS2N Laboratory, Institut Mines Telecom Atlantique, Nantes 44307, France\\
\affilnum{3}Khalifa University Center for Autonomous Robotic Systems (KUCARS), Khalifa University, Abu Dhabi, UAE}

\affiliation{\affilnum{1}Department of Mechanical Engineering, Khalifa University, Abu Dhabi, UAE\\
\affilnum{2}Khalifa University Center for Autonomous Robotic Systems (KUCARS), Khalifa University, Abu Dhabi, UAE\\
\affilnum{3}LS2N Laboratory, Institut Mines Telecom Atlantique, Nantes 44307, France}

\corrauth{Anup Teejo Mathew, Department of Mechanical Engineering, Khalifa University, Abu Dhabi, UAE.}

\email{anup.mathew@ku.ac.ae, anupteejo@gmail.com}

\begin{abstract}
Algorithms that use derivatives of governing equations have accelerated rigid robot simulations and improved their accuracy, enabling the modeling of complex, real-world capabilities. However, extending these methods to soft and hybrid soft-rigid robots is significantly more challenging due to the complexities in modeling continuous deformations inherent in soft bodies. A considerable number of soft robots and the deformable links of hybrid robots can be effectively modeled as slender rods. The Geometric Variable Strain (GVS) model, which employs the screw theory and the strain parameterization of the Cosserat rod, extends the rod theory to model hybrid soft-rigid robots within the same mathematical framework. Using the Recursive Newton-Euler Algorithm, we developed the analytical derivatives of the governing equations of the GVS model. These derivatives facilitate the implicit integration of dynamics and provide the analytical Jacobian of the statics residue, ensuring fast and accurate computations. We applied these derivatives to the mechanical simulations of six common robotic systems: a soft cable-driven manipulator, a hybrid serial robot, a fin-ray finger, a hybrid parallel robot, a contact scenario, and an underwater hybrid mobile robot. Simulation results demonstrate substantial improvements in computational efficiency, with speed-ups of up to three orders of magnitude. We validate the model by comparing simulations done with and without analytical derivatives. Beyond static and dynamic simulations, the techniques discussed in this paper hold the potential to revolutionize the analysis, control, and optimization of hybrid robotic systems for real-world applications.
\end{abstract}


\maketitle

\section{Introduction}

Rigid-body algorithms have been developed for the mechanical analysis of multi-body systems, enabling the modeling of dynamic response and static equilibrium of robots \cite{Featherstone, Murray}. Over the years, these algorithms have undergone remarkable improvements, enabling faster-than-real-time simulations \cite{Newbury2024}. One of the most significant advancements has been the incorporation of gradient information into these algorithms \cite{Carpentier2019, howell2023dojodifferentiablephysicsengine, Markus2017, Geilinger2020}. The ability to accurately and efficiently compute the derivatives of governing equations with respect to the system's state, model parameters, and control variables has been pivotal for implicit integration, design optimization, trajectory optimization, and optimal control.
The impact of these advancements is exemplified by leading humanoid and quadrupedal robots, such as those developed by Boston Dynamics and Unitree. These robots leverage the gradient information for trajectory optimization and model-predictive control (MPC), enabling real-time control, enhanced stability, and adaptability in complex environments \cite{Sukhija2023, Neunert2018, Wensing2024}.

Several methods exist for computing the derivatives of governing equations, each with its trade-offs \cite{Newbury2024}. The most straightforward approach is the numerical scheme of finite difference. The finite difference method can offer simplicity and ease of parallelization but often falls short in accuracy \cite{Todorov2012MuJoCo}. Automatic differentiation (AutoDiff) computes derivatives by recognizing that even complex functions are built from fundamental operations and functions. It systematically applies the chain rule to these operations within the algorithm, allowing the program to automatically calculate derivatives alongside the original calculations \cite{drake, Giftthaler2017}. However, AutoDiff involves intermediate computations that are generally hard to simplify. Analytical methods take a manual approach by directly applying the chain rule to recursive algorithms like the Recursive Newton-Euler Algorithm (RNEA) \cite{carpentier2018analytical, Singh2022}. By exploiting the inherent structure and spatial algebra at the core of rigid-body dynamics algorithms, analytical derivatives (AD) can simplify computations and achieve greater computational efficiency than automatic differentiation methods. Efficient implementation of analytical derivatives can lead to faster and more resource-efficient computations with improved accuracy and provide deeper insight into the mathematical structure of the derivatives. However, deriving analytical derivatives manually can be complex and time-consuming, making them challenging to implement \cite{Singh2022}.

Deriving analytical derivatives in soft robots is significantly more challenging than in rigid body systems. The primary difficulty stems from the complex nature of deformable bodies, which undergo large, continuous deformations, making it highly challenging to derive closed-form equations for their dynamics. The general class of soft robots, which cannot be modeled as a system rods, require numerical methods based on 3D continuum mechanics such as Finite Element Methods (FEM) \cite{Duriez_ICRA2013} or Material Point Method (MPM) \cite{Hu2018}. Analytical derivatives of FEM in robotics have been explored in several works, including \cite{Hoshyari2019, Hafner2019, Geilinger2020, Bacher2021, Jatavallabhula2021, Du2022}. On the other hand, MPM is often referred to as naturally differentiable due to its particle-grid representation and the smooth interpolation between particles and the grid, which makes gradient computation more efficient \cite{Spielberg2023, Huang2021PlasticineLabAS}. FEM and MPM use maximal coordinate representations for rigid bodies, increasing the degrees of freedom (DoF) and the computational cost for hybrid soft-rigid robots.

A large portion of soft robots and the compliant links of hybrid soft rigid robots can be effectively modeled as slender, rod-like structures, making them well-suited for analysis using the Cosserat rod theory \cite{Armanini2023}.  The Cosserat rod is a 1D continuum mechanics object that can model deformations of slender bodies, including twisting, bending in two axes, stretching, and shearing in two axes \cite{Cao2008}. Four distinct research communities have leveraged Cosserat rod theory to address their specific challenges, each producing specialized numerical methods tailored to their needs. Ranking them by their order of appearance, these communities are: the geometrically exact FEM community, the ocean engineering community, the computer graphics community, and the robotics community. The geometrically exact FEM community has focused primarily on developing FEM software that can predict the movements and stresses of mechanisms undergoing large deformations \cite{Simo_CMAME1988, Meier2019, Eugster2023}. Meanwhile, the ocean engineering community has applied Cosserat rod models to the simulation of towed submarine cables, addressing the unique challenges posed by underwater environments \cite{Burgess1992, Tjavaras1998}. On the other hand, the computer graphics community has prioritized computational speed for interactive simulations of filament-like structures, such as hair, using the Discrete Elastic Rod (DER) approaches \cite{Bergou2008, Gazzola_RSOS2018}. Finally, the robotics community has concentrated on the simulation and control of soft or continuum robots to safely interact with their surroundings \cite{Rucker2011, Till_IJRR2019, Boyer2022}.

To cater specifically to the needs of robotics, a novel parameterization of the configuration space of Cosserat rods has been proposed, focusing on strain fields rather than the traditional pose-based approach used in FEM and DER. This strain-based approach, referred to as the Geometric Variable Strain (GVS) method \cite{Renda_RAL2020, Boyer_TRO2020}, is geometrically exact and aligns well with the demands of soft robotic applications, providing a more efficient framework for modeling their dynamic and compliant behavior. Among the various models based on the Cosserat rod, the GVS model stands out for its ability to merge the screw theory-based formulation of rigid robots with the strain parameterization of the rod. Its Lagrangian mechanics formulation with minimal generalized coordinates enables efficient analysis of hybrid soft-rigid robotic systems within a unified mathematical framework. The implementation of the GVS model, based on the approximate Magnus expansion of the rod's strain field, makes the soft body computationally equivalent to multidimensional, discrete rigid joints \cite{Mathew2024}. The model has been extensively compared and validated with analytical models, FEM, and other rod models in previous studies \cite{BoyerTRO2023, Mathew2022}. In \cite{Mathew2022, Mathew2024}, we implemented the GVS model for hybrid soft robots using a built-in implicit integration scheme in MATLAB called $ode15s$. When analytical derivatives (Jacobian) are not supplied, $ode15s$ internally compute the Jacobian of the governing equations using a finite difference scheme. However, this can lead to longer computational times or cause the solver to stall due to errors introduced by the numerical approximation, particularly in high-DoF systems with constraints. The objective of this work is to develop and implement an analytical derivative for the GVS model, aiming to improve computational efficiency and robustness in the simulation of hybrid soft rigid robots.

\textit{Related works}: The Piecewise Constant Strain (PCS) model, which discretizes the continuous deformation of a Cosserat rod into a finite number of segments with constant strain, is a subclass of the GVS model. Based on the Comprehensive Motion Transformation Matrix (a Lie group of coordinate transformations of displacement, velocity, and acceleration), an analytical gradient of the PCS model was derived in \cite{Ishigaki2024}. A differentiable soft robot simulation environment called DisMech was introduced based on the implicit DER method \cite{Choi2024}. DisMech employs a finite difference scheme to compute the necessary gradients for the implicit integration. Recently, a new algorithm for the implicit dynamics of robots with rigid bodies and Cosserat rods has been proposed \cite{BoyerTRO2023}. By applying an exact symbolic differentiation of the robot's RNEA inverse dynamics (named $IDM$ in \cite{BoyerTRO2023}), a new RNEAlgorithm, called the tangent inverse dynamics model ($TIDM$), has been derived. This algorithm is then fed with unit inputs to numerically calculate the Jacobian of the inverse dynamics. This calculation is performed at the continuous level, i.e. directly on the partial differential equations (PDEs), and before spatial integration, for which spectral methods have been employed instead of Magnus expansion. Although the approach adheres to the true definition of geometrically exact methods - where discretization of time and space occurs only after all analytical calculations - due to its implicit nature, it does not directly provide Jacobian matrices in analytical matrix form, which can be advantageous for control and fast simulation of robots with MATLAB. In contrast, the approach here presented lies in its analytic and explicit formulation, which was made possible by the Magnus expansion. 

\textit{Contributions of this work}: By leveraging the ``pseudo-rigid joint" formulation of GVS and building upon established methods for analytical derivatives in rigid body systems \cite{carpentier2018analytical, Singh2022}, we developed analytical derivatives for soft and hybrid soft-rigid robots with slender soft bodies. We implemented the derivatives in two implicit integration algorithms: $ode15s$ of MATLAB and Newmark-$\beta$ scheme for dynamics and provided the analytical Jacobian for efficient static equilibrium computation. Our method significantly improved the computational speed of implicit integration for dynamic simulations, achieving speed-ups of up to three orders of magnitude compared to traditional methods without analytical derivatives. Similarly, for static analysis, we observed substantial improvements in computational efficiency. Six common robotic systems are considered for the simulation study. To the best of the authors' knowledge, this is the first work to derive and implement analytical derivatives for the mechanical analysis of hybrid soft-rigid robotic systems of this kind.

\textit{Organization of the paper}: A summary of the GVS model is presented in Section \ref{sec::GVS summary}. Section \ref{sec::Why and How Derivatives} details the implementation of analytical derivatives in dynamic and static algorithms for fast and accurate computations. In Section \ref{sec::Analytical Derivatives of Soft Body}, we focus initially on a single soft body, applying the RNEA and $ID$ framework to derive the analytical derivatives of the governing equations. This framework is extended to hybrid multi-body systems in Section \ref{sec::Analytical Derivatives of Hybrid multi-body}. We address two typical constraints in multi-body systems: joint actuation via joint coordinates and closed-chain (CC) systems. Section \ref{sec::Common External Forces} discusses the derivation of analytical derivatives for systems subjected to three common external forces: point forces, contact loads, and hydrodynamic forces. Simulations and validations across six robotic systems are presented across these sections, demonstrating the effectiveness of the approach. 

Table \ref{tab::symbols} lists all the important symbols used in this paper. Readers are encouraged to refer to the supplementary videos to visualize the dynamic simulation results presented in this work.

\begin{table}[h!]
\centering
\caption{List of symbols and their descriptions}
\label{tab::symbols}
\begin{tabular}{ | m{1.6cm} | m{6cm} | } 
  \hline
  \textbf{Symbol} & \textbf{Description} \\ 
  \hline
  $n_{dof}$ & Total degrees of freedom \\
  $n_{a}$ & Number of actuators \\
  $N$ & Number of Cosserat rods \\
  $n_{p}$ & Number of computational points \\
  $\bm{q}$ & Generalized coordinates \\ 
  $\dot{\bm{q}}$ & Generalized velocities \\ 
  $\ddot{\bm{q}}$ & Generalized accelerations \\ 
  $\bm{x}$ & Generalized robot state. $\bm{x}=[\bm{q}^T \; \dot{\bm{q}}^T]^T$\\
  $h$ & Time step \\
  $t$ & Time \\
  $X$ & Curvilinear abscissa of the soft body, $X \in [0,\;L]$ \\
  $FD$ & Forward dynamics \\ 
  $ID$ & Inverse dynamics \\ 
  $\bm{M}$ & Generalized mass matrix \\
  $\bm{F}$ & Generalized external and Coriolis force \\
  $\bm{\tau}$ & Generalized internal force \\
  $\bm{B}$ & Generalized actuation matrix \\
  $\bm{u}$ & Actuator strength \\
  $\bm{K}$ & Generalized stiffness matrix \\
  $\bm{D}$ & Generalized damping matrix \\
  $\bm{\xi}$ & Screw strain \\
  $\bm{\Phi}_\xi$ & Strain basis \\
  $(\bullet)_\alpha$ & Quantity at joint or computational point $\alpha$ \\
  $(\bullet)^B$ & Quantities computed during the forward pass \\
  $(\bullet)^{C,S}$ & Quantities computed during the backward pass \\
  $\bm{S}_\alpha$ & Joint motion subspace matrix \\
  $\bm{\mathcal{F}}_k$ & Local wrench\\
  $\bm{\mathcal{M}}_k$ & Screw inertia matrix\\
  $\bm{\eta}_k$ & Velocity twist\\
  $\dot{\bm{\eta}}_k$ & Acceleration twist\\
  $\bm{\mathcal{G}}$ & $\bm{\mathcal{G}} = [\bm{0}^T \bm{a}_g^T]^T$, where $\bm{a}_g$ is the acceleration due to gravity\\
  $\bm{g}$ & Homogeneous transformation matrix in $SE(3)$\\
  $\mathrm{Ad}_{(\cdot)}$, $\mathrm{Ad}_{(\cdot)}^*$ & Adjoint maps in $SE(3)$\\
  $\mathrm{ad}_{(\cdot)}$, $\mathrm{ad}_{(\cdot)}^*$, $\overline{\mathrm{ad}}_{(\cdot)}^*$ & adjoint operators in $\mathfrak{se}(3)$\\
  \hline
\end{tabular}
\end{table} 

\section{Summary of the GVS Model}
\label{sec::GVS summary}

\begin{figure*}
\centering
\includegraphics[width=\textwidth]{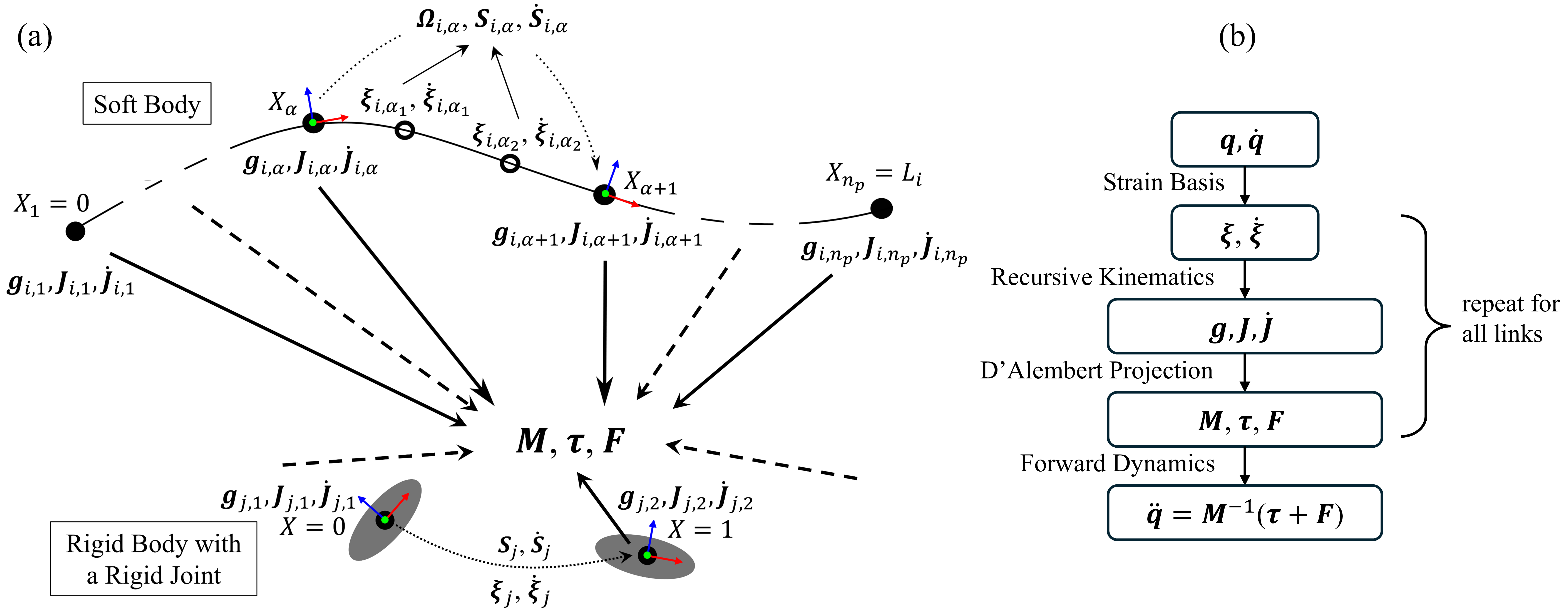}
\caption{Graphical summary of the GVS model: (a) Schematic of the implemented recursive scheme. (b) Block diagram showing the summary of the GVS $FD$ algorithm.}
\label{fig::GVS_Summary}
\end{figure*}

The GVS model introduces generalized coordinates ($\bm{q}$) using a variable strain parameterization of the Cosserat rod:

\begin{equation}\label{eq::strainparameterization}
\bm{\xi}_i(X, \bm{q}_i) = \bm{\Phi}_{\xi_i}(X)\bm{q}_i + \bm{\xi}_i^*(X)
\end{equation}
where $X \in [0, L_i]$ is the curvilinear abscissa of the rod, $\bm{\xi} \in \mathbb{R}^{6}$ is the screw strain, $\bm{\Phi}_{\xi_i} \in \mathbb{R}^{6\times {n_{dof}}_i}$ (${n_{dof}}_i$ is the degrees of freedom) is a strain basis, and $\bm{\xi}^*$ is the reference strain. The subscript $i$ indicates the $i^{th}$ Cosserat rod. In the GVS formulation, a rigid joint is equivalent to a fictitious Cosserat rod spanning from $X = 0$ to $X = 1$ \cite{Mathew2024}. For a rigid joint, the strain twist is equivalent to the joint twist in $\mathbb{R}^{6}$ and is independent of $X$.

For a hybrid robot with $N$ Cosserat rods (including rigid joints and soft bodies), the state of the robot $\bm{x}$ is governed by the generalized coordinates $\bm{q} \in \mathbb{R}^{n_{dof}}$ ($n_{dof}=\sum_{i=1}^N {n_{dof}}_i$) and its time derivative $\dot{\bm{q}}$. Using $\bm{q}$, $\dot{\bm{q}}$ and basis functions, $\bm{\xi}$ and $\dot{\bm{\xi}}$ can be computed at each rigid joint and at any computational point on the soft bodies.


Using these, a recursive scheme that employs the exponential map of the Lie algebra of $ SE(3)$ is implemented to compute the robot's forward and differential kinematics (Appendix \ref{app::A}). This yields $\bm{g}_i \in SE(3)$, the homogeneous transformation matrix mapping from the inertial frame to the local frame, $\bm{J}_i \in \mathbb{R}^{6 \times n_{dof}}$, the geometric Jacobian in the local frame, and its time derivative $\dot{\bm{J}}_i$ at discrete computational point along the rod domains.
\begin{subequations}
\begin{align}
\bm{g}_{i, \alpha+1} &= \bm{g}_{i, \alpha} \exp\left({\widehat{\bm{\Omega}}_{i, \alpha}}\right) \label{eq::POE} \\
\bm{J}_{i, \alpha+1} &= \mathrm{Ad}_{\exp\left({\widehat{\bm{\Omega}}_{i, \alpha}}\right)}^{-1}\left(\bm{J}_{i, \alpha} + \big[ \bm{0}_{6 \times n_i^-} \; \bm{S}_{\alpha} \; \bm{0}_{6 \times n_i^+} \big] \right) \label{eq::Geometric Jacobian} \\
\dot{\bm{J}}_{i, \alpha+1} &= 
\begin{aligned}[t]
    &\mathrm{Ad}_{\exp\left({\widehat{\bm{\Omega}}_{i, \alpha}}\right)}^{-1}\Big(\dot{\bm{J}}_{i, \alpha} + \big[ \bm{0}_{6 \times n_i^-} \; \dot{\bm{S}}_{\alpha} \\
    &\quad + \mathrm{ad}_{\bm{\eta}_{i, \alpha}} \, \bm{S}_{\alpha} \; \bm{0}_{6 \times n_i^+} \big] \Big) 
\end{aligned} \label{eq::Geometric Jacobian dot}
\end{align}
\end{subequations}
where, $\bm{\Omega}_{i, \alpha}(\bm{q}_i)$ is an approximation of the Magnus expansion of $\bm{\xi}_i$, representing an equivalent joint twist from $\alpha$ to $\alpha+1$, $\bm{S}_\alpha(\bm{q}_i)  \in \mathbb{R}^{6 \times {n_{dof}}_i}$ is the joint motion subspace matrix, and $\dot{\bm{S}}_\alpha(\bm{q}_i,\dot{\bm{q}_i})$ is its time derivative. $\bm{\eta}_{i, \alpha} = \bm{J}_{i, \alpha}\dot{\bm{q}}$ is the screw velocity in the local frame. The matrix dimensions $ n_i^- = \sum_{k=0}^{i-1} {n_{dof}}_k $ and $ n_i^+ = \sum_{k=i+1}^{N} {n_{dof}}_k $. $\widehat{(\cdot)}$ is an isomorphism from $\mathbb{R}^{6}$ to $\mathfrak{se}(3)$ and $\mathrm{Ad}_{(\cdot)}$ represents the Adjoint map in $SE(3)$. Note that these equations also apply to rigid joints, providing a mapping from $X = 0$ to $X = 1$. Readers may refer to Appendices \ref{app::A} and \ref{app:C} for their analytical expressions.

Based on D'Alembert's principle, the free dynamics of soft links and rigid bodies are projected onto the generalized coordinate space using the geometric Jacobian (Kane method \cite{DAlembertKane}), yielding the generalized mass matrix $\bm{M}$, the generalized internal force $\bm{\tau}$, and the generalized external and Coriolis forces $\bm{F}$. For a soft body, the computation of these components requires the spatial integration of distributed quantities of soft links using a quadrature method, such as Gauss quadrature or the trapezoidal rule. For this numerical integration, the continuous domain of $X$ is discretized into $n_p$ computational points, and the integral of a distributed quantity $\overline{\bm{Q}}$ is approximated as a weighted sum of the integrands.
\begin{equation}
\label{eq::Gauss Quadrature}
\int_{0}^{L_i} \overline{\bm{Q}}_i \, dX = \sum_{k=1}^{n_p} w_k \, {\overline{\bm{Q}}_i}_k
\end{equation}.

Finally, the Forward Dynamics ($FD$) computes $\ddot{\bm{q}}$ by solving,

\begin{equation}
\begin{split}
\label{eq::FD}
\bm{M}(\bm{q})\ddot{\bm{q}} =\bm{\tau}(\bm{q},\dot{\bm{q}},\bm{u})+\bm{F}(\bm{q},\dot{\bm{q}})
\end{split}
\end{equation}
where, $ \bm{u} \in \mathbb{R}^{n_a} $ is the vector of applied actuation forces.


Figure \ref{fig::GVS_Summary} presents a graphical summary of the model implementation. The dynamic simulation computes the temporal evolution of the robot's state by time-integrating $[\dot{\bm{q}}^T, \; \ddot{\bm{q}}^T]^T$, while static simulation involves the computation of $\bm{q}$ for which the following equation is satisfied.
\begin{equation}
\begin{split}
\label{eq::statics}
\bm{\tau}(\bm{q},\bm{0},\bm{u})+\bm{F}(\bm{q},\bm{0})=0
\end{split}
\end{equation}

For further details on the model and its computational implementation, readers are encouraged to refer to \cite{Mathew2024}.

\section{Derivatives for Dynamics and Statics}
\label{sec::Why and How Derivatives}

Analytical derivatives of governing equations, \eqref{eq::FD} and \eqref{eq::statics}, are crucial for efficiently solving dynamic and static problems. Here, we explain how the derivatives can be utilized to ensure accurate and fast computations.


\subsection{Implicit Euler Integration Using Inverse Dynamics}

 The temporal evolution of the robot state $\bm{x}=[\bm{q}^T, \; \dot{\bm{q}}^T]^T$ can be computed numerically by integrating $\dot{\bm{x}}$ using an explicit or implicit scheme \cite{butcher2016numerical}.

The explicit Euler method is given by:
\begin{equation}
\nonumber
\bm{x}_{n+1} = \bm{x}_n + h \dot{\bm{x}}(t_n, \bm{x}_n)
\end{equation}
where $t$ is the time, $h$ is the step size, and the subscripts $n$ and $n+1$ indicate current and next time steps respectively.

For an explicit integrator, $\dot{\bm{x}}$ is a function of the current robot state. In contrast, the implicit Euler method is:
\begin{equation}
\nonumber
\bm{x}_{n+1} = \bm{x}_n + h \dot{\bm{x}}(t_{n+1}, \bm{x}_{n+1})
\end{equation}

Since $\bm{x}_{n+1}$ appears on both sides of the equation, it necessitates a solution through iteration. The equation can be solved for the unknown $\bm{x}_{n+1}$ by making it a residual:

\begin{equation}
\nonumber
\bm{R}_{dyn}(\bm{x}_{n+1}) = \bm{x}_{n+1} - \bm{x}_n - h \dot{\bm{x}}(t_{n+1}, \bm{x}_{n+1})
\end{equation}

The Jacobian of this equation is given by:

\begin{equation}
\nonumber
\bm{J}_{dyn}(\bm{x}_{n+1}) = \bm{I} - h \frac{\partial \dot{\bm{x}}}{\partial \bm{x}}\bigg|_{n+1}
\end{equation}
With an initial guess of $\bm{x}_{{n+1}_{guess}} = \bm{x}_{n}$, an iterative solver such as the Newton-Raphson method uses the residual and the Jacobian to march towards the solution. We used the $ode15s$ function in MATLAB, a variable-step, variable-order ODE solver that uses an advanced form of the implicit Euler method. Similar to the Euler scheme, $ode15s$ utilizes the state derivative ($\dot{\bm{x}}$) and its Jacobian for time integration. For a Lagrangian system governed by \eqref{eq::FD},
\begin{subequations}
\begin{align}
\dot{\bm{x}} =& \begin{bmatrix} \dot{\bm{q}} \\ FD\end{bmatrix} \label{eq::xd} \\
\frac{\partial \dot{\bm{x}}}{\partial \bm{x}}=& \begin{bmatrix}
\bm{0}_{n_{dof} \times n_{dof}} & \bm{I}_{n_{dof}} \\
\frac{\partial FD}{\partial \bm{q}} & \frac{\partial FD}{\partial \dot{\bm{q}}}
\end{bmatrix}\label{eq::dxd_dx}
\end{align}
\end{subequations}

While the Jacobian can be computed numerically or analytically, the analytical computation can provide faster and more accurate results. To obtain the derivatives of $FD$ for the computation of the Jacobian, we take a partial derivative of \eqref{eq::FD} with respect to $\bm{q}$:
\begin{equation}
\nonumber
\frac{\partial \bm{M}}{\partial \bm{q}}\ddot{\bm{q}}+\bm{M}\frac{\partial FD}{\partial \bm{q}} = \frac{\partial \bm{\tau}}{\partial \bm{q}}+\frac{\partial \bm{F}}{\partial \bm{q}}
\end{equation}


By rearranging terms, we get:

\begin{equation}
\label{eq::dFD_dq}
\frac{\partial FD}{\partial \bm{q}} = \bm{M}^{-1}\left(\frac{\partial \bm{\tau}}{\partial \bm{q}}-\frac{\partial ID}{\partial \bm{q}}\right)
\end{equation}
where $ID$ is the inverse dynamics given by,
\begin{equation}
ID(\bm{q},\;\dot{\bm{q}},\;\ddot{\bm{q}}) = \bm{M}(\bm{q})\ddot{\bm{q}}-\bm{F}(\bm{q},\;\dot{\bm{q}})\label{eq::ID}
\end{equation}
with $\ddot{\bm{q}}$ treated as an independent variable.

Similarly, we get:
\begin{equation}
\label{eq::dFD_dqd}
\frac{\partial FD}{\partial \dot{\bm{q}}} = \bm{M}^{-1}\left(\frac{\partial \bm{\tau}}{\partial \dot{\bm{q}}}-\frac{\partial ID}{\partial \dot{\bm{q}}}\right)
\end{equation}




\subsection{Newmark-\texorpdfstring{$\beta$}{beta} Method}
The Newmark $\beta$ method is another advanced implicit scheme, widely used in structural mechanics where it has been improved over the years thanks to the HHT or the Generalized-$\alpha$-method among others \cite{Cardona1994, Chung_1993}. Recently, it has been shown that the Newmark scheme is symplectic and variational \cite{Kane2000}, which may explain the reasons for its success in the FEM community. In the Newmark scheme, $\ddot{\bm{q}}_{n+1}$, $\dot{\bm{q}}_{n+1}$, and $\bm{q}_{n+1}$ are solved using equation \eqref{eq::FD} together with:
\begin{equation}
\label{eq::beta_qd_qdd}
\begin{split}
\dot{\bm{q}}_{n+1} =& \frac{\gamma}{\beta h}(\bm{q}_{n+1}-\bm{q}_n) + \left(1-\frac{\gamma}{\beta}\right)\dot{\bm{q}}_{n} + h\left(1-\frac{\gamma}{2\beta}\right)\ddot{\bm{q}}_{n}\\
\ddot{\bm{q}}_{n+1} =& \frac{1}{\beta h^2}(\bm{q}_{n+1}-\bm{q}_n) -\frac{1}{\beta h}\dot{\bm{q}}_{n} + \left(1-\frac{1}{2\beta}\right)\ddot{\bm{q}}_{n}
\end{split}
\end{equation}
where $\beta$ and $\gamma$ are parameters that can be adjusted to control the stability and accuracy of the method. For instance, $(\beta,\;\gamma) = (1/4, 1/2)$ ensures second-order accuracy with no damping. Using equation \eqref{eq::beta_qd_qdd}, we can convert \eqref{eq::FD} into a residue of $\bm{q}_{n+1}$:

\begin{equation}
\label{eq::NB_Residue}
\bm{R}_{dyn}(\bm{q}_{n+1}) = \bm{\tau}(\bm{q}_{n+1})-ID(\bm{q}_{n+1})
\end{equation}
where the Jacobian is given by:

\begin{equation}
\begin{split}
\label{eq::NB_Jacobian}
\bm{J}_{dyn}(\bm{q}_{n+1}) =&  \frac{\partial \bm{\tau}}{\partial \bm{q}}+\frac{\gamma}{\beta h}\frac{\partial \bm{\tau}}{\partial \dot{\bm{q}}} \\ 
-& \left( \frac{\partial ID}{\partial \bm{q}}+\frac{\gamma}{\beta h}\frac{\partial ID}{\partial \dot{\bm{q}}}+\frac{1}{\beta h^2}\frac{\partial ID}{\partial \ddot{\bm{q}}} \right)
\end{split}
\end{equation} 

The solution for $\bm{q}_{n+1}$ can be obtained iteratively, starting with an initial guess $\bm{q}_{{n+1}_{guess}} = \bm{q}_{n}$ and using a time step $h$. The process can be computationally efficient if the analytical Jacobian is provided. The Newmark-$\beta$ method can be efficient, particularly for index-3 Differential Algebraic Equations (DAEs) with a large number of kinematic constraints. We implemented from scratch a custom MATLAB code based on dynamic equations \eqref{eq::NB_Residue} and \eqref{eq::NB_Jacobian}. Unlike $ode15s$, the Newmark scheme works with a fixed time step, which has been chosen for each example, so that the position offset between the custom code output and $ode15s$, is of the order of millimeters.

\subsection{Computation of Static Equilibrium}
The role of the Jacobian in the statics simulation is straightforward. The residual and Jacobian for solving the static equilibrium can be derived from \eqref{eq::statics} and \eqref{eq::ID} with $\dot{\bm{q}} =\bm{0}$ and $\ddot{\bm{q}} =\bm{0}$.
\begin{subequations}
\begin{align}
\bm{R}_{sta}(\bm{q})=&\bm{\tau}-ID \label{eq::statics_res} \\
\bm{J}_{sta}(\bm{q})=&\frac{\partial \bm{\tau}}{\partial \bm{q}}-\frac{\partial ID}{\partial \bm{q}} \label{eq::staticsJacobian}
\end{align}
\end{subequations}




Numerical solvers such as $fsolve$ in MATLAB, use \eqref{eq::statics_res} with an initial guess $\bm{q}_{guess}$ and march towards a solution using the Jacobian \eqref{eq::staticsJacobian}. If the analytical Jacobian is not provided, the solver uses numerical techniques, such as finite differences, to compute it. Therefore, with the analytical derivative of the residue, the problem can be solved quickly and efficiently.

\section{Derivatives of Soft Body Dynamics}

\label{sec::Analytical Derivatives of Soft Body}

\begin{figure}
\centering
\includegraphics[width=0.9\columnwidth]{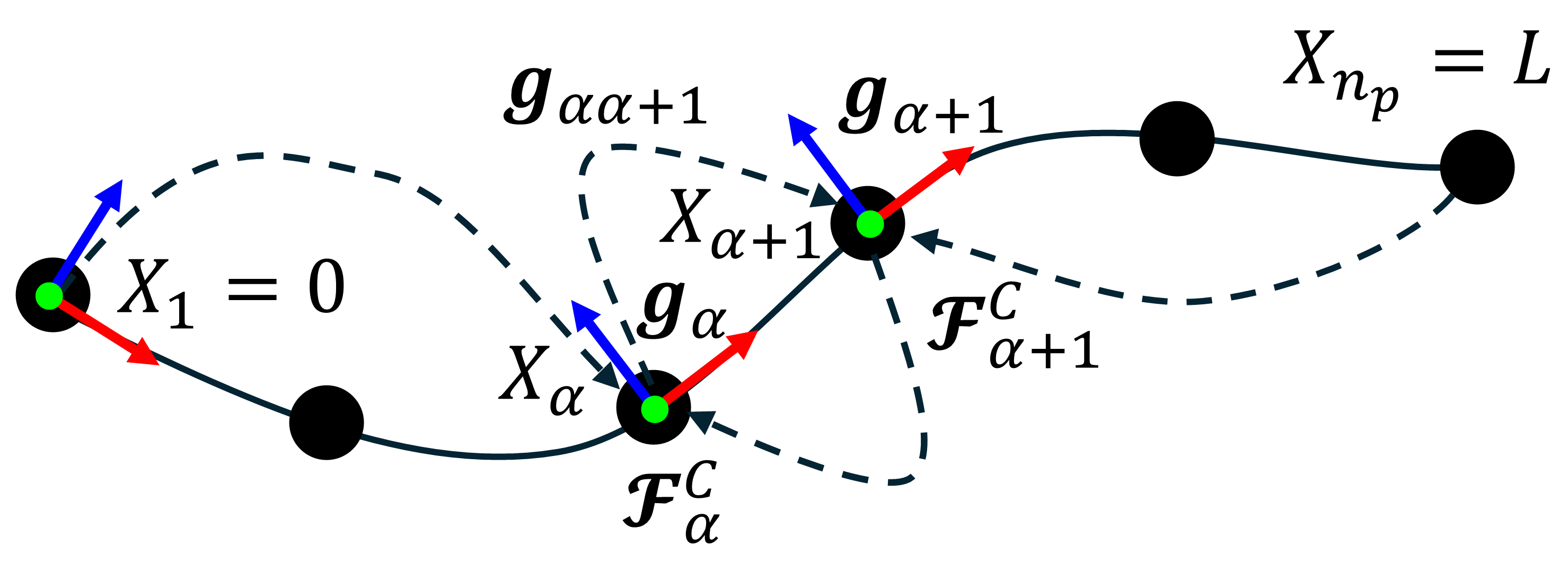}
\caption{Schematic of the RNEA for a single soft body showing the forward pass for kinematics and the backward pass for joint wrench computation. The joint motion subspace ($\bm{S}_\alpha$) projects the resultant wrench acting on the joint into the generalized coordinate space.}
\label{fig::SingleSoftBody}
\end{figure}

We begin by deriving the derivative for a single soft body and then extend this approach to a hybrid multi-body system. For simplicity, we omit the subscript $i$ in this section. 
By virtue of the Magnus expansion, the soft body is computationally equivalent to $n_p-1$ rigid joints governed by the same $\bm{q}$. The computation of the partial derivative of $FD$ requires determining the partial derivative of $ID$, according to equations \eqref{eq::dFD_dq} and \eqref{eq::dFD_dqd}. The most efficient algorithm for the computation of $ID$ is the RNEA \cite{Featherstone}. It is a two-pass algorithm with a forward pass that calculates link kinematics and a backward pass that determines the joint wrench needed to produce this motion. The schematic of RNEA for a soft body is shown in Figure \ref{fig::SingleSoftBody}. According to RNEA, the $ID$ of the discretized soft body is given by:

\begin{equation}
\label{eq::IDsum}
\begin{split}
ID = \sum_{\alpha=1}^{n_p-1} ID_\alpha =\sum_{\alpha=1}^{n_p-1} \bm{S}_\alpha^T\bm{\mathcal{F}}_\alpha^C
\end{split}
\end{equation}
where $\bm{\mathcal{F}}_\alpha^C$ is the resultant of all the dynamic (inertial and Coriolis) and external wrenches (such as gravity) acting on all points $k>\alpha$, transformed to the frame of $\alpha$ (Figure \ref{fig::SingleSoftBody}). 

\begin{equation}
\bm{\mathcal{F}}_\alpha^C = \sum_{k=\alpha+1}^{n_p}\mathrm{Ad}_{\bm{g}_{\alpha k}}^*\bm{\mathcal{F}}_k
\label{eq::FC}
\end{equation}
where $\mathrm{Ad}_{(\cdot)}^* \in \mathbb{R}^{6 \times 6} $ is coadjoint map of $SE(3)$ and $\bm{\mathcal{F}}_k$ is the resultant point wrench in the local frame of $k$, given by the sum of the inertial and Coriolis forces minus the external force. For a distributed wrench ($\overline{\bm{\mathcal{F}}}_k$), an equivalent point wrench, given by $\bm{\mathcal{F}}_k = w_k \overline{\bm{\mathcal{F}}}_k$, where $w_k$ is the quadrature weight for integration, is considered. For convenience, we drop the overbar above all the distributed quantities using this rule. We have,
\begin{equation}
\label{eq::TotalWrench}
\bm{\mathcal{F}}_k = \bm{\mathcal{M}}_k\dot{\bm{\eta}}_k+\mathrm{ad}_{\bm{\eta}_k}^*\bm{\mathcal{M}}_k\bm{\eta}_k-\bm{\mathcal{M}}_k\mathrm{Ad}_{\bm{g}_{k}}^{-1}\bm{\mathcal{G}}
\end{equation}
where, $\bm{\mathcal{M}}_k \in \mathbb{R}^{6 \times 6}$ is the screw inertia matrix, $\bm{\eta}_k$ and $\dot{\bm{\eta}}_k$ and the local velocity and acceleration twists, $\bm{\mathcal{G}} = [\bm{0}^T \bm{a}_g^T]^T$, where $\bm{a}_g$ is the acceleration due to gravity in the global frame, and $\mathrm{ad}_{(\cdot)}^*$ is the coadjoint operator (Appendix \ref{app::A}).

With this we proceed to compute the partial derivative of $ID$ with respect to the states of the robot. A summary of the results is presented below. For detailed derivations, readers are encouraged to refer to Appendix \ref{app::C} and \ref{app::D}.

\subsection{Partial Derivative of ID With Respect to Generalized Coordinates}
\label{sec::dID_dq}
Substituting \eqref{eq::FC} in \eqref{eq::IDsum}, we can see that the partial derivative of $ID_\alpha$ with respect to $\bm{q}$ is given by:
\begin{equation}
\label{eq::dtau_dq_start}
\begin{split}
\frac{\partial ID_\alpha}{\partial \bm{q}} =& \frac{\partial \bm{S}_\alpha^T}{\partial \bm{q}}\bm{\mathcal{F}}_\alpha^C+\bm{S}_\alpha^T\sum_{k=\alpha+1}^{n_p}\frac{\partial \mathrm{Ad}_{\bm{g}_{\alpha k}}^*}{\partial \bm{q}}\bm{\mathcal{F}}_k\\
+&\bm{S}_\alpha^T\sum_{k=\alpha+1}^{n_p}\mathrm{Ad}_{\bm{g}_{\alpha k}}^*\frac{\partial \bm{\mathcal{F}}_k}{\partial \bm{q}}
\end{split}
\end{equation}

The analytical formula for the first term is provided in the Appendix \ref{app::C}. The second and third terms involve sum over $k$ from $\alpha+1$ to $n_p$. The derivations of their summands are given in the Appendix \ref{app::D}. We get,

\begin{equation}
\label{eq::Adg*alphak_dq}
\frac{\partial \mathrm{Ad}_{\bm{g}_{\alpha k}}^{*}}{\partial \bm{q}}\bm{\mathcal{F}}_k = \sum_{\beta=\alpha}^{k-1}\mathrm{Ad}_{\bm{g}_{\alpha \beta}}^*\overline{\mathrm{ad}}_{\mathrm{Ad}_{\bm{g}_{\beta k}}^*\bm{\mathcal{F}}_k}^*\bm{S}_{\beta}
\end{equation}
and

\begin{equation}
\label{eq::dFk_dq_full}
\begin{split}
\frac{\partial \bm{\mathcal{F}}_k}{\partial \bm{q}} =&\bm{\mathcal{N}}_k\sum_{\beta=1}^{k-1}\mathrm{Ad}_{\bm{g}_{\beta k}}^{-1}\bm{R}_\beta+\bm{\mathcal{M}}_k\sum_{\beta=1}^{k-1}\mathrm{Ad}_{\bm{g}_{\beta k}}^{-1}\bm{Q}_\beta
\end{split}
\end{equation}
where,

\begin{subequations}
\begin{align}
\bm{\mathcal{N}}_k =& \overline{\mathrm{ad}}_{\bm{\mathcal{M}}_k\bm{\eta}_k}^*+\mathrm{ad}_{\bm{\eta}_k}^*\bm{\mathcal{M}}_k -\bm{\mathcal{M}}_k\mathrm{ad}_{\bm{\eta}_k} \label{eq::N_k} \\
\bm{R}_\beta =&\mathrm{ad}_{\bm{\eta}_\beta^+}\bm{S}_{\beta}+\frac{\partial \bm{S}_\beta}{\partial \bm{q}}\dot{\bm{q}} \label{eq::R_beta}\\
\bm{Q}_\beta=&\mathrm{ad}_{\dot{\bm{\eta}}_\beta^+}\bm{S}_{\beta}+\mathrm{ad}_{\bm{\eta}_\beta^+}\bm{R}_\beta \label{eq::Q_beta}\\
&+\mathrm{ad}_{\bm{\eta}_\beta}\frac{\partial \bm{S}_\beta}{\partial \bm{q}}\dot{\bm{q}}+\frac{\partial \dot{\bm{S}}_\beta}{\partial \bm{q}}\dot{\bm{q}}+\frac{\partial \bm{S}_\beta}{\partial \bm{q}}\ddot{\bm{q}}-\mathrm{ad}_{\mathrm{Ad}_{\bm{g}_{\beta}}^{-1}\bm{\mathcal{G}}}\bm{S}_{\beta} \nonumber
\end{align}
\end{subequations}

The definitions of the adjoint operators and the analytical formulas of $\frac{\partial \bm{S}_\beta}{\partial \bm{q}}\dot{\bm{q}}$, $\frac{\partial \dot{\bm{S}}_\beta}{\partial \bm{q}}\dot{\bm{q}}$, and $\frac{\partial \bm{S}_\beta}{\partial \bm{q}}\ddot{\bm{q}}$ are provided in the Appendices \ref{app::A} and \ref{app::C}. $\bm{\eta}_\beta^+$, $\dot{\bm{\eta}}_\beta^+$ are the velocity and acceleration twists of $\beta+1$ expressed in the frame of $\beta$. $\bm{\mathcal{N}}_k$ has the same dimensions as $\bm{\mathcal{M}}_k$ ($\mathbb{R}^{6 \times 6}$), while $\bm{R}_\beta$ and $\bm{Q}_\beta \in \mathbb{R}^{6 \times n_{dof}}$, similar to $\bm{S}_\beta$. Now we proceed to take the sum over $k$ from $\alpha+1$ to $n_p$. Substituting \eqref{eq::Adg*alphak_dq} and \eqref{eq::dFk_dq_full} into \eqref{eq::dtau_dq_start} we get:

\begin{empheq}[box=\fbox]{gather}
\label{eq::dtau_dq_end}
\begin{aligned}
\frac{\partial ID_\alpha}{\partial \bm{q}} =& \frac{\partial \bm{S}_\alpha^T}{\partial \bm{q}}\bm{\mathcal{F}}_\alpha^C\\
&+\bm{S}_\alpha^T\left(\bm{\mathcal{N}}_\alpha^C\bm{R}_{\alpha}^B+\bm{\mathcal{M}}_\alpha^C\bm{Q}_{\alpha}^B+\bm{U}_\alpha^S+\bm{P}_\alpha^S\right)
\end{aligned}
\end{empheq}
where, $\bm{R}_\alpha^B$, $\bm{Q}_\alpha^B$, $\bm{\mathcal{F}}_\alpha^C$, $\bm{\mathcal{N}}_\alpha^C$, $\bm{\mathcal{M}}_\alpha^C$,  $\bm{U}_\alpha^S$, and $\bm{P}_\alpha^S$ are computed efficiently using recursive formulas according to:
\begin{subequations}
\label{eq::recursion_q}
\begin{align}
\bm{R}_\alpha^B =& \mathrm{Ad}_{\bm{g}_{\alpha-1 \alpha}}^{-1}(\bm{R}_{\alpha-1}+\bm{R}_{\alpha-1}^B) \label{eq::RB} \\
\bm{Q}_\alpha^B=& \mathrm{Ad}_{\bm{g}_{\alpha-1 \alpha }}^{-1}(\bm{Q}_{\alpha-1}+\bm{Q}_{\alpha-1}^B)  \label{eq::QB} \\
\bm{\mathcal{F}}_\alpha^C =&\mathrm{Ad}_{\bm{g}_{\alpha \alpha+1}}^*(\bm{\mathcal{F}}_{\alpha+1}+\bm{\mathcal{F}}_{\alpha+1}^C)\\
\bm{\mathcal{N}}_\alpha^C =& \mathrm{Ad}_{\bm{g}_{\alpha \alpha+1}}^*(\bm{\mathcal{N}}_{\alpha+1}+\bm{\mathcal{N}}_{\alpha+1}^C)\mathrm{Ad}_{\bm{g}_{\alpha \alpha+1}}^{-1} \label{eq::NC} \\
\bm{\mathcal{M}}_\alpha^C=& \mathrm{Ad}_{\bm{g}_{\alpha \alpha+1}}^*(\bm{\mathcal{M}}_{\alpha+1}+\bm{\mathcal{M}}_{\alpha+1}^C)\mathrm{Ad}_{\bm{g}_{\alpha \alpha+1}}^{-1} \label{eq::MC} \\
\bm{U}_\alpha^S=& \bm{\mathcal{N}}_\alpha^C\bm{R}_\alpha+\bm{\mathcal{M}}_\alpha^C\bm{Q}_\alpha+\mathrm{Ad}_{\bm{g}_{\alpha \alpha+1}}^*\bm{U}_{\alpha+1}^S  \label{eq::OS} \\
\bm{P}_\alpha^S =& \overline{\mathrm{ad}}_{\bm{\mathcal{F}}_\alpha^C}^*\bm{S}_{\alpha}+\mathrm{Ad}_{\bm{g}_{\alpha \alpha+1}}^*\bm{P}_{\alpha+1}^S \label{eq::PS}
\end{align}
\end{subequations}

Finally, we obtain:

\begin{equation}
\label{eq::dID_dq_full}
\frac{\partial ID}{\partial \bm{q}} = \sum_{\alpha=1}^{n_p-1}\frac{\partial ID_\alpha}{\partial \bm{q}}
\end{equation}

Note that $\bm{R}_\alpha^B$ is the partial derivative of $\bm{\eta}_\alpha$, and $\bm{Q}_\alpha^B$ without the gravity term is the partial derivative of $\dot{\bm{\eta}}_\alpha$ with respect to $\bm{q}$. Quantities with the superscript $B$ are kinematic quantities determined during the forward pass, while those with superscripts $C$ or $S$ are evaluated during the backward pass. Similar to $\bm{\mathcal{F}}_\alpha^C$, $\bm{\mathcal{N}}_\alpha^C$ and $\bm{\mathcal{M}}_\alpha^C$ are the resultant $\bm{\mathcal{N}}_k$ and $\bm{\mathcal{M}}_k$ acting on all points $k>\alpha$, transformed to the point of $\alpha$. All quantities with superscript $B$ and $S$ has a dimension of $\mathbb{R}^{6 \times n_{dof}}$. Hence, the $\bm{S}_\alpha$ projection in \eqref{eq::dtau_dq_end} returns quantities with dimension $\mathbb{R}^{n{dof} \times n_{dof}}$. The same rules apply for analytical derivatives with respect to $\dot{\bm{q}}$ and $\ddot{\bm{q}}$, discussed in the next sections.

\subsection{Partial Derivative of ID With Respect to Generalized Velocities}
\label{sec::dID_dqd}
The partial derivative of $ID_\alpha$ wrt $\dot{\bm{q}}$ is given by:
\begin{equation}
\label{eq::dtau_dqd_start}
\frac{\partial ID_\alpha}{\partial \dot{\bm{q}}} =\bm{S}_\alpha^T\sum_{k=\alpha+1}^{n_p}\mathrm{Ad}_{\bm{g}_{\alpha k}}^*\frac{\partial \bm{\mathcal{F}}_k}{\partial \dot{\bm{q}}}
\end{equation}

Only the Coriolis force is a function of $\dot{\bm{q}}$. We derive (details in Appendix \ref{app::D}),

\begin{equation}
\label{eq::dFk_dqd_full}
\begin{split}
\frac{\partial \bm{\mathcal{F}}_k}{\partial \dot{\bm{q}}} 
=&\bm{\mathcal{N}}_k\sum_{\beta=1}^{k-1}\mathrm{Ad}_{\bm{g}_{\beta k}}^{-1}\bm{S}_\beta+\bm{\mathcal{M}}_k\sum_{\beta=1}^{k-1}\mathrm{Ad}_{\bm{g}_{\beta k}}^{-1}\bm{Y}_\beta
\end{split}
\end{equation}
where
\begin{equation}
\bm{Y}_\beta = \bm{R}_\beta+\mathrm{ad}_{\bm{\eta}_\beta}\bm{S}_\beta+\dot{\bm{S}}_\beta
\end{equation}

Substituting \eqref{eq::dFk_dqd_full} into \eqref{eq::dtau_dqd_start} and taking the sum over $k$ from $\alpha+1$ to $n_p$, we get:

\begin{equation}
\label{eq::dtau_dqd_end}
\boxed{\frac{\partial ID_\alpha}{\partial \dot{\bm{q}}}=\bm{S}_\alpha^T\left(\bm{\mathcal{N}}_\alpha^C\bm{S}_{\alpha}^B+\bm{\mathcal{M}}_\alpha^C\bm{Y}_{\alpha}^B+\bm{V}_\alpha^S\right)}
\end{equation}
where, $\bm{S}_\alpha^B$, $\bm{Y}_\alpha^B$, and $\bm{V}_\alpha^S$ are recursively computed according to:
\begin{subequations}
\label{eq::recursion_qd}
\begin{align}
\bm{S}_\alpha^B =& \mathrm{Ad}_{\bm{g}_{\alpha-1 \alpha}}^{-1}(\bm{S}_{\alpha-1}+\bm{S}_{\alpha-1}^B) \label{eq::SB} \\
\bm{Y}_\alpha^B=& \mathrm{Ad}_{\bm{g}_{\alpha-1 \alpha}}^{-1}(\bm{Y}_{\alpha-1}+\bm{Y}_{\alpha-1}^B)  \label{eq::UB} \\
\bm{V}_\alpha^S=& \bm{\mathcal{N}}_\alpha^C\bm{S}_\alpha+\bm{\mathcal{M}}_\alpha^C\bm{Y}_\alpha+\mathrm{Ad}_{\bm{g}_{\alpha \alpha+1}}^*\bm{V}_{\alpha+1}^S  \label{eq::LS} 
\end{align}
\end{subequations}

Note that $\bm{S}_\alpha^B$ is the partial derivative of $\bm{\eta}_\alpha$, and $\bm{Y}_\alpha^B$ is the partial derivative of $\dot{\bm{\eta}}_\alpha$ with respect to $\dot{\bm{q}}$. This implies that, $\bm{J}_\alpha = \bm{S}_\alpha^B$ and $\dot{\bm{J}}_\alpha = \bm{Y}_\alpha^B-\bm{R}_\alpha^B$.

Summing \eqref{eq::dtau_dqd_end} over $\alpha$ from 1 to $n_p-1$, we have:

\begin{equation}
\label{eq::dID_dqd_full}
\frac{\partial ID}{\partial \dot{\bm{q}}} = \sum_{\alpha=1}^{n_p-1}\frac{\partial ID_\alpha}{\partial \dot{\bm{q}}}
\end{equation}

\subsection{Partial Derivative of ID With Respect to Generalized Accelerations}
\label{sec::dID_dqdd}
The partial derivative of $ID_\alpha$ with respect to $\ddot{\bm{q}}$ is given by:
\begin{equation}
\label{eq::dtau_dqdd_start}
\frac{\partial ID_\alpha}{\partial \ddot{\bm{q}}} =\bm{S}_\alpha^T\sum_{k=\alpha+1}^{n_p}\mathrm{Ad}_{\bm{g}_{\alpha k}}^*\frac{\partial \bm{\mathcal{F}}_k}{\partial \ddot{\bm{q}}}
\end{equation}

Only the inertial force is a function of $\ddot{\bm{q}}$. From \eqref{eq::TotalWrench}, we get:
\begin{equation}
\label{eq::dFk_dqdd_full}
\frac{\partial \bm{\mathcal{F}}_k}{\partial \ddot{\bm{q}}} =\bm{\mathcal{M}}_k\sum_{\beta=1}^{k-1}\mathrm{Ad}_{\bm{g}_{\beta k}}^{-1}\bm{S}_\beta
\end{equation}

Substituting this into \eqref{eq::dtau_dqdd_start}, we get:
\begin{equation}
\label{eq::dtau_dqdd_end}
\boxed{\frac{\partial ID_\alpha}{\partial \ddot{\bm{q}}} =\bm{S}_\alpha^T\left(\bm{\mathcal{M}}_\alpha^C\bm{S}_{\alpha}^B+\bm{W}_\alpha^S\right)}
\end{equation}
where, $\bm{W}_\alpha^S$ is computed recursively according to:
\begin{equation}
\label{eq::WS} 
\bm{W}_\alpha^S= \bm{\mathcal{M}}_\alpha^C\bm{S}_\alpha+\mathrm{Ad}_{\bm{g}_{\alpha \alpha+1}}^*\bm{W}_{\alpha+1}^S  
\end{equation}

Finally summing over all $\alpha$, we get:

\begin{equation}
\label{eq::dID_dqdd_full}
\frac{\partial ID}{\partial \ddot{\bm{q}}} = \sum_{\alpha=1}^{n_p-1}\frac{\partial ID_\alpha}{\partial \ddot{\bm{q}}}=\bm{M}(\bm{q})
\end{equation}

Note that \eqref{eq::dID_dqdd_full}, incidentally, represents a novel way of computing the generalized mass matrix of a soft robot.

\subsection{Partial Derivatives of Internal Forces}
\label{sec::dtau_dq}
The internal forces include the elastic, damping, and actuation forces. It is given by:
\begin{equation}
\label{eq::internalforce_components}
\bm{\tau} = \bm{B}\bm{u}-\bm{K}\bm{q}-\bm{D}\dot{\bm{q}}
\end{equation}
where $ \bm{D} \in \mathbb{R}^{n_{dof} \times n_{dof}} $ is the generalized damping matrix, $ \bm{K} \in \mathbb{R}^{n_{dof} \times n_{dof}} $ is the generalized stiffness matrix, $ \bm{B}(\bm{q}) \in \mathbb{R}^{n_{dof} \times n_a} $ ($ n_{a} $ being the total number of actuators) is the generalized actuation matrix

The actuation matrix ($\bm{B}$) for tendon-like actuators was derived in \cite{Renda_RAL2020, Renda2024}, while a simple Hooke-like linear elastic and damping model is used for computing $\bm{K}$ and $\bm{D}$. The expressions for $\bm{B}$, $\bm{K}$, and $\bm{D}$ can be found in the Appendix \ref{app::D}.

The partial derivative of $\bm{\tau}$ with respect to $\bm{q}$ and $\dot{\bm{q}}$ are given by:
\begin{subequations}
\begin{align}
\frac{\partial \bm{\tau}}{\partial \bm{q}} =& \frac{\partial \bm{B}}{\partial \bm{q}}\bm{u}-\bm{K} \label{eq::dtau_dq}\\
\frac{\partial \bm{\tau}}{\partial \dot{\bm{q}}} =& -\bm{D} \label{eq::dtau_dqd}
\end{align}
\end{subequations}
For tendon-like actuators, the partial derivative of $\bm{B}\bm{u}$ with respect to $\bm{q}$ is provided in the Appendix \ref{app::D}.

\subsection{Cable-Driven Soft Manipulator}
\label{sec::CDM}
\begin{figure}
\centering
\includegraphics[width=1\columnwidth]{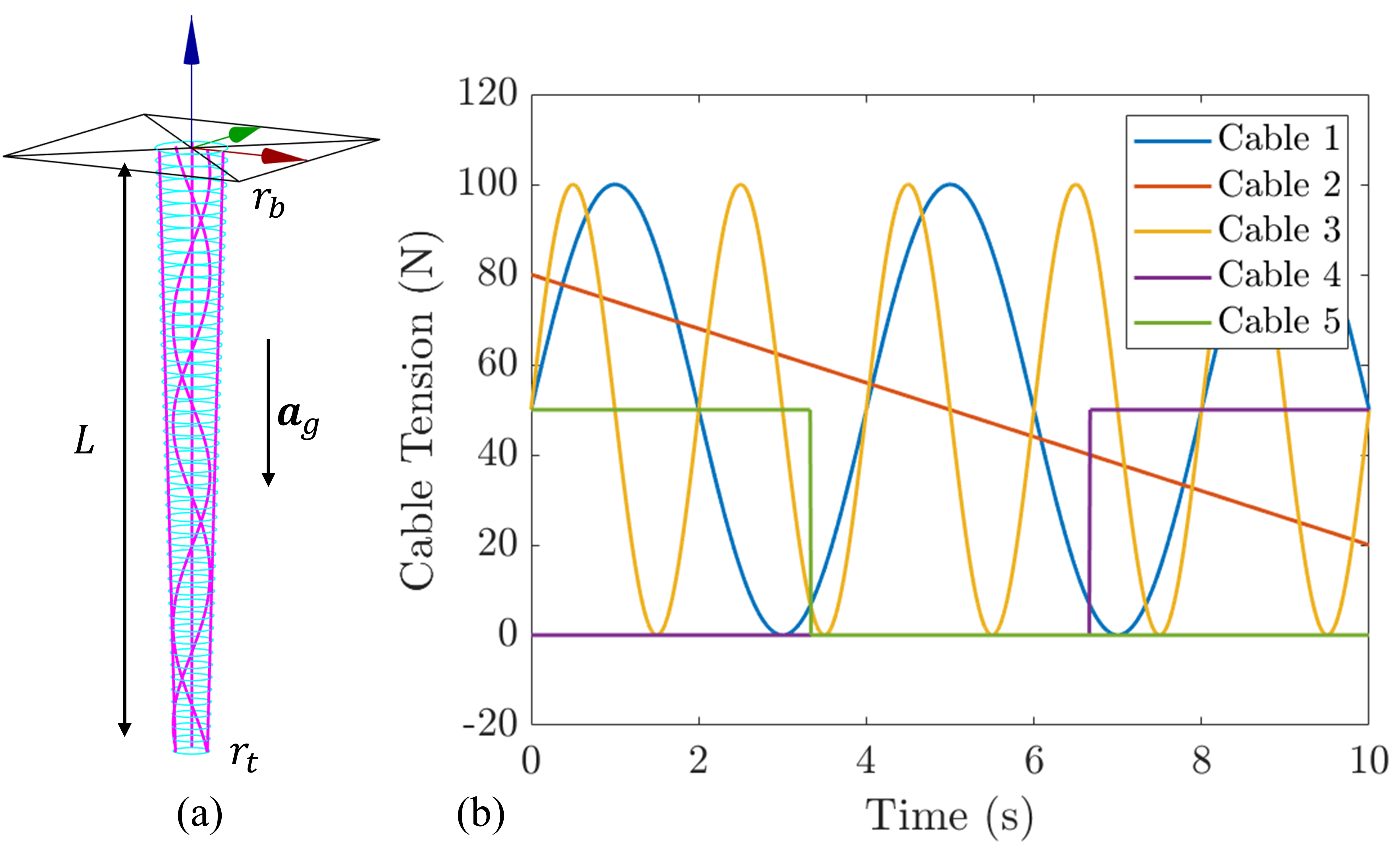}
\caption{(a) Schematics of cable-driven soft manipulator with cable routing. $L = 50\;\text{cm}$, base radius $r_b = 3 \;\text{cm}$, and tip radius $r_t = 1.5\;\text{cm}$. RGB arrows indicate $x$, $y$, and $z$ axes of the global frame. (b) Actuation input used for the simulation.}
\label{fig::CDM}
\end{figure}

Based on the results so far, we set up the first simulation example. We simulate a cable-driven soft manipulator (CDM) shown in Figure \ref{fig::CDM}(a). The manipulator radius varies linearly from base to tip according to $r(X)=r_b+(r_t-r_b)X/L$. The material properties used in the simulation are as follows: Young's modulus of 1 MPa, Poisson's ratio of 0.5, density of $1000 \; \text{kg/m}^3$, and a material damping coefficient of $10^4 \; \text{Pa} \cdot \text{s}$. The soft manipulator has five tendon-like actuators, as shown in Figure \ref{fig::CDM}(a). Their routing, local coordinates for a given $X$, are provided in Table \ref{tab::Cables}. Gravity acts in the negative $z$-axis direction. We used a Legendre polynomial basis with 4th-order angular strains (torsion about $x$, bending about $y$ and $z$) and 2nd-order linear strains (elongation about $x$, shear in $y$ and $z$). Hence, the total DoF of the system is 24. For the numerical integration, we used 5 Gauss quadrature points. Including the computational points at the base and the tip, the total number of points $n_p = 7$.

\begin{table}[ht]
\centering
\caption{Cable routing for the cable-driven manipulator. $r(X)$ is the manipulator radius and $r_t=r(L)$. }
\begin{tabular}{|c|c|c|}
\hline
\begin{tabular}[c]{@{}c@{}}Cable\\ Number\end{tabular} &
  \begin{tabular}[c]{@{}c@{}}y-Coordinate\end{tabular} &
  \begin{tabular}[c]{@{}c@{}}z-Coordinate\end{tabular} \\ \hline
1 & $0$      & $r(X)$          \\ 
2 & $-\frac{\sqrt{3}}{2}r(X)$     & $-\frac{1}{2}r(X)$          \\ 
3 & $\frac{\sqrt{3}}{2}r(X)$     & $-\frac{1}{2}r(X)$  \\ 
4 & $r_t sin\left(\frac{4\pi}{L}X\right)$         & $r_t cos\left(\frac{4\pi}{L}X\right)$ \\ 
5 & $r_t sin\left(\frac{4\pi}{L}X\right)$         & $-r_t cos\left(\frac{4\pi}{L}X\right)$ \\ \hline
\end{tabular}
\label{tab::Cables}
\end{table}

The manipulator is actuated by arbitrary time-varying inputs for 10 s, as shown in Figure \ref{fig::CDM}(b). We used $ode15s$ of MATLAB as the ODE integrator, as it demonstrated the fastest computational speed for this simulation. The $\dot{\bm{x}}$ for the ODE is computed using \eqref{eq::FD}, while its Jacobian is computed using \eqref{eq::dxd_dx}, \eqref{eq::dFD_dq}, \eqref{eq::dFD_dqd} and analtyical derivatives \eqref{eq::dID_dq_full}, \eqref{eq::dID_dqd_full}, \eqref{eq::dID_dqdd_full}, \eqref{eq::dtau_dq} and \eqref{eq::dtau_dqd}. To validate and compare the impact of analytical derivatives, the ode15s solver was executed both without (method 1) and with (method 2) their inclusion. For the Newmark-$\beta$ scheme (method 3), a time step $h$ of 0.002 s is used in this example. Throughout all simulations, the default tolerances in MATLAB were used for $ode15s$ and $fsolve$: `RelTol' of $10^{-3}$, and `AbsTol' of $10^{-6}$. The PC specifications for all computations presented here are as follows: 13th Gen Intel(R) Core(TM) i9-13900HX, 2.20 GHz, 64.0 GB RAM. The MATLAB version used is R2024a.

\begin{figure*}[ht]
\centering
\includegraphics[width=\textwidth]{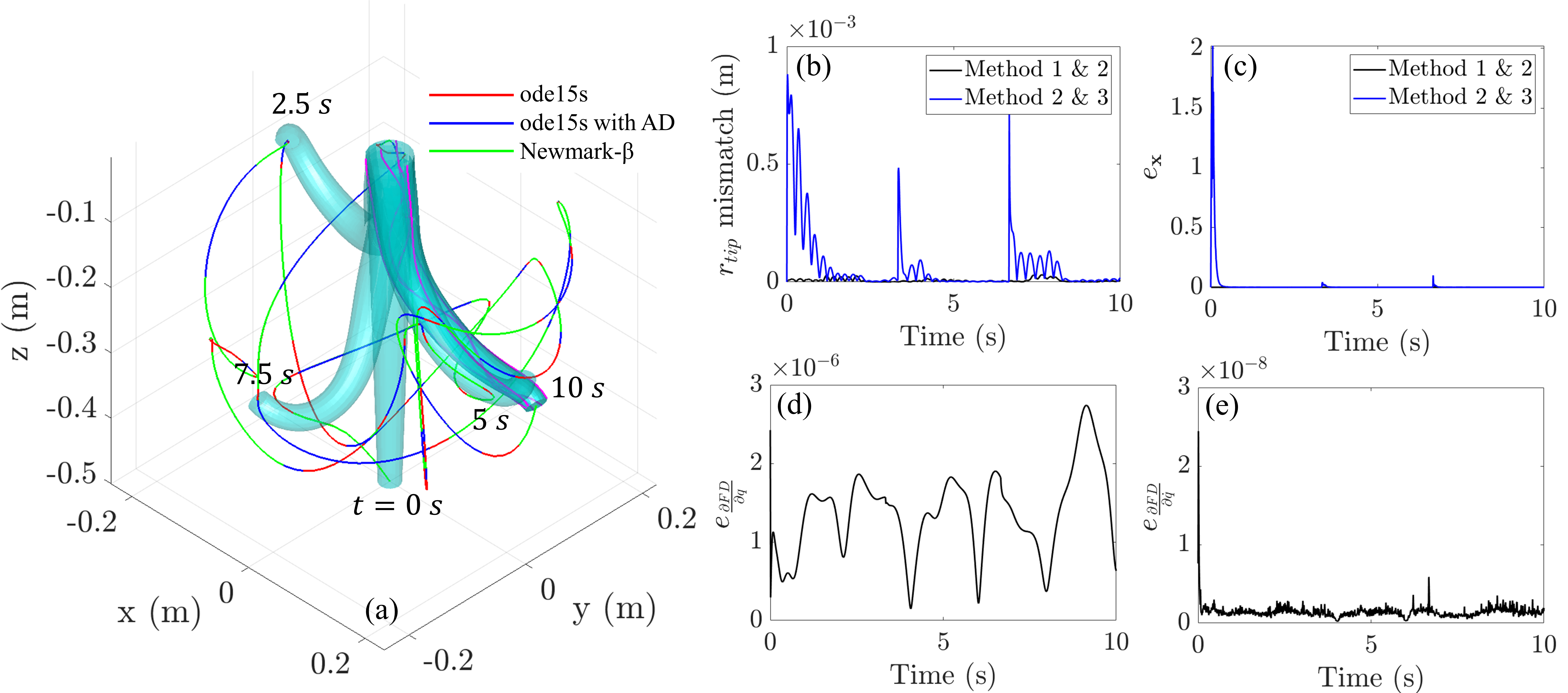}
\caption{Dynamic response of the CDM: (a) Snapshots of the manipulator at different times and the tip trajectory with and without using analytical derivatives (AD) of $FD$. (b) Mismatch between the tip positions ($\bm{r}_{tip}$). (c) State mismatch. Mismatch between numerical and analytical derivatives of $FD$ (d) with respect to $\bm{q}$ and (e) with respect to $\dot{\bm{q}}$}
\label{fig::Exp1_dynamics}
\end{figure*}

Figure \ref{fig::Exp1_dynamics} shows the simulation results for three cases. The dynamic response is very close for all three methods. Figure \ref{fig::Exp1_dynamics}(a) plots the states of the robot and the tip trajectory. The tip position mismatch computed using $\|\Delta\bm{r}_{tip}\|$ is plotted in Figure \ref{fig::Exp1_dynamics}(b). The maximum error is less than $30 \; \mu \text{m}$ between methods 1 and 2, while it is less than $1 \; \text{mm}$ between methods 2 and 3. Similarly, the state mismatch between the first two methods is on the order of $10^{-4}$, while the mismatch between the second and third methods is higher, as shown in Figure \ref{fig::Exp1_dynamics}(c). The following metric is used to calculate the mismatch between the states of the body, obtained from two different methods, at a given time instant:

\begin{equation}
\label{eq::mismatch}
e_{\bm{x}} = \frac{\|\Delta\bm{x}\|}{\|avg(\bm{x})\|}
\end{equation}

Without analytical derivatives, $ode15s$ took $5.40 \; \text{s}$, whereas, with analytical derivatives, it was completed in $1.25 \; \text{s}$. This indicates that using analytical derivatives makes the computation more than four times faster. Simulation using the Newmark-$\beta$ method was completed in 24.43 s, which can be explained by the fact that this scheme has a fixed time-step in contrast to $ode15s$ whose time-step is adaptive.

The analytical derivatives of the forward dynamics (FD) are validated as follows. Using the finite difference method, we calculated the numerical derivatives of $FD(\bm{q}, \dot{\bm{q}}, \bm{u})$ with respect to $\bm{q}$ and $\dot{\bm{q}}$. The $i^{th}$ column of the numerical Jacobians is computed using the forward finite difference method:

\begin{subequations}
\begin{equation}
\left( \frac{\partial FD}{\partial q_i} \right)_{num} = \frac{FD(\bm{q} + \epsilon \bm{n}_i, \dot{\bm{q}}, \bm{u}) - FD(\bm{q}, \dot{\bm{q}}, \bm{u})}{\epsilon}
\end{equation}
\begin{equation}
\left( \frac{\partial FD}{\partial \dot{q_i}} \right)_{num} = \frac{FD(\bm{q}, \dot{\bm{q}} + \epsilon \bm{n}_i, \bm{u}) - FD(\bm{q}, \dot{\bm{q}}, \bm{u})}{\epsilon}
\end{equation}
\end{subequations}
where $\bm{n}_i$ is the unit vector in the direction of the $i$-th component, and $\epsilon$ is a perturbation value set to $10^{-6}$. We used the forward finite difference method as it requires the least function evaluations to compute numerical derivatives, though this may result in lower accuracy compared to other methods.

Using a similar metric like \eqref{eq::mismatch}, we computed the mismatch between analytical and numerical Jacobian at every time step. The results are plotted in Figure \ref{fig::Exp1_dynamics}(d) and (e). Numerical values on the order of $10^{-6}$ and $10^{-8}$ indicate that their differences are insignificant. Computing the Jacobians numerically took, on average, 10.5 ms, while the analytical Jacobian required only 1.3 ms, making the analytical method more than 8 times faster.

\begin{figure}
\centering
\includegraphics[width=1\columnwidth]{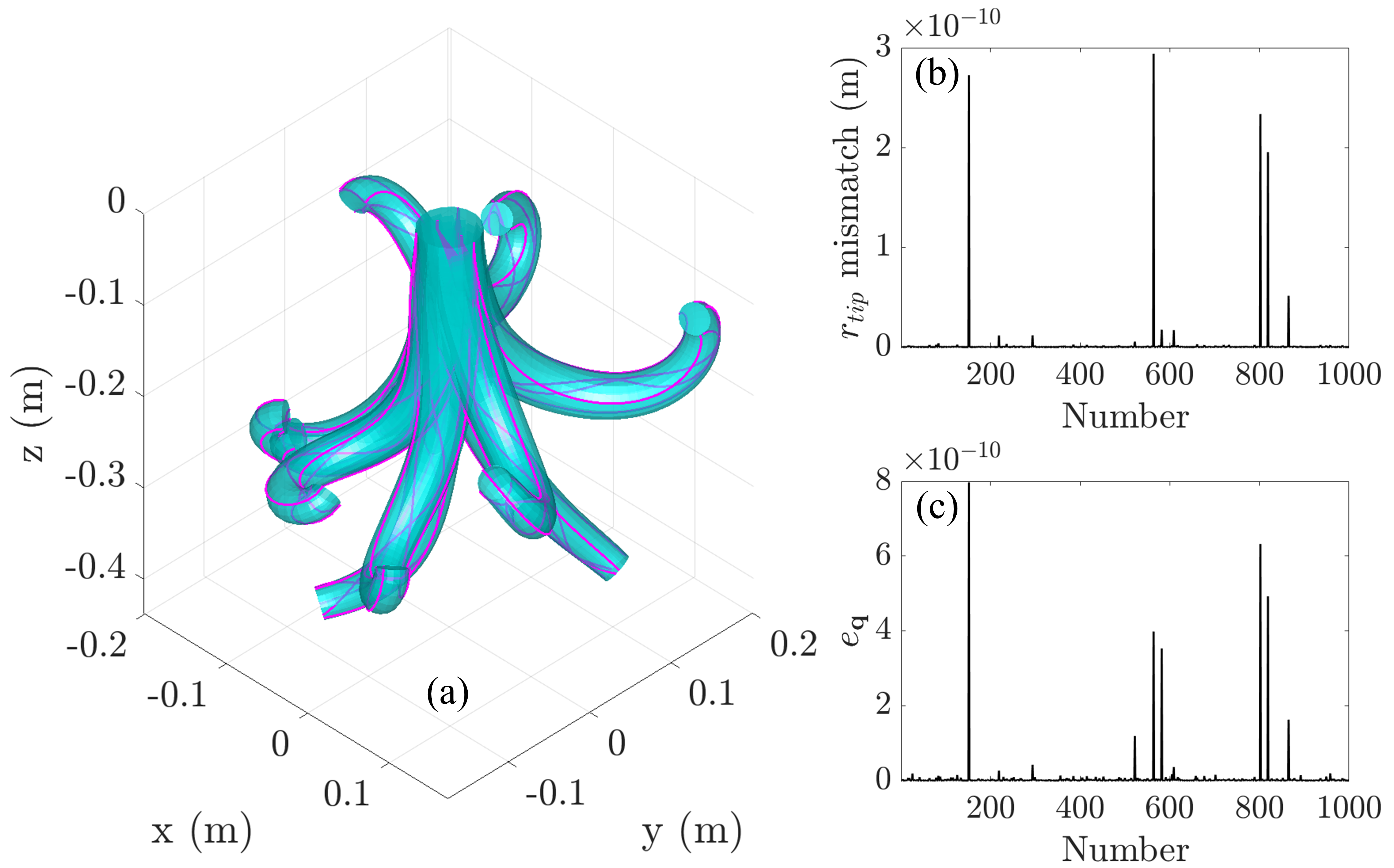}
\caption{Static simulation results: (a) Ten arbitrary equilibrium shapes. (b) The mismatch between tip positions. (c) The mismatch static equilibrium solutions.}
\label{fig::Exp1_statics}
\end{figure}

For the static equilibrium simulation, the residue is computed using \eqref{eq::statics_res}, while the analytical Jacobian is obtained from \eqref{eq::staticsJacobian}, \eqref{eq::dID_dq_full}, and \eqref{eq::dtau_dq} with $\dot{\bm{q}}=\bm{0}$ and $\ddot{\bm{q}}=\bm{0}$. We performed 1000 static equilibrium simulations on the same system. In each simulation, we used random values of cable tensions in the range of 0 to 100 N. Figure \ref{fig::Exp1_statics}(a) illustrates 10 equilibrium shapes from these 1000 simulations. Both methods (with and without the analytical Jacobian) yielded identical results. The mismatch in tip positions and static equilibrium solutions ($e_{\bm{q}}$) using the two approaches are shown in Figure \ref{fig::Exp1_statics}(b) and (c). Without analytical derivatives, the static simulations took an average of 28.2 ms, whereas, with analytical derivatives, it took 3.3 ms, making the computation more than eight times faster.

\section{Extension to Hybrid Multi-body System}
\label{sec::Analytical Derivatives of Hybrid multi-body}

A hybrid multi-body system may incorporate rigid joints (up to 6 DoF), rigid links, soft links, branched chains, and closed-chain joints (Figure \ref{fig::multi-body}). The forward pass of RNEA goes through the whole multibody ($i=1$ to $N$) and the backward pass from $i=N$ to $1$. We propose the following rules to simplify the analysis of hybrid multi-body systems:

\begin{figure}
\centering
\includegraphics[width=0.9\columnwidth]{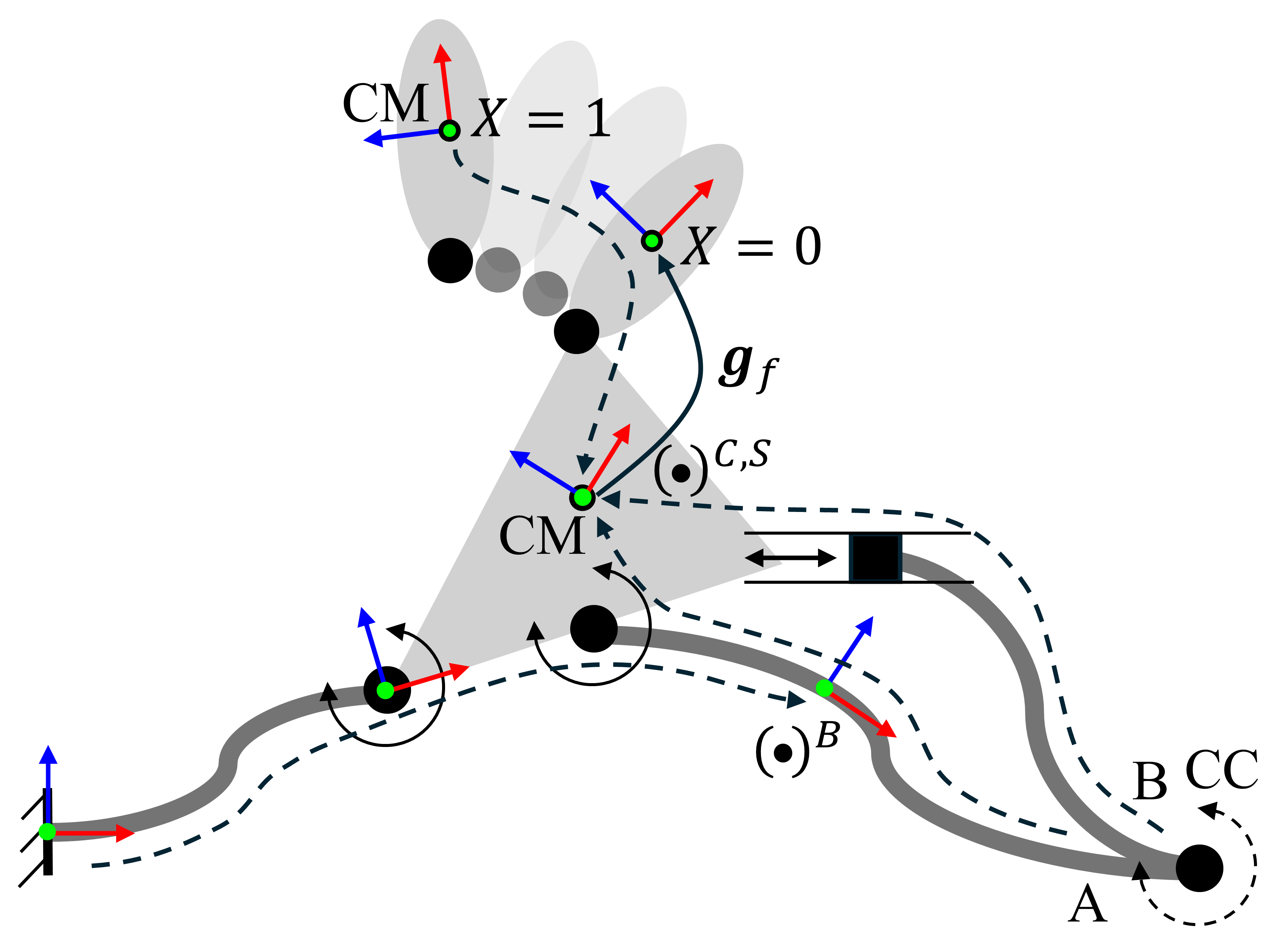}
\caption{Schematics of a generic hybrid multi-body system featuring soft and rigid links, rigid joints, branched chains, and closed-chain (CC) joints.}
\label{fig::multi-body}
\end{figure}

\begin{enumerate}
\setcounter{enumi}{0}
    
    \item Each link is defined by a rigid joint and a body (rigid or soft). Accordingly, every rigid link has two computational points: one at $X=0$ and another at $X=1$ of the joint. These points are defined in the body’s reference frame, typically located at the center of mass (CM), where the inertia matrix is computed. Soft links have $n_p+1$ computational points: at $X=0$ of the rigid joint and $n_p$ along the soft body. $X=1$ of the joint coincides with $X=0$ of the soft body.

    \item For a rigid body, the inertial and gravitational forces act on the computational frame on the joint at $X=1$. At $X=0$ of a rigid joint, $\bm{\mathcal{F}}_k = \bm{0}$, $\bm{\mathcal{M}}_k = \bm{0}$ and $\bm{\mathcal{N}}_k = \bm{0}$.
    
    \item Let $\bm{g}_f$ denote a fixed transformation between two adjacent frames (Figure \ref{fig::multi-body}). The recursive transformation rule, as outlined in equations \eqref{eq::recursion_q}, \eqref{eq::recursion_qd}, and \eqref{eq::WS}, applies to both the forward and backward transformations between these two frames when $\bm{g}_{\alpha-1 \alpha}$ or $\bm{g}_{\alpha \alpha+1}$ is replaced by $\bm{g}_f$.

    \item In a branched chain system, the kinematic term $(\bullet)^B$ denotes the contribution of all joints from the global frame to the computational point $\alpha$, following a serial chain. Meanwhile, $(\bullet)^C$ and $(\bullet)^S$ encompass all the children's joints, including all branches ahead of point $\alpha$. The dashed arrows in Figure \ref{fig::multi-body} illustrate these points.

    \item Computation of $(\bullet)^B$ in the forward pass follows the same rule of the geometric Jacobian in \eqref{eq::Geometric Jacobian}:

    \begin{equation}
    \begin{split}
        (\bullet)_{i, \alpha}^B =& \mathrm{Ad}_{\bm{g}_{\alpha \alpha-1}}\Bigl((\bullet)_{i, \alpha-1}^B \\ 
        &+ \big[ \bm{0}_{6 \times n_i^-} \; (\bullet)_{i,\alpha-1} \; \bm{0}_{6 \times n_i^+} \big] \Bigr)
    \end{split}
    \end{equation}

    Hence, while the joint quantities, $\bm{S}_\alpha$, $\bm{R}_\alpha$, $\bm{Q}_\alpha$, and $\bm{Y}_\alpha$ have dimensions $\mathbb{R}^{6 \times {n_{dof}}_i}$, their $(\bullet)^B$ counterparts have dimensions $\mathbb{R}^{6 \times n_{dof}}$.

    \item Computation of $(\bullet)^S$ in the backward pass follows a similar rule:
    
    \begin{equation}
    (\bullet)_{i,\alpha}^S = \big[ \bm{0}_{6 \times n_i^-} \; (\bullet)_{i,\alpha} \; \bm{0}_{6 \times n_i^+} \big]+\mathrm{Ad}_{\bm{g}_{\alpha \alpha+1}}^*(\bullet)_{i,\alpha+1}^S 
    \end{equation}

    Hence, $(\bullet)^S$ also has dimensions $\mathbb{R}^{6 \times n_{dof}}$.

    \item For the $i^{th}$ Cosserat rod, equations \eqref{eq::dtau_dq_end}, \eqref{eq::dtau_dqd_end}, and \eqref{eq::dtau_dqdd_end} are generalized for the multi-body as follows:

    \begin{subequations}
    \begin{align}
    &\frac{\partial ID_{i,\alpha}}{\partial \bm{q}} = \left[\bm{0}_{{n_{dof}}_i \times n_i^-} \; \frac{\partial \bm{S}_{i,\alpha}^T}{\partial \bm{q}_i}\bm{\mathcal{F}}_{i,\alpha}^C \; \bm{0}_{{n_{dof}}_i \times n_i^+} \right]\\
    \nonumber &+\bm{S}_{i,\alpha}^T\left(\bm{\mathcal{N}}_{i,\alpha}^C\bm{R}_{i,\alpha}^B+\bm{\mathcal{M}}_{i,\alpha}^C\bm{Q}_{i,\alpha}^B+\bm{U}_{i,\alpha}^S+\bm{P}_{i,\alpha}^S\right)\\
    &\frac{\partial ID_{i,\alpha}}{\partial \dot{\bm{q}}}=\bm{S}_{i,\alpha}^T\left(\bm{\mathcal{N}}_{i,\alpha}^C\bm{S}_{i,\alpha}^B+\bm{\mathcal{M}}_{i,\alpha}^C\bm{Y}_{i,\alpha}^B+\bm{V}_{i,\alpha}^S\right) \\
    &\frac{\partial ID_{i,\alpha}}{\partial \ddot{\bm{q}}} =\bm{S}_{i,\alpha}^T\left(\bm{\mathcal{M}}_{i,\alpha}^C\bm{S}_{i,\alpha}^B+\bm{W}_{i,\alpha}^S\right)
    \end{align}
    \end{subequations}
    
    \item Finally, the partial derivatives of $ID$ and $\bm{\tau}$ are computed by vertically concatenating row-blocks of each Cosserat rod according to:
    \begin{subequations}
    \begin{align}
        \frac{\partial ID}{\partial \bm{q}} =& \big\Vert_{i=1}^N \sum_{\alpha=1}^{n_p-1} \frac{\partial ID_{i,\alpha} }{\partial \bm{q}}\\
        \frac{\partial ID}{\partial \dot{\bm{q}}} =& \big\Vert_{i=1}^N \sum_{\alpha=1}^{n_p-1} \frac{\partial ID_{i,\alpha}}{\partial \dot{\bm{q}}}\\
        \frac{\partial ID}{\partial \ddot{\bm{q}}}  =& \big\Vert_{i=1}^N \sum_{\alpha=1}^{n_p-1} \frac{\partial ID_{i,\alpha}}{\partial \ddot{\bm{q}}} = \bm{M}\\
        \frac{\partial \bm{\tau}}{\partial \bm{q}} =& \big\Vert_{i=1}^N \frac{\partial \bm{\tau}_i}{\partial \bm{q}}\\
        \frac{\partial \bm{\tau}}{\partial \dot{\bm{q}}} =& \big\Vert_{i=1}^N \frac{\partial \bm{\tau}_i}{\partial \dot{\bm{q}}}
    \end{align}
    \end{subequations}
    where $\big\Vert$ is the vertical concatenation operator. In the backward pass from $N$ to 1, the corresponding derivatives of each Cosserat rod (as block matrices of size $\mathbb{R}^{{n_{dof}}_i \times n_{dof}}$) are concatenated from bottom to top.
\end{enumerate}

Analytical derivatives of arbitrary hybrid multi-body systems can be evaluated following these rules. In addition to these, the dynamics of a multi-body system is often subject to equality constraints that can be expressed as $\bm{c}(\bm{q}, \dot{\bm{q}}) = \bm{0}$. These constraints introduce an equal number of unknown Lagrange multipliers to enforce the equality condition. As a result, the system's ODE is transformed into a DAE. We explore two of the most commonly used types of constraints in robotic systems: joint coordinate controlled rigid joints and closed-chain systems.

\subsection{Joint Coordinate Controlled Rigid Joints}
\label{sec::JointCoordinateControlled}

 Rigid joints can be actuated by specifying joint wrench $\bm{u}$ or providing joint coordinates. The latter can be expressed as an equality constraint, with the Lagrange multiplier being $\bm{u}$. Similar to the soft body actuation \eqref{eq::internalforce_components}, the generalized actuation force of a rigid joint is given by $\bm{B}\bm{u}$. The expression of the generalized actuation matrix ($\bm{B}$) and the derivative of $\bm{B}\bm{u}$ with respect to $\bm{q}$ are provided in the Appendix \ref{app::D}. For 1 DoF joints, $\bm{B}$ is independent of $\bm{q}$, and therefore, its derivative is $\bm{0}$.
 
 The computation of $FD$ and its derivatives remains the same if the joint is actuated by specifying $\bm{u}$. However, if some of the joints are actuated by specifying joint coordinates, the $FD$ computation is different. One approach is to split $\ddot{\bm{q}}$ into $\ddot{\bm{q}}_u$ and $\ddot{\bm{q}}_k$ and accordingly, $\bm{u}$ into $\bm{u}_k$ and $\bm{u}_u$. The subscripts $k$ and $u$ stand for known and unknown quantities. Note that $\ddot{\bm{q}}_k$ is the second time derivative of the specified joint coordinates $\bm{q}_k=\bm{f}(t)$, making it an index-1 DAE. Bringing the unknown terms to the LHS and known terms to the RHS, the generalized dynamics equation can be rewritten as:

\begin{equation}
\label{eq::FD_jointactuation}
[\bm{M}_u\; -\bm{B}_u]\begin{bmatrix}
\ddot{\bm{q}}_u \\
\bm{u}_u
\end{bmatrix}=[\bm{B}_k\; -\bm{M}_k]\begin{bmatrix}
\bm{u}_k\\
\ddot{\bm{q}}_k
\end{bmatrix}
+(\bullet)
\end{equation}
where, the subscripts of $\bm{M}$ and $\bm{B}$ indicate corresponding columns of known and unknown quantities.

Equation \eqref{eq::FD_jointactuation} is used to solve $[\ddot{\bm{q}}_u^T\;\bm{u}_u^T]^T$ and $FD$ is found by combining $\ddot{\bm{q}}_u$ and $\ddot{\bm{q}}_k$. To ensure numerical stability, during each iteration of the $FD$, the known joint angles and their derivatives are replaced with the specified $\bm{q}_k$ and its time derivative $\dot{\bm{q}}_k$. The derivative of \eqref{eq::FD_jointactuation} with respect to $\bm{q}$ gives:

\begin{equation}
\label{eq::dFD_jointactuation_dx}
\begin{bmatrix}
\frac{\partial \ddot{\bm{q}}_u}{\partial \bm{q}_u} \\
\frac{\partial \bm{u}_u}{\partial \bm{q}_u}
\end{bmatrix}=[\bm{M}_u\; -\bm{B}_u]^{-1}\left(\frac{\partial \bm{\tau}}{\partial \bm{q}_u}-\frac{\partial ID}{\partial \bm{q}_u}\right)
\end{equation}

Finally, $\frac{\partial FD}{\partial \bm{q}}$ is found by combining $\frac{\partial \ddot{\bm{q}}_u}{\partial \bm{q}_u}$ with $\frac{\partial \ddot{\bm{q}}_k}{\partial \bm{q}} = \bm{0}^{n_k \times n_{dof}}$ and $\frac{\partial \ddot{\bm{q}}_u}{\partial \bm{q}_k} = \bm{0}^{n_{u} \times n_{k}}$. The partial derivative of $FD$ with respect to $\dot{\bm{q}}$ is computed similarly.

The same problem can be solved using index-3 DAE formulation using the Newmark-$\beta$ scheme. In this case, we have $\bm{q}_k = \bm{f}(t)$ and the residue \eqref{eq::NB_Residue} and Jacobian \eqref{eq::NB_Jacobian} are modified to include the unknown $\bm{u}_{u,n+1}$ and partial derivative with respect to $\bm{q}_{u,n+1}$ and $\bm{u}_{u,n+1}$.

\begin{subequations}
\begin{equation}
\bm{R}_{dyn} = \bm{\tau}(\bm{q}_{n+1},\;\bm{u}_{n+1}) - ID(\bm{q}_{n+1})
\label{eq::NB_Residue_jq}
\end{equation}
\begin{equation}
\bm{J}_{dyn} = 
\begin{bmatrix}  
\frac{\partial \bm{\tau}}{\partial \bm{q}_u} -\frac{\partial ID}{\partial \bm{q}_u} & \bm{B}_u
\end{bmatrix}
\label{eq::NB_Jacobian_jq}
\end{equation}
\end{subequations}

The statics problem involves finding the unknown generalized coordinates and joint forces ($\bm{q}_u$ and $\bm{u}_u$). It can be seen that the residue and Jacobian of the static equilibrium are identical to \eqref{eq::NB_Residue_jq} and \eqref{eq::NB_Jacobian_jq} with $\dot{\bm{q}}$ and $\ddot{\bm{q}}$ set to zero.

\begin{figure}
\centering
\includegraphics[width=1\columnwidth]{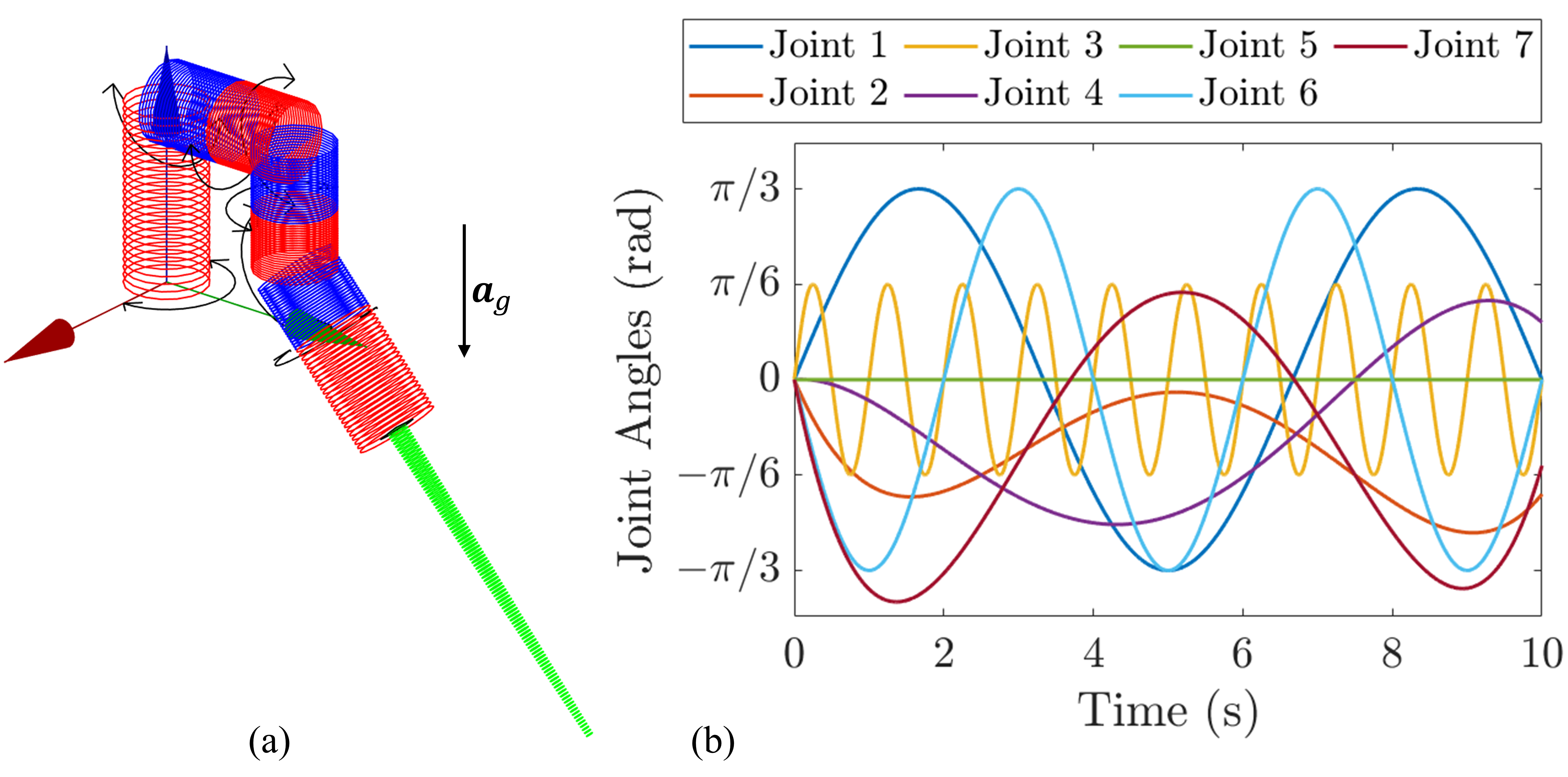}
\caption{(a) Hybrid serial robot consisting of 7 rigid links and 1 soft link with a fixed joint. The rigid links shown in red have revolute joints rotating about their local x-axis, while those in blue rotate about the y-axis. (b) Arbitrary joint angles as a function of time.}
\label{fig::HybridSerialRobot}
\end{figure}

We modeled a hybrid serial robot by combining a rigid serial robotic arm with a soft manipulator, as shown in Figure \ref{fig::HybridSerialRobot}(a). The robot consists of 7 revolute joints that are actuated by arbitrary joint angles according to Figure \ref{fig::HybridSerialRobot}(b). The soft link has a fixed joint, a length of 50 cm, and a radius linearly varying from 1.5 cm to 1 cm. The mechanical properties of the soft body are identical to that of the first example. The soft body is modeled as a Kirchhoff rod (no shear deformation) with a fourth-order strain basis. Hence, in total, the robot has 27 DoF.

\begin{figure*}[ht]
\centering
\includegraphics[width=\textwidth]{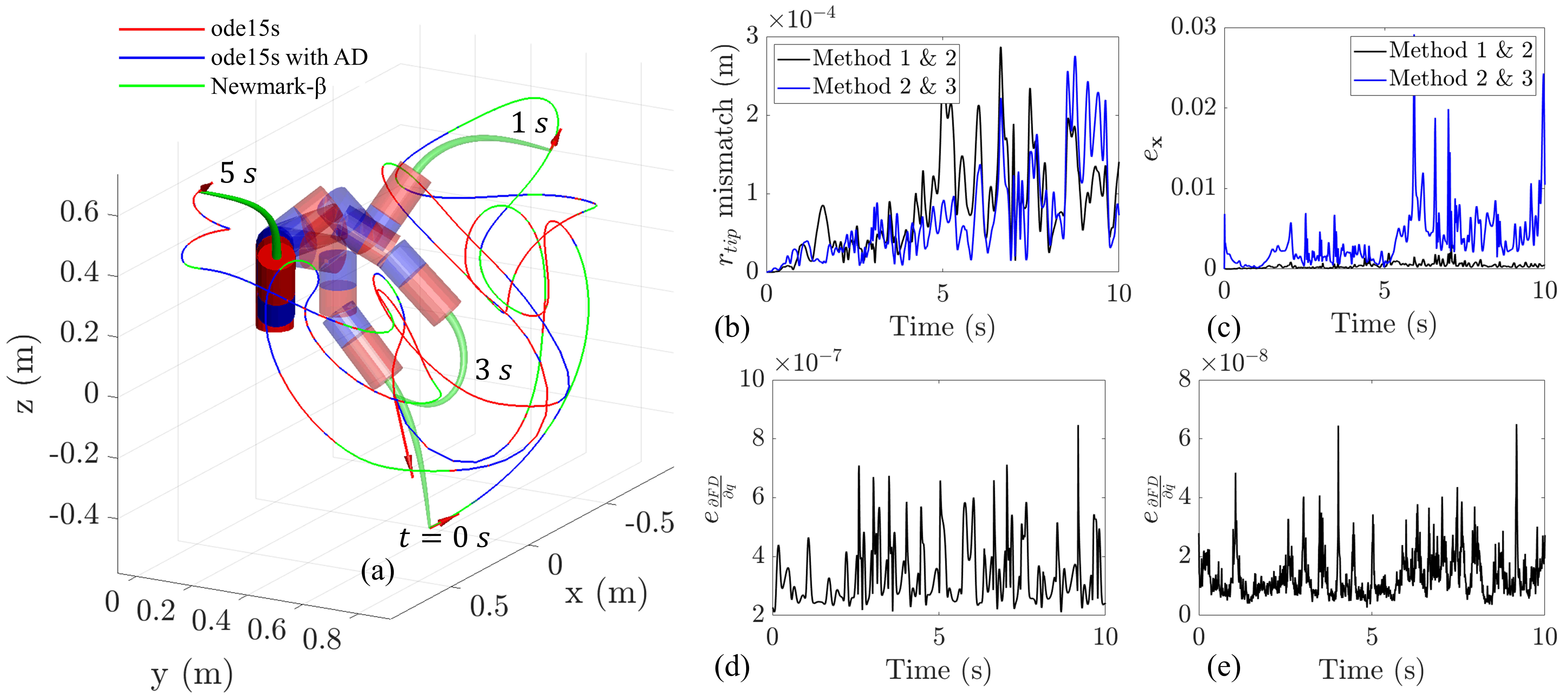}
\caption{Dynamic response of the serial robot: (a) Superimposed snapshots of the robot at different times and the tip trajectory with (blue) and without (red) using analytical derivatives. (b) Tip position mismatch. (c) State mismatch. The mismatch between numerical and analytical derivatives of $FD$ (d) with respect to $\bm{q}$ and (e) with respect to $\dot{\bm{q}}$.}
\label{fig::Exp2_dynamics}
\end{figure*}

The dynamic response of the robot is computed over 10 seconds using three methods: $ode15s$ without providing analytical derivatives, $ode15s$ with analytical derivatives, and the Newmark-$\beta$ method. For the Newmark-$\beta$ approach, a time step of 0.001 s was used, as the solution failed to converge for larger time steps. Snapshots of the simulation results for the first 5 seconds are displayed in Figure \ref{fig::Exp2_dynamics}(a). Without analytical derivatives, the $ode15s$ took 7.08 s, while with analytical derivatives, it was completed in 2.49 s, making it approximately three times faster. While for our custom Newmark-$\beta$ code, the simulation time was 3 min and 51 s. Mismatch metrics, similar to those in the first example, are shown in Figures \ref{fig::Exp2_dynamics}(b), (c), (d), and (e). The position and state mismatch between the first two methods is in the order of 0.1 mm and $10^{-3}$, respectively. The differences between the numerical and analytical derivatives of $FD$ indicate that their discrepancies are insignificant. Numerically computing the Jacobians took an average of 32.1 ms, whereas the analytical Jacobian required only 2.5 ms, making the analytical approach nearly 13 times faster.

\begin{figure}
\centering
\includegraphics[width=1\columnwidth]{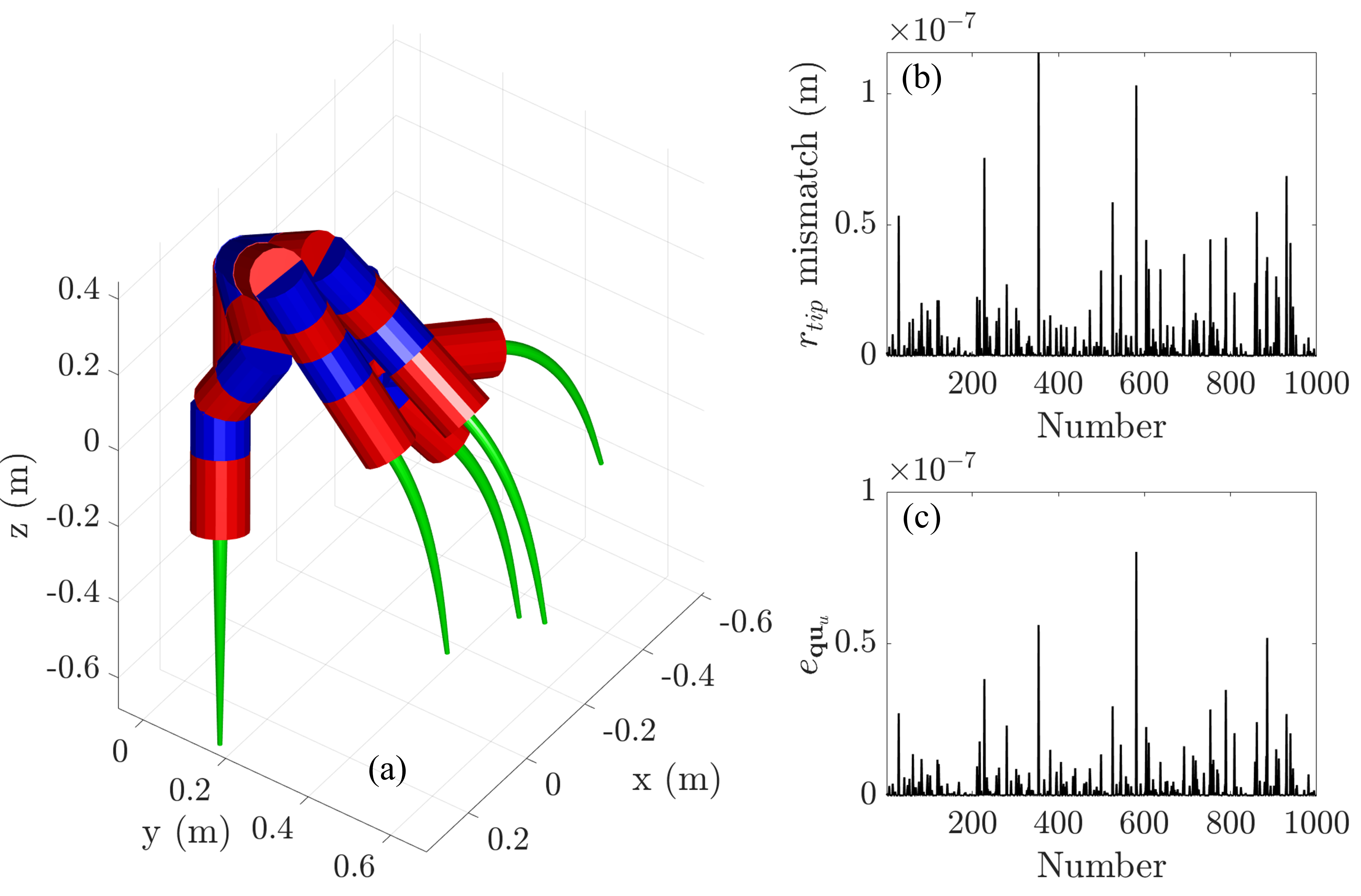}
\caption{Static simulation results: (a) Five arbitrary equilibrium shapes. (b) The mismatch between tip positions. (c) The mismatch static equilibrium solutions ($\bm{q}_u$ and $\bm{u}_u$).}
\label{fig::Exp2_statics}
\end{figure}

Static equilibrium simulation of the robot involves solving $\bm{q}_u$ of the soft body and $\bm{u}_u$ of the rigid joints. Similar to the first example, we performed 1000 static equilibrium simulations on the same system where we used random values of joint angles in the range of $-\pi/4$ to $\pi/4$. Figure \ref{fig::Exp2_statics}(a) illustrates 5 equilibrium shapes from these 1000 simulations. Both methods (with and without the analytical Jacobian) yielded very close results. The mismatch in tip positions and static equilibrium solutions using the two approaches are shown in Figure \ref{fig::Exp2_statics}(b) and (c). Without analytical derivatives, the static simulations took an average of 49.9 ms, whereas, with analytical derivatives, it took 6.3 ms, making the computation approximately eight times faster.

\subsection{Closed Chain Systems}
\label{sec::ClosedChainSystems}

Robots can have a closed-chain (CC) structure, as shown in Figure \ref{fig::multi-body}. Closed-chain systems can be modeled using kinematic constraints ($\bm{e}(\bm{q})=\bm{0}$) expressed in the Pfaffian form (velocity level): $\bm{A}(\bm{q})\dot{\bm{q}}=\bm{0}$. Accordingly, their generalized dynamic equation is given by \cite{Armanini_TRO2022}:
\begin{subequations}
\begin{align}
&\bm{M}\ddot{\bm{q}} = \bm{\tau} + \bm{F} + \bm{A}^T\bm{\lambda} \label{eq::FD_closedchain_a} \\
&\bm{A}\ddot{\bm{q}} + \dot{\bm{A}}\dot{\bm{q}} + \frac{2}{T_B}\bm{A}\dot{\bm{q}} + \frac{1}{T_B^2}\bm{e} = 0 \label{eq::FD_closedchain_b}
\end{align}
\end{subequations}
where $\bm{\lambda} = [\bm{\lambda}_1^T,\;\bm{\lambda}_2^T\;...\;\bm{\lambda}_{n_{CL}}^T]^T$ represents the unknown constraint wrenches, where $n_{CL}$ is the number of closed-chain joints and $T_B$ is the Baumgart stabilization constant. $\bm{A}$, $\dot{\bm{A}}$, and $\bm{e}$ are given by:
\begin{subequations}
\begin{align}
\bm{A}=& \big\Vert_{k=1}^{n_{CL}}\bm{\Phi}_{\perp k}^T(\mathrm{Ad}_{\bm{g}_{kBA}}\bm{J}_{kA}-\bm{J}_{kB})
\label{eq::PfaffianMatrix}\\
\dot{\bm{A}}=& \big\Vert_{k=1}^{n_{CL}}\bm{\Phi}_{\perp k}^T(\mathrm{Ad}_{\bm{g}_{kBA}}\mathrm{ad}_{\bm{\eta}_{kBA}}\bm{J}_{kA}+\mathrm{Ad}_{\bm{g}_{kBA}}\dot{\bm{J}}_{kA}-\dot{\bm{J}}_{kB})
\label{eq::PfaffianMatrix_derivative}\\
\bm{e}=& \big\Vert_{k=1}^{n_{CL}}\bm{\Phi}_{\perp k}^T\left(\text{log}(\bm{g}_{kBA})\right)^\vee
\label{eq::poseerror}
\end{align}
\end{subequations}
where $\bm{g}_{kBA} = \bm{g}_{kB}^{-1}\bm{g}_{kA}$, $\bm{\Phi}_{\perp k}$ is the basis for the constrained degrees of freedom and $\bm{J}_{kA}$ and $\bm{J}_{kB}$ are geometric Jacobians at points $A$ and $B$ where the closed-loop joint is defined. 

Combining \eqref{eq::FD_closedchain_a} and \eqref{eq::FD_closedchain_b} we get:

\begin{equation}
\bm{M}_A\begin{bmatrix}
\ddot{\bm{q}} \\
\bm{\lambda}
\end{bmatrix} = 
\begin{bmatrix}
\bm{\tau}+\bm{F} \\
-\dot{\bm{A}}\dot{\bm{q}} - \frac{2}{T_B}\bm{A}\dot{\bm{q}} - \frac{1}{T_B^2}\bm{e}
\end{bmatrix}
\label{eq::FD_closedchain_combined_matrix}
\end{equation}
where,

\begin{equation}
    \bm{M}_A=\begin{bmatrix}
\bm{M} & -\bm{A}^T \\
\bm{A} & 0
\end{bmatrix}
\end{equation}

Equation \eqref{eq::FD_closedchain_combined_matrix} is used to solve for $\ddot{\bm{q}}$ and $\bm{\lambda}$, and $\bm{\lambda}$ is ignored for the time integration. The partial derivative of \eqref{eq::FD_closedchain_combined_matrix} with respect to $\bm{q}$ gives
\begin{equation}
\begin{bmatrix}
\frac{\partial FD}{\partial \bm{q}} \\
\frac{\partial \bm{\lambda}}{\partial \bm{q}}
\end{bmatrix} = \bm{M}_A^{-1}
\begin{bmatrix}
\frac{\partial \bm{\tau}}{\partial \bm{q}} - \frac{\partial ID(\bm{q},\dot{\bm{q}},\ddot{\bm{q}},\bm{\lambda})}{\partial \bm{q}} \\
-\frac{\partial \bm{A}}{\partial \bm{q}}\ddot{\bm{q}}-\frac{\partial \dot{\bm{A}}}{\partial \bm{q}}\dot{\bm{q}} - \frac{2}{T_B}\frac{\partial \bm{A}}{\partial \bm{q}}\dot{\bm{q}} - \frac{1}{T_B^2}\frac{\partial \bm{e}}{\partial \bm{q}}
\end{bmatrix}
\label{eq::dFD_closedchain_combined_matrix_dq}
\end{equation}
where $ID$ includes the contributions of the constraint force: a local wrench of $\mathrm{Ad}_{\bm{g}_{kBA}}^{-*}\bm{\Phi}_{\perp k}\bm{\lambda}_k$ on point $A$ and $-\bm{\Phi}_{\perp k}\bm{\lambda}_k$ on point $B$ of the closed-loop joint $k$. $ID$ treats $\bm{\lambda}$ as an independent variable.

Similarly, the derivative of \eqref{eq::FD_closedchain_combined_matrix} with respect to $\dot{\bm{q}}$ gives:
\begin{equation}
\begin{bmatrix}
\frac{\partial FD}{\partial \dot{\bm{q}}} \\
\frac{\partial \bm{\lambda}}{\partial \dot{\bm{q}}}
\end{bmatrix} = \bm{M}_A^{-1}
\begin{bmatrix}
\frac{\partial \bm{\tau}}{\partial \dot{\bm{q}}} - \frac{\partial ID(\bm{q},\dot{\bm{q}},\ddot{\bm{q}},\bm{\lambda})}{\partial \dot{\bm{q}}} \\
-\frac{\partial \dot{\bm{A}}}{\partial \dot{\bm{q}}}\dot{\bm{q}}-\dot{\bm{A}} - \frac{2}{T_B}\bm{A}
\end{bmatrix}
\label{eq::dFD_closedchain_combined_matrix_dqd}
\end{equation}

For the index-3 DAE formulation of the problem, we have additional unknowns $\bm{\lambda}$ and additional equations given by the kinematic constraint $\bm{e}(\bm{q})=\bm{0}$. Hence, the residue \eqref{eq::NB_Residue} and Jacobian \eqref{eq::NB_Jacobian} of the Newmark-$\beta$ scheme are modified as follows.

\begin{subequations}
\begin{equation}
\label{eq::NB_Residue_CC}
\bm{R}_{dyn}(\bm{q}_{n+1},\bm{\lambda}_{n+1}) = \begin{bmatrix}
    \bm{\tau}(\bm{q}_{n+1}) - ID(\bm{q}_{n+1},\bm{\lambda}_{n+1})\\
    \bm{e}(\bm{q}_{n+1})
    \end{bmatrix}
\end{equation}
\begin{equation}
\begin{split}
\label{eq::NB_Jacobian_CC}
\bm{J}_{dyn}(\bm{q}_{n+1}) =&  \begin{bmatrix}\frac{\partial \bm{\tau}}{\partial \bm{q}} - \frac{\partial ID}{\partial \bm{q}} & \bm{A}^T\\
\frac{\partial \bm{e}}{\partial \bm{q}} & \bm{0}
\end{bmatrix}
\end{split}
\end{equation}
\end{subequations}

The static equilibrium simulation involves solving $\bm{q}$ and $\bm{\lambda}$ where, $\dot{\bm{q}}$ and $\ddot{\bm{q}}$ are zero. In this case, the residue and Jacobian are identical to \eqref{eq::NB_Residue_CC} and \eqref{eq::NB_Jacobian_CC}.
Analytical derivatives of all quantities in \eqref{eq::dFD_closedchain_combined_matrix_dq}, \eqref{eq::dFD_closedchain_combined_matrix_dqd}, and \eqref{eq::NB_Jacobian_CC} can be found in the Appendix \ref{app::D}.

Figure \ref{fig::FinRayFinger} shows a fin-ray finger, a multi-body system with 17 soft links, revolute joints, a prismatic joint, and 6 closed-chain joints. The finger is actuated by a displacement of the prismatic joint, as shown in Figure \ref{fig::FinRayFinger}(b). The ribs (horizontal links) are connected to the sidewalls via revolute joints. Each link has a rectangular cross-section with a width of 1 mm. The sidewalls have a height of 2.5 cm and a length of 3 cm, while the ribs have a height of 1 cm. All soft links are modeled using linear bending, constant elongation, and shear strains, resulting in a planar multi-body with 74 DoF. The material used is Acrylonitrile-Butadiene-Styrene (ABS), with a Young’s modulus of 2 GPa and a Poisson’s ratio of 0.35, commonly used in 3D printing. Five of the six closed-chain joints are revolute, and the top joint is fixed. The problem includes both the joint coordinate constraint and the closed-chain constraint.

\begin{figure}
\centering
\includegraphics[width=1\columnwidth]{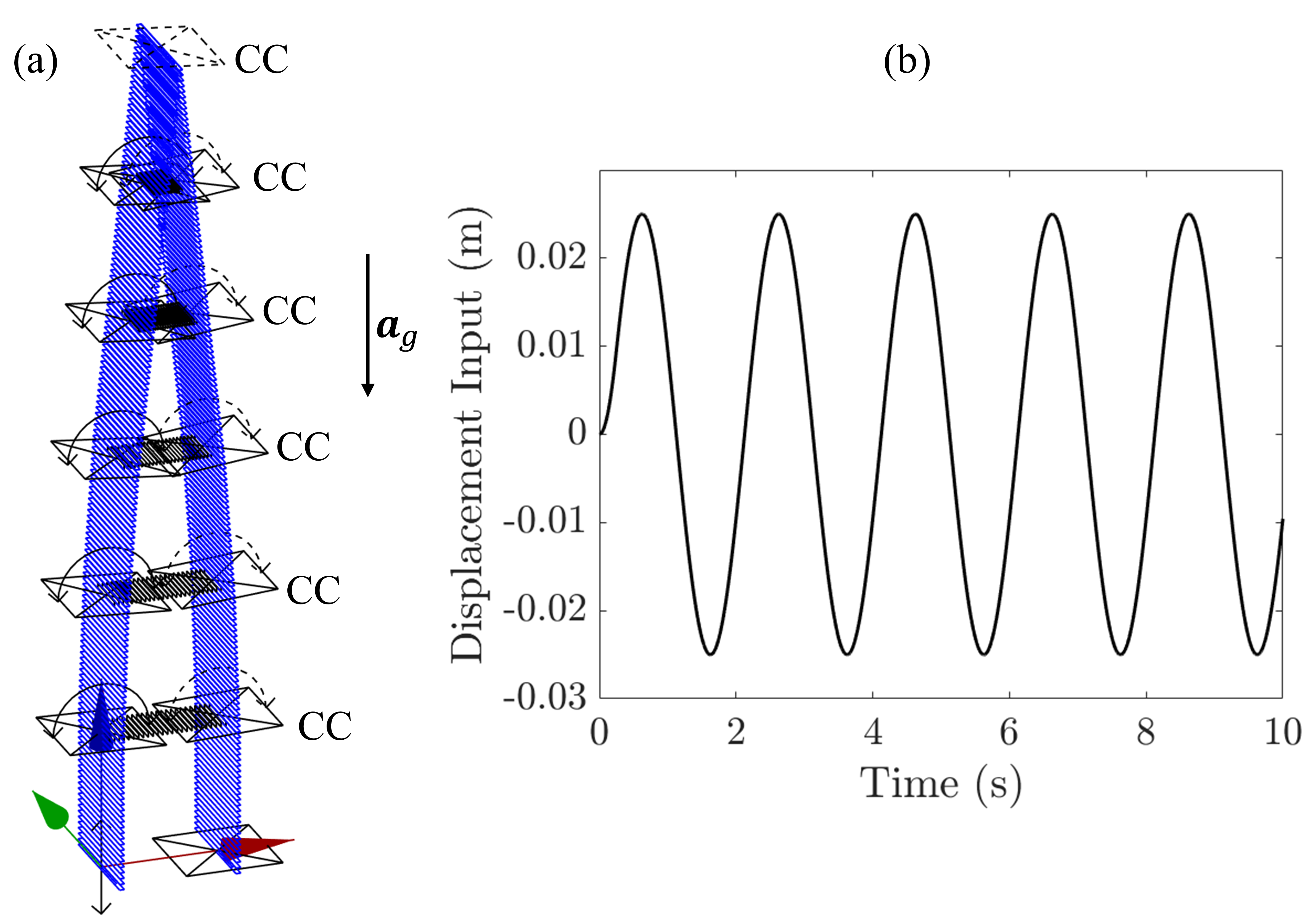}
\caption{(a) A fin-ray finger with 17 soft links and 6 closed-chain joints. The robot is actuated by a prismatic joint at its base. (b) Displacement input of the prismatic joint.}
\label{fig::FinRayFinger}
\end{figure}

\begin{figure*}[ht]
\centering
\includegraphics[width=\textwidth]{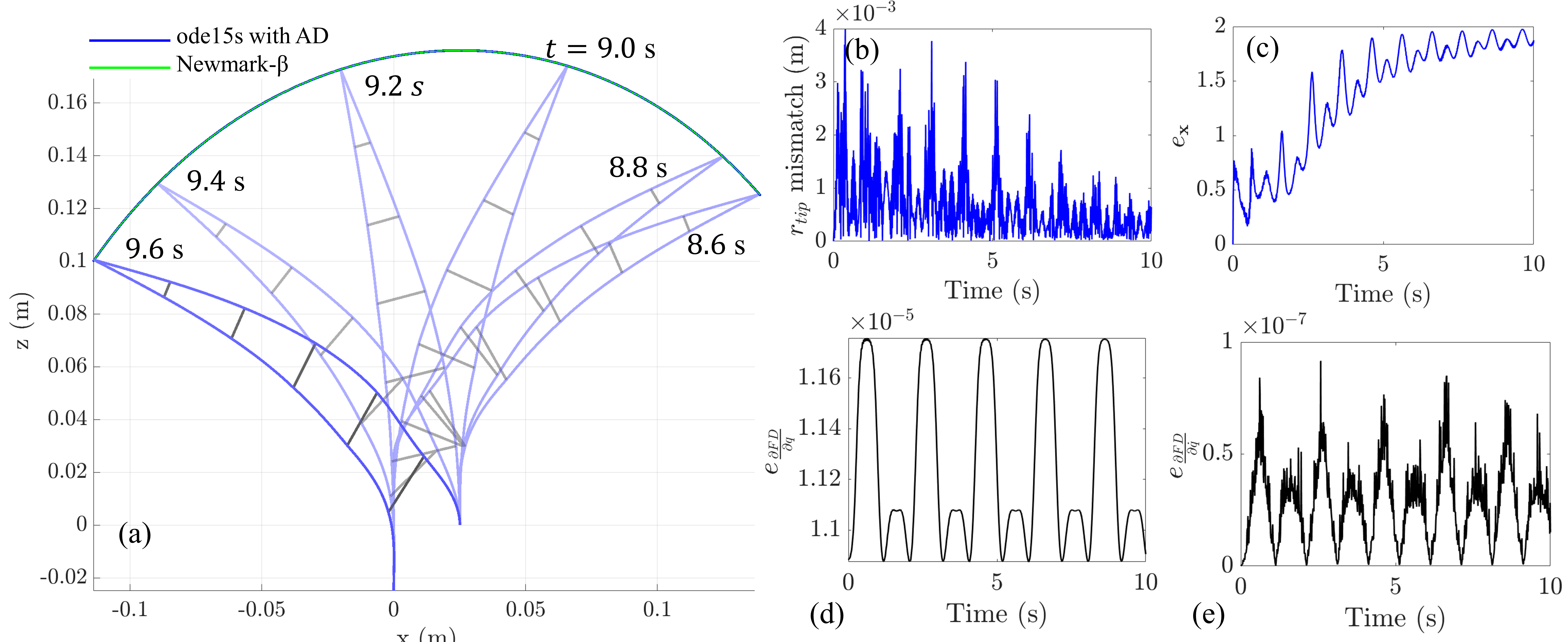}
\caption{Dynamic response of the fin-ray finger: (a) Snapshots of the finger at different times with the trails of the tip position. (b) Average values of mismatch between tip positions. (c) State mismatch. The mismatch between numerical and analytical derivatives of $FD$ (d) with respect to $\bm{q}$ and (e) with respect to $\dot{\bm{q}}$.}
\label{fig::Exp3_dynamics}
\end{figure*}

In this example, when the analytical Jacobian was not provided, the dynamics simulation using $ode15s$ was practically stuck at 0.04 s. Other MATLAB ODE integrators, such as $ode45$ and $ode113$, progressed extremely slowly, completing only $\sim 0.1$ s in 12 hours. However, when $ode15s$ was supplied with the analytical derivatives of $FD$, the simulation was completed in 16 min and 34 s. Meanwhile, the Newmark-$\beta$ integrator (with $h=0.01$ s) completed the same simulation in just 1 min and 9 s, making it the fastest DAE solver for this example. Figure \ref{fig::Exp3_dynamics}(a) illustrates the dynamic response of the finger, which exhibits periodic behavior as expected. Figure \ref{fig::Exp3_dynamics}(b) and (c) compare the tip position and the robot states, respectively. The tip position has a mismatch in the order of mm, while the state mismatch is higher. Using a lower value of time step can reduce this mismatch further. Figure \ref{fig::Exp3_dynamics}(d) and (e) indicates that the discrepancies between the numerical and analytical derivatives of $FD$ are insignificant. The numerical derivatives took an average of 498 ms to compute, while the analytical derivatives took 23.74 ms, making the analytical approach more than 20 times faster.

\section{Examples of Common External Forces}
\label{sec::Common External Forces}
In this section, we derive the analytical derivatives of various external loads to which robots are commonly subjected: point wrenches, contact forces, and hydrodynamic forces.

\subsection{Point Wrench}
\label{sec::PointWrench}
The point wrench can be expressed in the local frame (as a follower load) or the global frame. In the former case, the partial derivative of $\bm{\mathcal{F}}_{k}$ with respect to $\bm{q}$ is $\bm{0}$. In our formulation, an external wrench expressed in the global frame $\bm{\mathcal{F}}_{g k}$ must be transformed into the local frame for the projection into the space of generalized coordinates. The transformed wrench in the local frame is given by:

\begin{equation}
\label{eq::forcelocal}
\bm{\mathcal{F}}_{k} = \mathrm{Ad}_{\bm{g}_{\theta k}}^{-*}\bm{\mathcal{F}}_{gk}
\end{equation}
where, $\bm{g}_{\theta k}$ is the rotational component of $\bm{g}_{k}$.

The derivative of this term with respect to $\bm{q}$ is given by:
\begin{equation}
\label{eq::dforcelocal_dq}
\frac{\partial \bm{\mathcal{F}}_{k}}{\partial \bm{q}} = \mathrm{ad}_{\bm{\mathcal{F}}_{k}}^*\bm{I}_\theta \bm{S}_{k}^B
\end{equation}
where, $\bm{I}_\theta = \text{diag}([1\;1\;1\;0\;0\;0])$.

Accordingly, the partial derivative of $ID$ with respect to $\bm{q}$ includes an additional term at every computational point $k$:
\begin{equation}
\label{eq::ID+dforcelocal_dq}
\frac{\partial ID_k}{\partial \bm{q}} = (\bullet) + \bm{S}_k^T\left(\frac{\partial \bm{\mathcal{F}}_k
}{\partial \bm{q}}^C\right)
\end{equation}
where, $\frac{\partial \bm{\mathcal{F}_k}
}{\partial \bm{q}}^C$ is computed similarly to $\bm{\mathcal{F}}_k^C$:
\begin{equation}
    \frac{\partial \bm{\mathcal{F}}_k
}{\partial \bm{q}}^C =\mathrm{Ad}_{\bm{g}_{k k+1}}^*\left(\frac{\partial \bm{\mathcal{F}}_{k+1}
}{\partial \bm{q}}+\frac{\partial \bm{\mathcal{F}}_{k+1}
}{\partial \bm{q}}^C\right)
\end{equation}

\begin{figure}
\centering
\includegraphics[width=1\columnwidth]{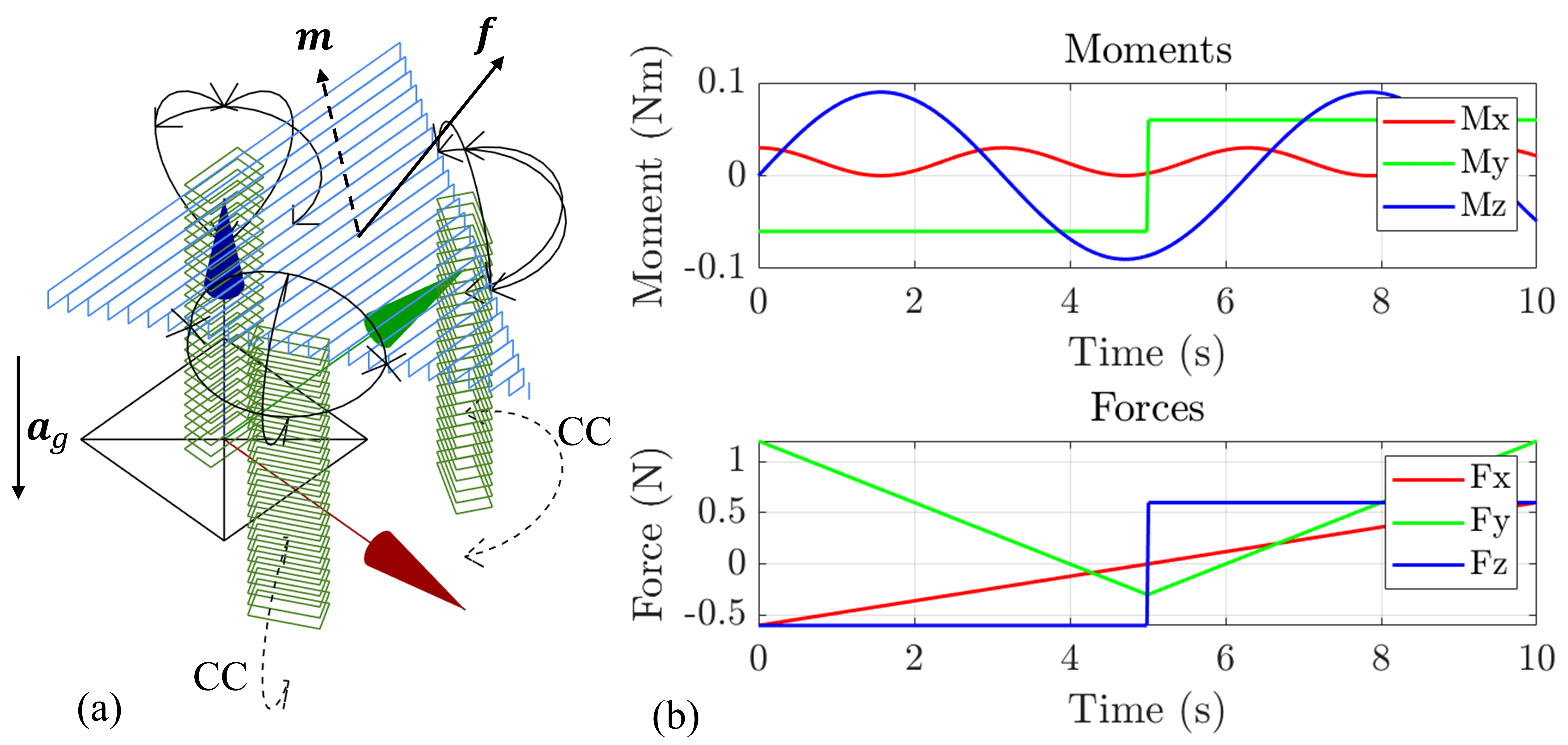}
\caption{(a) Hybrid parallel robot consisting of three soft pillars, a rigid platform with three spherical joints, and two closed-chain revolute joints subject to a point wrench (force $\bm{f}$ and moment $\bm{m}$). (b) Applied external wrenches.}
\label{fig::HybridParallelRobot}
\end{figure}

Figure \ref{fig::HybridParallelRobot}(a) shows an example of a hybrid parallel robot with three soft pillars and a rigid platform. The properties of the soft body are kept identical to those of CDM. The soft body is modeled as a cuboid with dimensions 1.5 cm × 3 cm × 15 cm. The top rigid platform is an equilateral triangle with a side length of 20 cm and a height of 1 cm. The soft pillars are connected to the platform via spherical joints and are parameterized using cubic angular strains and first-order linear strains. Hence, the total DoF of the robot is 63. The dynamic simulation is solved using $FD$ computed from \eqref{eq::FD_closedchain_combined_matrix}, with analytic Jacobians according to \eqref{eq::dFD_closedchain_combined_matrix_dq} and \eqref{eq::dFD_closedchain_combined_matrix_dqd}. The platform is subjected to an external point wrench (force and moment) in the global frame according to Figure \ref{fig::HybridParallelRobot}(b), and the robot's dynamic response is calculated over 10 seconds.


\begin{figure*}[ht]
\centering
\includegraphics[width=\textwidth]{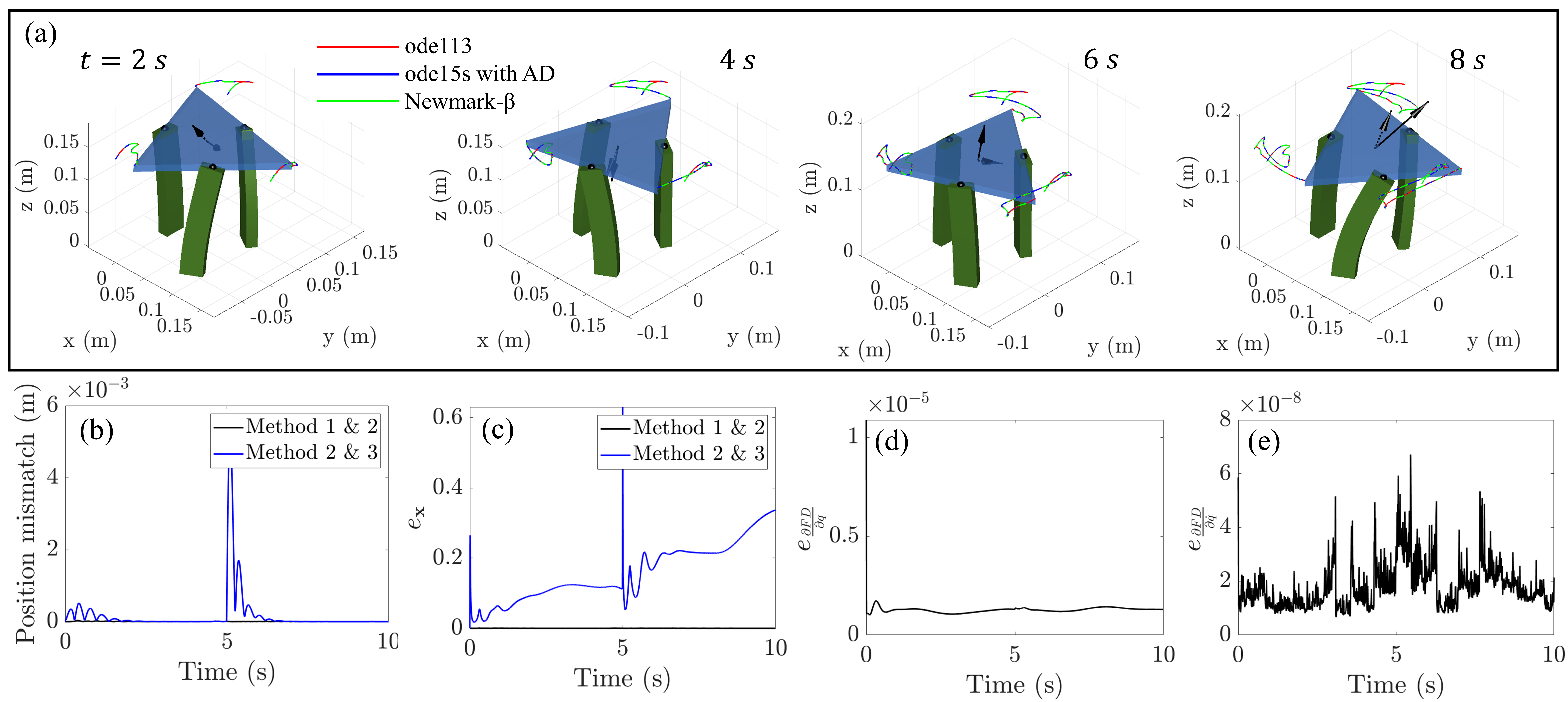}
\caption{Dynamic response of the hybrid parallel robot: (a) Snapshots of the robot at different times with the trails of the corner positions of the rigid platform. Solid and dotted arrows indicate the applied point forces and moments, respectively. (b) Average values of mismatch between corner positions. (c) State mismatch. The mismatch between numerical and analytical derivatives of $FD$ (d) with respect to $\bm{q}$ and (e) with respect to $\dot{\bm{q}}$.}
\label{fig::Exp4_dynamics}
\end{figure*}

Figure \ref{fig::Exp4_dynamics}(a) shows snapshots of the dynamic response. In this example, when the analytical Jacobian was not provided, the simulation using $ode15s$ advanced at an impractically slow pace. Consequently, we switched to MATLAB's $ode45$ and $ode113$ (method 1) integrators. The simulation took 1 hr 46 mins with $ode45$ and 1 hr 15 mins with $ode113$. In contrast, $ode15s$ with the analytical Jacobian completed the simulation in just 2.46 s, making it about 1800 times faster than $ode113$. For the Newmark-$\beta$ scheme, the computational time was 17.62 s for a time step of 0.01 s. Figures \ref{fig::Exp4_dynamics}(b) and (c) demonstrate how closely the dynamic simulation results of all three methods align. The position and state mismatch between the first two methods is in the order of $10^{-5}$ m and $10^{-4}$, respectively. The difference between the analytical and numerical derivatives of $FD$, displayed in Figures \ref{fig::Exp4_dynamics}(d) and (e), validates the analytical derivatives. Numerically computing the derivatives took an average of 82.9 ms, whereas the analytical derivatives required only 5.1 ms, making the analytical approach nearly 16 times faster. Thus, it is not just the speed of computation; its accuracy also contributed to the significant improvement in the overall computational efficiency of dynamic simulations.

\begin{figure}
\centering
\includegraphics[width=1\columnwidth]{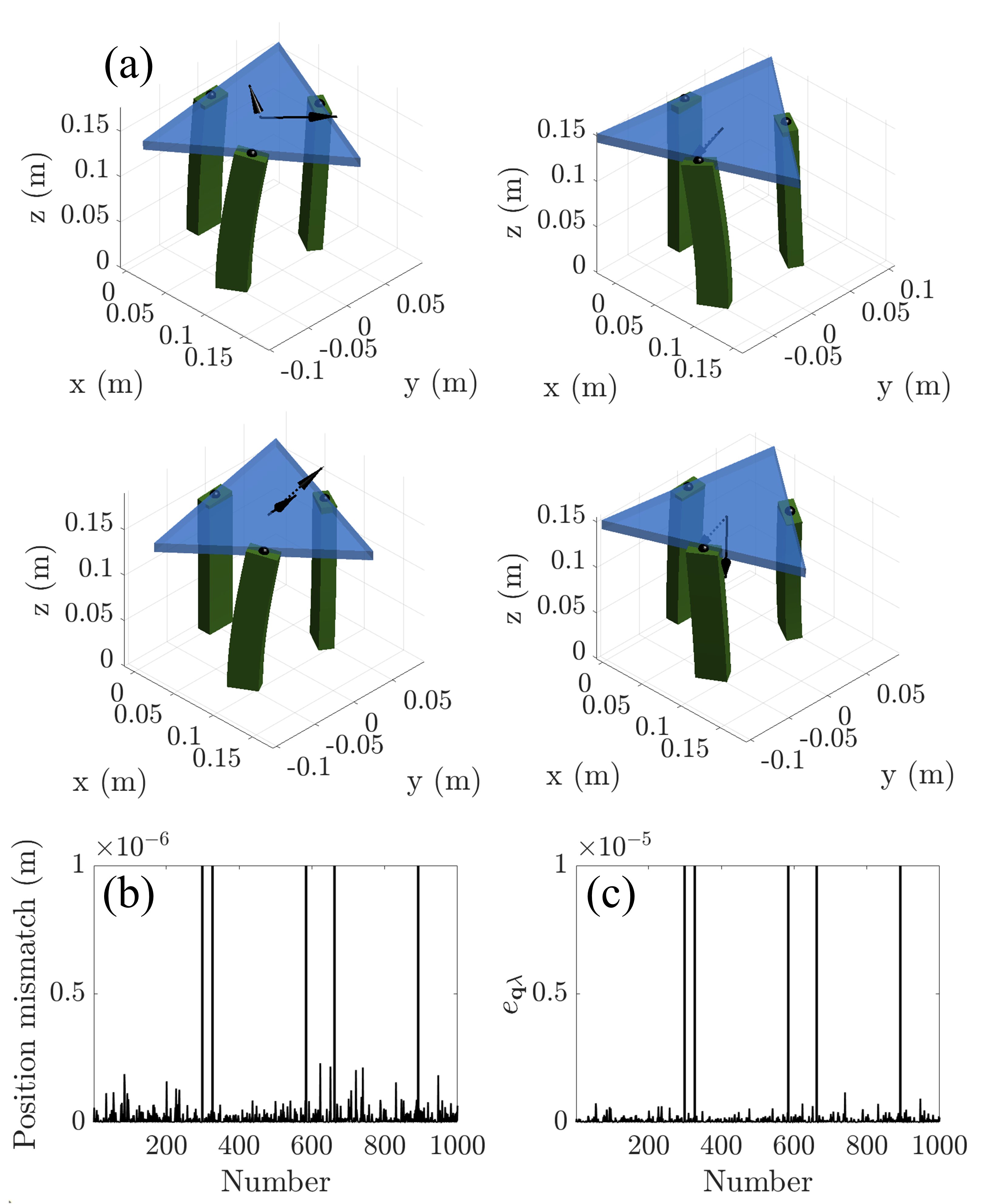}
\caption{Static simulation results of the hybrid parallel robot: (a) Four arbitrary equilibrium shapes. (b) The mismatch between tip positions. (c) The mismatch static equilibrium solutions ($\bm{q}$ and $\bm{\lambda}$). Vertical lines in (b) and (c) indicate cases of multiple static solutions obtained from both methods.}
\label{fig::Exp3_statics}
\end{figure}

For the static equilibrium simulations, the system is subjected to 1000 randomly applied point wrenches: forces in the range -0.5 to 0.5 N and moments in the range -0.05 to 0.05 Nm. The equilibrium values of $\bm{q}$ and $\bm{\lambda}$ are computed by solving \eqref{eq::NB_Residue_CC}. When the analytical derivatives \eqref{eq::NB_Jacobian_CC} were not provided, the simulation took an average of 166.39 ms. In contrast, the simulation was completed in just 8.97 ms with the analytical derivatives, making it approximately 18 times faster. Figure \ref{fig::Exp3_statics}(a) presents four static equilibrium shapes from the simulation. The average corner position mismatch is depicted in Figure \ref{fig::Exp3_statics}(b), while the solution mismatch, including the equilibrium values of $\bm{q}$ and $\bm{\lambda}$, is shown in Figure \ref{fig::Exp3_statics}(c). These figures suggest that, with and without analytical derivatives, the solutions obtained are identical, except in cases where multiple solutions are present.

\subsection{Contact Force}
\label{sec::Contact}
Accurate modeling of contact mechanics is paramount in robotics and has been addressed through various models and algorithms \cite{Lidec2024}. Here, we examine a simplified internal contact scenario where a cable-driven soft manipulator, such as the one depicted in Figure \ref{fig::CDM}, is actuated inside a hollow cylinder, with the cylinder's centerline aligned to the global z-axis. The following assumptions are made to simplify the contact force:

\begin{enumerate}
    \item To compute the contact force, the manipulator is discretized into spheres at each computational point ($n_p=12$ in this example). The radius of each sphere corresponds to the manipulator's radius at that point $k$.
    \item Only normal forces are considered; tangential (frictional) forces are assumed to be zero. Hence, the contact force does not cause any local moments.
    \item Penetration into the cylinder is allowed, and the contact force is assumed to be a function of the penetration $\delta$.
    \item The direction of the contact force is normal to the surface of the cylindrical wall, taking the form $[n_x,\; n_y,\; 0]^T$.
\end{enumerate}

To model the contact force, we used the Hertz model \cite{Flores2022}, a two-parameter non-linear contact model given by:

\begin{equation}
\label{eq::contactmodel}
\bm{\mathcal{F}}_{gk} = 
\begin{cases} 
\begin{pmatrix} 
\bm{0} \\ 
k_c\delta_k^p\bm{u}_{\perp k} 
\end{pmatrix} & \text{if } \delta > 0 \\ 
\bm{0} & \text{otherwise}
\end{cases}
\end{equation}
where $k_c$ represents the contact stiffness coefficient and $p$ is the force exponent, typically dependent on the material properties and geometry of the contact. $\bm{u}_{\perp}$ is the unit normal from the center of the sphere to the wall of the cylinder. The penetration $\delta_k$ is given by:
\begin{equation}
\label{eq::penetration}
\delta_k = \|\bm{n}_{\perp k}\|+r_k-r_{cyl}
\end{equation}
where $\bm{n}_{\perp k} = \bm{C}_1\bm{r}_{g k}$, $\bm{C}_1 = \text{diag}([1\;1\;0])$ and $\bm{r}_{g k}$ is the vector from the global frame to point $k$. Note that $\bm{u}_{\perp k}=\bm{n}_{\perp k}/\|\bm{n}_{\perp k}\|$. Since $\bm{\mathcal{F}}_{gk}$ is in the global frame, we need to transform it into the local frame according to \eqref{eq::forcelocal}. The derivative of the local force with respect to $\bm{q}$ is given by:
\begin{equation}
\label{eq::dcontactforcelocal_dq}
\frac{\partial \bm{\mathcal{F}}_{k}}{\partial \bm{q}} = \mathrm{ad}_{\bm{\mathcal{F}}_{k}}\bm{I}_\theta \bm{S}_{k}^B+\mathrm{Ad}_{\bm{g}_{\theta k}}^{-1}\frac{\partial \bm{\mathcal{F}}_{gk}}{\partial \bm{q}}
\end{equation}

The partial derivative of the contact force in the global frame is given by:

\begin{equation}
\label{eq::dcontactforce_dq}
\frac{\partial \bm{\mathcal{F}}_{gk}}{\partial \bm{q}} = \begin{pmatrix} 
\bm{0} & \bm{0}\\ 
\bm{0} & \bm{k}^*\bm{C}_1
\end{pmatrix}
\mathrm{Ad}_{\bm{g}_{\theta k}}\bm{S}_{k}^B
\end{equation}
where $\bm{k}^*$ is given by
\begin{equation}
\label{eq::kstar}
\bm{k}^*=k_c p\delta_k^{p-1}\bm{u}_{\perp k}\bm{u}_{\perp k}^T+\frac{k_c\delta_k^{p}}{\|\bm{n}_{\perp k}\|}(\bm{I}_3-\bm{u}_{\perp k}\bm{u}_{\perp k}^T)
\end{equation}

Similar to the case of point force, the derivative of the local force \eqref{eq::dcontactforcelocal_dq} is projected backward, and their contribution is accounted into the partial derivative of $ID$. Note that the penetration condition of the contact force \eqref{eq::contactmodel} also applies to its derivative. Readers interested in detailed derivation may refer to Appendix \ref{app::D}.

\begin{figure*}[ht]
\centering
\includegraphics[width=\textwidth]{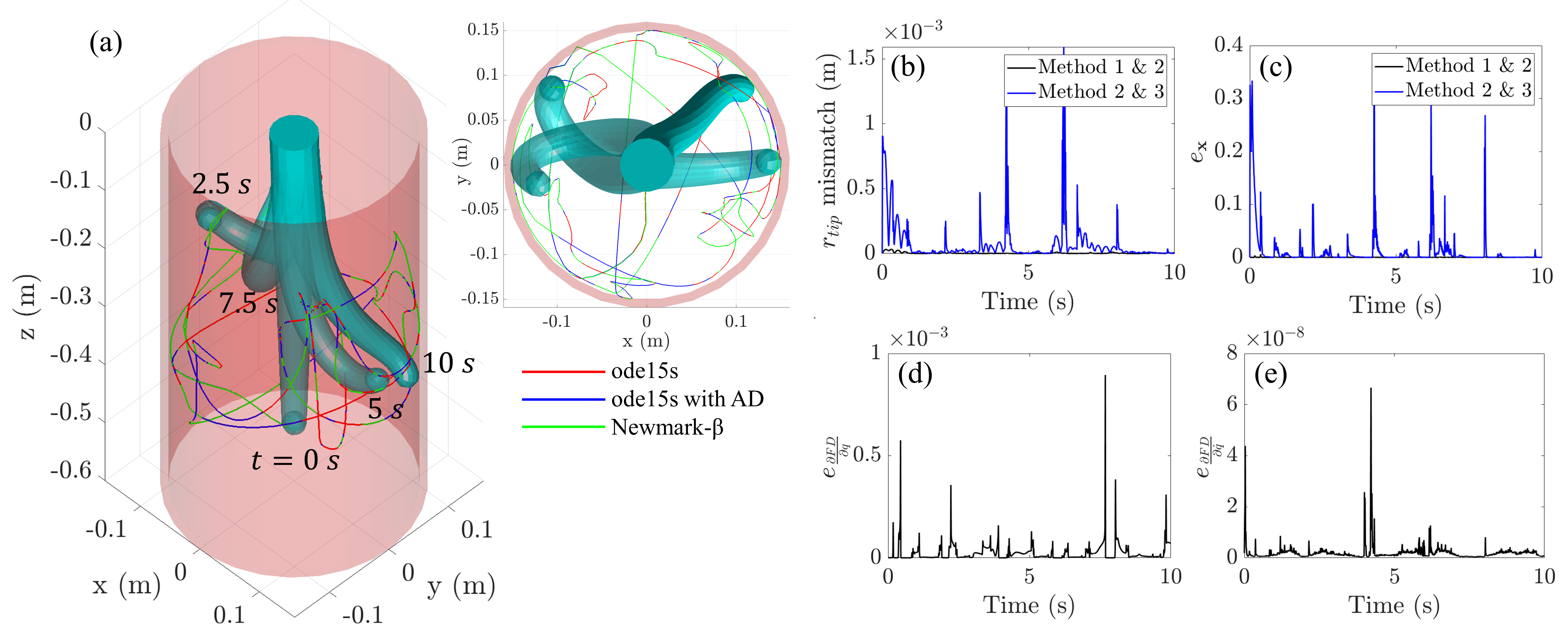}
\caption{Contact simulation results: (a) Snapshots of dynamics of CDM inside a hollow cylinder with tip trajectories. Inset showing a top view. (b) Mismatch between tip positions. (c) State mismatch. Mismatch between numerical and analytical derivatives of $FD$: (d) with respect to $\bm{q}$ and (e) with respect to $\dot{\bm{q}}$.}
\label{fig::Exp5_dynamics}
\end{figure*}

For the dynamic simulation, we used the same manipulator and actuation inputs as shown in Figure \ref{fig::CDM}. We used an inner wall radius of $r_{cyl} = 15 \; \text{cm}$, $k_c = 10^5 \; \text{N/m}$, and $p = 1.5$. The simulation results are displayed in Figure \ref{fig::Exp5_dynamics}(a). Simulation using $ode15s$ without providing analytical derivatives was completed in 30.61 s, while with analytical Jacobian, it was completed in 7.80 s, making it nearly 4 times faster. Integration using the Newmark-$\beta$ approach with a time step of 0.002 s was completed in 46.64 s. The validation between the methods in terms of tip position and robot state are shown in Figure \ref{fig::Exp5_dynamics}(b) and (c), respectively. Between methods 1 and 2, the tip position mismatch is less than $40 \; \mu$m and the state mismatch is in the order of $10^{-3}$. Figure \ref{fig::Exp5_dynamics}(d) and (e) compare the derivatives of $FD$ obtained using numerical and analytical methods. While the discrepancies are generally small, the relatively larger mismatches in the partial derivatives of $FD$ with respect to $\bm{q}$ can be attributed to the penetration condition. The average time to compute the numerical derivatives was 23.7  ms, while for the analytical derivatives, it was 2.3 ms, making the analytical method approximately 10 times faster.

\subsection{Hydrodynamic Force}
\label{sec::Hydrodynamic}
Several hybrid soft robots are designed for underwater exploration. In addition to gravity, they are subject to external forces due to buoyancy, drag, lift, and added mass (fluid displacement). Buoyancy acts as an acceleration opposing gravity, while the added mass force can be simplified as additional inertia that the body experiences. To model the effect of buoyancy, we can use the modified acceleration $\bm{a}_G' = (1 - \rho_w/\rho_b) \bm{a}_G$, where $\rho_w$ is the density of the surrounding water and $\rho_b$ is the density of the robot's body. To tackle the hydrodynamic forces along rods, we follow \cite{Boyer_2006} and use a model that combines a simplified version of the reactive theory of \cite{Lighthill_LAEBT}, with the resistive empirical model of \cite{Morison_1950}. In this model, the rod inertia matrices are replaced by the modified inertia matrix $\bm{\mathcal{M}}_k' = \bm{\mathcal{M}}_k + \bm{\mathcal{M}}_{Ak}$, where $\bm{\mathcal{M}}_{Ak}$ represents the added inertia matrix on the $k$-th computational point on the robot. The effects of viscosity are modeled by a field of drag-lift forces applied along the ﬂagellum, and given by \cite{Armanini_TRO2021}:
\begin{equation}
\label{eq::DragLift}
\bm{\mathcal{F}}_{Dk} = \bm{\mathcal{D}}_k  \|\bm{v}_k\|  \bm{\eta}_k
\end{equation}
where, $\bm{\mathcal{D}}_k$ is the drag-lift matrix and $\|\bm{v}_k\|= \sqrt{\bm{\eta}_k^T\bm{I}_v\bm{\eta}_k}$, where $\bm{I}_{v}=\text{diag}([0\;0\;0\;1\;1\;1])$. Hence, we have:
\begin{equation}
\label{eq::dDragLift_dq}
\frac{\partial \bm{\mathcal{F}}_{Dk}}{\partial \bm{q}} = \bm{\mathcal{D}}_k^*\frac{\partial \bm{\eta}_k}{\partial \bm{q}}
\end{equation}
where $\bm{\mathcal{D}}_k^* = \bm{\mathcal{D}}_k\left(\frac{1}{\|\bm{v}_k\|}\bm{\eta}_k \bm{\eta}_k^T\bm{I}_{v} + \|\bm{v}_k\|  \bm{I}_6 \right)$.

Substituting the partial derivative of the velocity twist (Appendix \ref{app::D}), we get the derivative of the drag-lift force:

\begin{equation}
\label{eq::dDragLift_dq_end}
\frac{\partial \bm{\mathcal{F}}_{Dk}}{\partial \bm{q}} = \bm{\mathcal{D}}_k^*\sum_{\beta<k}\mathrm{Ad}_{\bm{g}_{\beta k}}^{-1}\bm{R}_\beta
\end{equation}

Similarly, we can write the partial derivative of the drag-lift force with respect to $\dot{\bm{q}}$ as follows,

\begin{equation}
\label{eq::dDraglift_dqd}
\frac{\partial \bm{\mathcal{F}}_{Dk}}{\partial \dot{\bm{q}}} = \bm{\mathcal{D}}_k^*\sum_{\beta<k}\mathrm{Ad}_{\bm{g}_{\beta k}}^{-1}\bm{S}_\beta
\end{equation}

With the form of \eqref{eq::dDragLift_dq_end} and \eqref{eq::dDraglift_dqd}, it is easy to see that the contribution of drag-lift force can be accounted for by including an additional term in \eqref{eq::N_k}:

\begin{equation}
    \bm{\mathcal{N}}_k = \overline{\mathrm{ad}}_{\bm{\mathcal{M}}_k\bm{\eta}_k}^*+\mathrm{ad}_{\bm{\eta}_k}^*\bm{\mathcal{M}}_k -\bm{\mathcal{M}}_k\mathrm{ad}_{\bm{\eta}_k}+\bm{\mathcal{D}}_k^* \label{eq::N_k_draglift} 
\end{equation}

\begin{figure}
\centering
\includegraphics[width=1\columnwidth]{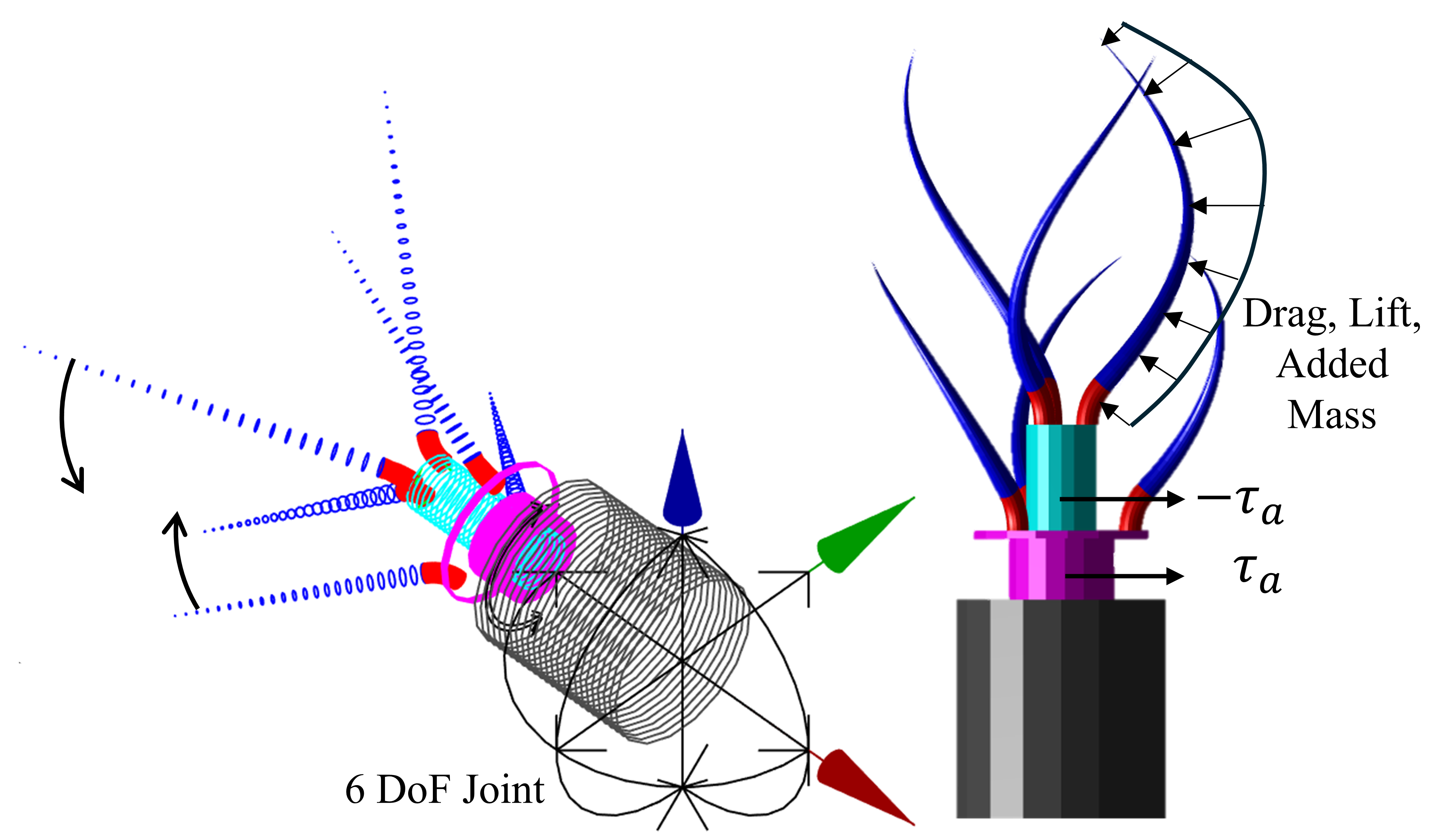}
\caption{Schematic of the underwater hybrid mobile robot. The robot is actuated by joint torque and is subject to drag-lift and added mass forces.}
\label{fig::UWMobileRobot}
\end{figure}

Using this, we simulated the dynamics of an underwater (UW) flagellated vehicle. The schematic of the robot is shown in Figure \ref{fig::UWMobileRobot}. The robot is a hybrid-branched chain system with a mobile (6 DoF) body, two shafts (rigid bodies with revolute joints), six soft bodies (flagella), and six hooks that connect the flagella with the shafts. The three front flagella are 35 cm long, while the rear ones are 50 cm long. Each flagellum has a base radius of 12.5 cm, tapering smoothly to a point at the tips. The material properties of the soft body are kept identical to previous examples. The robot is actuated by joint torques while subject to external hydrodynamic forces. In this example, we assume that all the bodies are neutrally buoyant ($\rho_w=\rho_b$). Each flagellum is modeled as an inextensible Kirchhoff rod with a cubic strain field, resulting in 80 DoFs for the robot.

\begin{figure*}[ht]
\centering
\includegraphics[width=\textwidth]{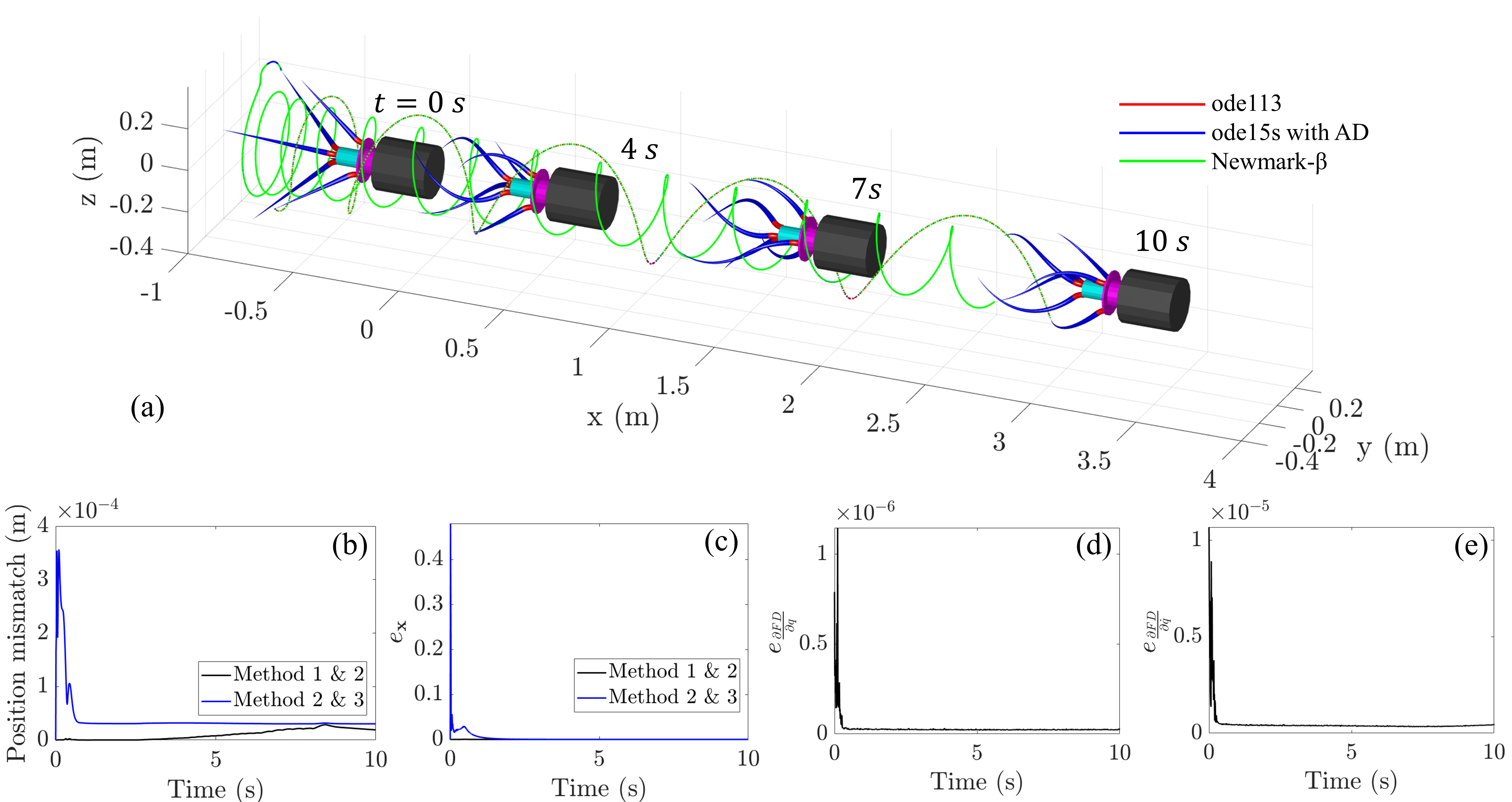}
\caption{Dynamic response of the underwater mobile robot: (a) Snapshots of the robot at different times with the trajectory of two flagella. (b) Average values of mismatch between the tips of all six flagella. (c) State mismatch. The mismatch between numerical and analytical derivatives of $FD$ (d) with respect to $\bm{q}$ and (e) with respect to $\dot{\bm{q}}$.}
\label{fig::Exp6_dynamics}
\end{figure*}

We applied a joint torque of 0.25 Nm on the first shaft and -0.25 Nm on the second to cancel out the rotation of the robot body. The dynamics of the robot are computed for 10 s. Similar to the case of the hybrid parallel robot, the simulation using $ode15s$ without providing the Jacobian progressed very slowly. As a result, we used $ode45$ and $ode113$ (method 1), obtaining simulation times of 11 min 58 s and 8 min 16 s, respectively. In contrast, ode15s with the analytical Jacobian completed the simulation in just 1.17 s, making it more than 400 times faster. Using the Newmark-$\beta$ method, the simulation was finished in 60.91s. The dynamic response of the robot is displayed in Figure \ref{fig::Exp6_dynamics}(a). The mismatch metrics comparing the outputs of all three methods are provided in Figure \ref{fig::Exp6_dynamics}(b) and (c). Between methods 1 and 2, the state mismatch is in the order of $10^{-4}$. The validation of the analytical derivatives of $FD$ is provided in Figure \ref{fig::Exp6_dynamics}(d) and (e). The numerical method took 216.4 ms on average, while the average time for the analytical approach was 11.9 ms, making it nearly 18 times faster.

\section{Discussions and Conclusions}
\label{sec::Conclusion}

In this work, we used the RNEA algorithm to derive the analytical derivatives of the GVS model for the mechanical analysis of hybrid soft-rigid robots. The ``pseudo rigid joint" formulation of the GVS model, facilitated by the Magnus expansion of soft rods, allows for the extension and generalization  analytical derivatives of rigid body dynamics algorithms \cite{carpentier2018analytical, Singh2022}. These derivatives were applied to the dynamic and static simulations of various hybrid multi-body systems. We presented six dynamic and three static simulations, demonstrating significant computational improvements. For dynamic simulations, three integration methods were used: method 1 employed either $ode15s$ or $ode113$ (when $ode15s$ was slower or stalled), method 2 utilized $ode15s$ with analytical derivatives, and method 3 applied the Newmark-$\beta$ approach. Method 2 provided exceptional speedup, particularly in multi-body systems with constraints, where speedup factors of over 3 orders of magnitude ($>10^3$) were observed. The Newmark-$\beta$ approach was observed to be faster for the highly constrained fin-ray finger simulation, while the solution using method 1 was impractically slow for the same case. The results for dynamic simulations are summarized in Table \ref{tab::summary}. For static simulations, $fsolve$ with analytical Jacobian was at least 8 times faster in all cases due to the improved convergence achieved with the accurate derivative information and efficient recursive implementation.

\begin{table*}[htbp]
\centering
\caption{Summary of computational times for 10-second dynamic simulations.}
\label{tab::summary}
\resizebox{\textwidth}{!}{
\begin{tabular}{|c|c|c|c|c|c|c|c|c|c|}
\hline
\multirow{2}{*}{Example} & \multirow{2}{*}{$N$} & \multirow{2}{*}{DoF} & \multirow{2}{*}{$n_p$} & \multicolumn{2}{c|}{Method 1} & \multicolumn{2}{c|}{$ode15s$ with AD (Method 2)} & \multicolumn{2}{c|}{Neumark-$\beta$ (Method 3)} \\
\cline{5-10}
 & & & & Computational Time & Solver & Computational Time & Speed-up Factor & $h$ & Computational Time \\
\hline
CDM & 1 & 24 & 7 & 5.4 s & $ode15s$ & 1.25 s & 4.32 & 0.002 s & 24.43 s \\
\hline
Serial Robot & 8 & 27 & 22 & 7.08 s & $ode15s$ & 2.49 s & 2.84 & 0.001 s & 3 min 51 s \\
\hline
Fin-ray Finger & 23 & 74 & 136 & Very slow & N/A & 16 min 34 s & Very high & 0.01 s & 1 min 9 s \\
\hline
Parallel Robot & 6 & 63 & 26 & 1 hr 15 min & $ode113$ & 2.46 s & 1829.27 & 0.01 s & 17.62 s \\
\hline
Contact Scenario & 1 & 24 & 12 & 30.61 s & $ode15s$ & 7.8 s & 3.92 & 0.002 s & 46.64 s \\
\hline
UW Vehicle & 19 & 80 & 80 & 8 min 16 s & $ode113$ & 1.17 s & 424.27 & 0.01 s & 1 min 9 s \\
\hline
\end{tabular}}
\end{table*}

A limitation of the $ode15s$ solver in MATLAB is that it does not allow passing $\dot{\bm{x}}$ to the Jacobian function. Since the partial derivative of $FD$ \eqref{eq::dFD_dq} requires $\ddot{\bm{q}}$, we need to compute $\ddot{\bm{q}}$ twice, slowing down the simulation. Adding the capability to directly pass $\dot{\bm{x}}$ could improve simulation speed. Even though the Newmark-$\beta$ approach is not a built-in MATLAB function, and our implementation is currently far from optimized, it offers several advantages. Its high stability allows large time steps without sacrificing accuracy, especially in handling DAE index-3 problems, as they are commonly considered in nonlinear structural dynamics of hybrid soft-rigid systems, where this formulation is generally preferred to the index-1 formulation of the Baumgart algorithm. In addition, the Newmark-$\beta$ scheme works particularly well for highly constrained systems. Moreover, preserving the symplectic structure of mechanical systems while maintaining simulation stability and second-order accuracy, it is ideal for long-duration simulations. Here, we used a fixed time step that resulted in a positional mismatch at the millimeter level compared to method 2, however, developing a variable time-step version of the method could further enhance its efficiency and performance. Similarly, replacing the nominal version of the Newmark scheme with the generalized HHT or $\alpha$ schemes should improve its performance, while requiring only minor modifications to the scheme. 


We have implemented the GVS model in a GUI-based MATLAB toolbox called SoRoSim \cite{Mathew2022}. The toolbox allows users from various backgrounds to easily simulate hybrid multi-body systems without requiring deep expertise in underlying algorithms. Implementation of analytical derivatives for the analysis of arbitrary robotic systems is a complex task, particularly for new researchers. To further simplify this process, we plan to upgrade the SoRoSim into a fully differentiable simulator, making it an even more accessible and powerful tool for robotics research.

The applications of analytical derivatives extend far beyond the algorithms presented for static and dynamic analysis. They are foundational for a wide range of advanced control and optimization techniques. For example, in design optimization and trajectory optimization, analytical derivatives enable efficient fine-tuning of system parameters and motion planning. They are also crucial for Model Predictive Control (MPC), where real-time system predictions are needed to maintain optimal performance. Moreover, their use can significantly improve methods like Differential Dynamic Programming (DDP), especially in handling non-linear constraints and complex robotic configurations. 
Analytical derivatives can make several rigid-body algorithms accessible for soft robotic systems, allowing researchers to simulate and control complex hybrid systems efficiently. The techniques presented in this paper offer broad applicability across the robotics field for enhancing the efficiency and precision of simulations. We believe that this work will play a pivotal role in addressing challenges in soft and hybrid soft rigid robots, driving innovation, and enabling real-world applications.

\section*{Acknowledgments}
By the Khalifa University of Science and Technology under Grants RIG-2023-048 and RC1-2018-KUCARS, as well as the French ANR COSSEROOTS (ANR-20-CE33-0001).

\appendix
\addcontentsline{toc}{section}{Appendices}
\section*{Appendices}
\section{Basic SE(3) Formulae}
\label{app::A}

Adjoint operator of $\mathfrak{se}(3)$:
\small
\begin{equation}
\begin{split}
\nonumber
\mathrm{ad}_{\mathcal{V}} = \left( \begin{array}{cc} \widetilde{\bm{\mathsf{w}}} & \bm{0}_{3\times3} \\
\widetilde{\bm{\mathsf{v}}} & \widetilde{\bm{\mathsf{w}}} \end{array} \right) \in \mathbb{R}^{6\times6}
\end{split}
\end{equation}
\normalsize
where, $\bm{\mathcal{V}} = [\bm{\mathsf{w}}^T \bm{\mathsf{v}}^T]^T \in \mathbb{R}^{6}$ is a screw vector.

Coadjoint operator of $\mathfrak{se}(3)$:
\small
\begin{equation}
\begin{split}
\nonumber
\mathrm{ad}_{\mathcal{V}}^* = \left( \begin{array}{cc} \widetilde{\bm{\mathsf{w}}} & \widetilde{\bm{\mathsf{v}}} \\ 
\bm{0}_{3\times3} & \widetilde{\bm{\mathsf{w}}} \end{array} \right) 
\in \mathbb{R}^{6\times6}
\end{split}
\end{equation}
\normalsize

Coadjointbar operator of $\mathfrak{se}(3)$:
\small
\begin{equation}
\nonumber
\overline{\mathrm{ad}}_{\bm{\mathcal{V}}}^* = -\left( \begin{array}{cc} \widetilde{\bm{\mathsf{w}}}& \widetilde{\bm{\mathsf{v}}} \\ \widetilde{\bm{\mathsf{v}}}& \bm{0}_{3\times3}\end{array} \right) \in \mathbb{R}^{6\times6} 
\end{equation}
\normalsize

Adjoint map of $SE(3)$:
\small
\begin{equation}
\begin{split}
\nonumber
\mathrm{Ad}_{\bm{g}(X)} = \left( \begin{array}{cc} \bm{R} & \bm{0}_{3\times3} \\ \widetilde{\bm{\mathsf{r}}}\bm{R} & \bm{R} \end{array} \right) \in \mathbb{R}^{6\times6}
\end{split}
\end{equation}
\normalsize

CoAdjoint map of $SE(3)$:
\small
\begin{equation}
\begin{split}
\nonumber
\mathrm{Ad}_{\bm{g}(X)}^* = \left( \begin{array}{cc} \bm{R} & \widetilde{\bm{\mathsf{r}}}\bm{R} \\ \bm{0}_{3\times3} & \bm{R} \end{array} \right) \in \mathbb{R}^{6\times6}
\end{split}
\end{equation}
\normalsize

Exponential map of $SE(3)$ (replace $\bm{\Omega}$ with $\bm{\xi}$ for rigid joints):

\begin{equation}
\begin{split}
\nonumber
\exp\left({\widehat{\bm{\Omega}}}\right) =& \bm{I}_4 + \widehat{\bm{\Omega}} + \frac{1}{\theta^2}\left(1-\cos\left(\theta\right)\right)\widehat{\bm{\Omega}}^2 \\
&+ \frac{1}{\theta^3}\left(\theta-\sin\left(\theta\right)\right)\widehat{\bm{\Omega}}^3
\end{split}
\end{equation}
where, $\theta = \sqrt{\Omega^T I_{\theta}\Omega}$, $\bm{I}_{\theta}=\text{diag}([1\;1\;1\;0\;0\;0])$

Tangent operator and its derivative

\small
\begin{equation}
\begin{split}
\nonumber
\bm{T}(\bm{\Omega})=\bm{I}_6 + \sum_{r=1}^{4}f_r(\theta)\mathrm{ad}_{\bm{\Omega}}^r
\end{split}
\end{equation}
\begin{equation}
\begin{split}
\nonumber
\dot{\bm{T}}(\bm{\Omega},\dot{\bm{\Omega}})=\sum_{r=1}^{4} \left(f_r^{'}(\theta)\dot{\theta}\mathrm{ad}_{\bm{\Omega}}^r+f_r(\theta)\dot{\left(\mathrm{ad}^r_{\bm{\Omega}}\right)}\right)
\end{split}
\end{equation}
\normalsize
where, $\dot{\theta} = \frac{1}{\theta}\Omega^T I_{\theta}\dot{\Omega}$ and $\dot{\left(\mathrm{ad}^r_{\bm{\Omega}}\right)} = \sum_{u=1}^{r}\mathrm{ad}_{\bm{\Omega}}^{r-u}\mathrm{ad}_{\dot{\bm{\Omega}}}\mathrm{ad}_{\bm{\Omega}}^{u-1}$

\small
\begin{subequations}
\begin{align}
f_1(\theta) =& \frac{1}{2\theta^2}\left(4-4\cos\left(\theta\right)-\theta\sin\left(\theta\right)\right) \nonumber \\
f_2(\theta) =& \frac{1}{2\theta^3} \left(4\theta-5\sin\left(\theta\right)+\theta\cos\left(\theta\right)\right)  \nonumber\\
f_3(\theta) =& \frac{1}{2\theta^4} \left(2-2\cos\left(\theta\right)-\theta\sin\left(\theta\right)\right) \nonumber \\
f_4(\theta) =& \frac{1}{2\theta^5}\left(2\theta-3\sin\left(\theta\right)+\theta\cos\left(\theta\right)\right)  \nonumber
\end{align}
\end{subequations}

\begin{subequations}
\begin{align}
f_1^{'}(\theta) =& \frac{1}{2\theta^3}\left(-8 +(8-\theta^2)\cos\left(\theta\right) + 5\theta\sin\left(\theta\right)\right) \nonumber \\
f_2^{'}(\theta) =& \frac{1}{2\theta^4}\left(-8\theta+(15-\theta^2)\sin\left(\theta\right)-7\theta\cos\left(\theta\right)\right)  \nonumber\\
f_3^{'}(\theta) =& \frac{1}{2\theta^5}\left(-8+(8-\theta^2)\cos\left(\theta\right)+5\theta\sin\left(\theta\right)\right) \nonumber \\
f_4^{'}(\theta) =& \frac{1}{2\theta^6}\left(-8\theta+(15 - \theta^2)\sin\left(\theta\right)-7\theta\cos\left(\theta\right)\right)  \nonumber
\end{align}
\end{subequations}
\normalsize

\section{Identities}
\label{app::B}

\subsection*{Screw Theory Identities}
\small
\begin{equation}
\mytag
\mathrm{ad}_{\bm{\mathcal{V}}}^* = -\mathrm{ad}_{\bm{\mathcal{V}}}^T 
\end{equation}
\begin{equation}
\mathrm{ad}_{\bm{\mathcal{V}}_1}\bm{\mathcal{V}}_2 = -\mathrm{ad}_{\bm{\mathcal{V}}_2}\bm{\mathcal{V}}_1 \mytag \label{I:ad to -ad}
\end{equation}
\begin{equation}
\mytag \label{I:ad* to ad_bar*}
\mathrm{ad}_{\bm{\mathcal{V}}_1}^*\bm{\mathcal{V}}_2 =\overline{\mathrm{ad}}_{\bm{\mathcal{V}}_2}^*\bm{\mathcal{V}}_1 \;
\end{equation}
\normalsize
\small
\begin{equation}
\mytag \label{I:Ad_g*ad_v}
\mathrm{Ad}_{\bm{g}}\mathrm{ad}_{\bm{\mathcal{V}}}=\mathrm{ad}_{\mathrm{Ad}_{\bm{g}}\bm{\mathcal{V}}}\mathrm{Ad}_{\bm{g}}
\end{equation}
\begin{equation}
\mytag \label{I:ad_v*Ad_g}
\mathrm{ad}_{\bm{\mathcal{V}}}\mathrm{Ad}_{\bm{g}}=\mathrm{Ad}_{\bm{g}}\mathrm{ad}_{\mathrm{Ad}_{\bm{g}}^{-1}\bm{\mathcal{V}}}
\end{equation}
\begin{equation}
\mytag
\mathrm{Ad}_{\bm{g}}^*\mathrm{ad}_{\bm{\mathcal{V}}}^*=\mathrm{ad}_{\mathrm{Ad}_{\bm{g}}\bm{\mathcal{V}}}^*\mathrm{Ad}_{\bm{g}}^*
\end{equation}
\begin{equation}
\mytag
\mathrm{ad}_{\bm{\mathcal{V}}}^*\mathrm{Ad}_{\bm{g}}^*=\mathrm{Ad}_{\bm{g}}^*\mathrm{ad}_{\mathrm{Ad}_{\bm{g}}^{-1}\bm{\mathcal{V}}}^*
\end{equation}
\begin{equation}
\mytag
\mathrm{Ad}_{\bm{g}}^*\overline{\mathrm{ad}}_{\bm{\mathcal{V}}}^*=\overline{\mathrm{ad}}_{\mathrm{Ad}_{\bm{g}}^*\bm{\mathcal{V}}}^*\mathrm{Ad}_{\bm{g}}
\end{equation}
\begin{equation}
\mytag \label{I:ad_v*bar Ad_g}
\overline{\mathrm{ad}}_{\bm{\mathcal{V}}}^*\mathrm{Ad}_{\bm{g}}=\mathrm{Ad}_{\bm{g}}^*\overline{\mathrm{ad}}_{\mathrm{Ad}_{\bm{g}}^{-*}\bm{\mathcal{V}}}^*
\end{equation}
\begin{equation}
\mytag \label{I:Adg_dot}
\dot{\mathrm{Ad}}_{\bm{g}}=\mathrm{Ad}_{\bm{g}}\mathrm{ad}_{\bm{\eta}}
\end{equation}
\begin{equation}
\mytag \label{I:Adg*_dot}
\dot{\mathrm{Ad}}_{\bm{g}}^*=\mathrm{Ad}_{\bm{g}}^*\mathrm{ad}_{\bm{\eta}}^*
\end{equation}
\begin{equation}
\mytag \label{I:Adginv_dot}
\dot{\mathrm{Ad}}_{\bm{g}}^{-1}=-\mathrm{ad}_{\bm{\eta}}\mathrm{Ad}_{\bm{g}}^{-1}
\end{equation}
\begin{equation}
\mytag \label{I:Adginv*_dot}
\dot{\mathrm{Ad}}_{\bm{g}}^{-*}=-\mathrm{ad}_{\bm{\eta}}^*\mathrm{Ad}_{\bm{g}}^{-*}
\end{equation}
\begin{equation}
\mytag \label{dexpdt}
\dot{\exp\left({\widehat{\bm{\xi}}}\right)} = \exp\left({\widehat{\bm{\xi}}}\right)\widehat{\left(\mathrm{Ad}_{\exp\left({\widehat{\bm{\xi}}}\right)}^{-1}\mathrm{T}(\bm{\xi})\dot{\bm{\xi}}\right)}
\end{equation}
\begin{equation}
\mytag \label{I:gdot}
\dot{\bm{g}} = \bm{g}\widehat{\bm{\eta}}
\end{equation}
\normalsize

\subsection*{Summation Identities}

For $\bm{M}$ and $\bm{N}$ that are matrix functions,
\small
\begin{equation}
\mytag \label{I:Summation 1}
\sum_{i=1}^n\sum_{j=i}^n \bm{M}_i\bm{N}_j=\sum_{j=1}^n\left(\sum_{i=1}^j \bm{M}_i\right)\bm{N}_j
\end{equation}
\begin{equation}
\mytag \label{I:Summation 2}
\sum_{i=1}^n\sum_{j=1}^{i-1} \bm{M}_i\bm{N}_j=\sum_{j=1}^{n-1}\left(\sum_{i=j+1}^n \bm{M}_i\right)\bm{N}_j
\end{equation}
\begin{equation}
\mytag \label{I:Summation 3}
\sum_{i=k+1}^n\sum_{j=k}^{i-1} \bm{M}_i\bm{N}_j=\sum_{j=k}^{n-1}\left(\sum_{i=j+1}^n \bm{M}_i\right)\bm{N}_j
\end{equation}
\begin{equation}
\mytag \label{I:Summation 4}
\begin{split}
\sum_{i=k+1}^n\sum_{j=1}^{i-1} \bm{M}_i\bm{N}_j=&\left(\sum_{i=k+1}^{n}\bm{M}_i\right)\left(\sum_{j=1}^{k-1} \bm{N}_j\right)\\
+&\sum_{j=k}^{n-1}\left(\sum_{i=j+1}^{n}\bm{M}_i\right) \bm{N}_j
\end{split}
\end{equation}
\normalsize

\section{Derivatives of GVS Elements}
\label{app::C}
\label{app:C}

\subsection*{Magnus Expansion and its Derivatives}
Since a generic Cosserat strain field is non-constant along the material domain, we used the approximate form of Magnus expansion given by the fourth-order Zannah collocation approach \cite{Zanna_JNA1999}  to compute the kinematic and differential kinematic map from one computational point to the next. According to this, the Magnus expansion of $\bm{\xi}$ between $X_{\alpha}$ and $X_{\alpha+1}$ is given by:
\small
\begin{equation}
\mytagb \label{eq::Omega}
\bm{\Omega}_{\alpha} = \frac{h_{\alpha}}{2} \left( {\bm{\xi}_{\alpha}}_{z1} + {\bm{\xi}_{\alpha}}_{z2} \right) + \frac{\sqrt{3}h_{\alpha}^2}{12} \mathrm{ad}_{{\bm{\xi}_{\alpha}}_{z1}}{\bm{\xi}_{\alpha}}_{z2}
\end{equation}
\normalsize
where $h_{\alpha} = X_{\alpha+1}-X_{\alpha}$ and subscripts `$z1$' and `$z2$' refer to quantities evaluated at the first and second Zannah collocation points: $X_{\alpha}+(1/2\mp \sqrt{3}/6)h_{\alpha}$.

First and second time derivatives of $\bm{\Omega}$ are given by:
\small
\begin{subequations}
\begin{align}
\dot{\bm{\Omega}}_{\alpha} =& \frac{h_\alpha}{2} \left( \dot{{\bm{\xi}}_{\alpha}}_{z1} + \dot{{\bm{\xi}}_{\alpha}}_{z2} \right)+\frac{\sqrt{3}h_\alpha^2}{12} \left( \mathrm{ad}_{{\bm{\xi}_{\alpha}}_{z1}}\dot{{\bm{\xi}}_{\alpha}}_{z2} - \mathrm{ad}_{{\bm{\xi}_{\alpha}}_{z2}}\dot{{\bm{\xi}}_{\alpha}}_{z1} \right) \mytagb \label{eq::Omegad_full} \\
\ddot{\bm{\Omega}}_{\alpha}=& \frac{h_\alpha}{2} \left( \ddot{{\bm{\xi}}_\alpha}_{z1} + \ddot{{\bm{\xi}}_\alpha}_{z2} \right)+ \frac{\sqrt{3}h_\alpha^2}{12} \left( \mathrm{ad}_{{\bm{\xi}_{\alpha}}_{z1}}\ddot{{\bm{\xi}}_\alpha}_{z2} - \mathrm{ad}_{{\bm{\xi}_{\alpha}}_{z2}}\ddot{{\bm{\xi}}_\alpha}_{z1} \right) \nonumber \\
&+ \frac{\sqrt{3}h_\alpha^2}{6}\mathrm{ad}_{\dot{{\bm{\xi}}_\alpha}_{z1}}\dot{{\bm{\xi}}_\alpha}_{z2} \mytagb  \label{eq::Omegadd_full}
\end{align}
\end{subequations}
\normalsize

By virtue of the Magnus expansion, the soft body is computationally identical to $n_p-1$ rigid joints with equivalent constant joint strain ($\bm{\Omega}_{\alpha}$) and its time derivatives ($\dot{\bm{\Omega}}_{\alpha}$ and $\ddot{\bm{\Omega}}_{\alpha}$). By introducing $\bm{q}$ from \eqref{eq::strainparameterization} and taking partial derivative of \eqref{eq::Omega} with respect to $\bm{q}$ we get,

\begin{equation}
\mytagb \label{eq::Omega_qDerivative}
\frac{\partial \bm{\Omega}_{\alpha}}{\partial \bm{q}} = \bm{\mathcal{Z}}_{\alpha}
\end{equation}
 where $\bm{\mathcal{Z}}_{\alpha}  \in \mathbb{R}^{6\times {n_{dof}}_i}$ is given by:

\small
\begin{equation}
\mytagb \label{eq::Z}
\begin{split}
\bm{\mathcal{Z}}_{\alpha} = \frac{h_{\alpha}}{2} \left( {\bm{\Phi}_{\xi}}_{z1} + {\bm{\Phi}_{\xi}}_{z2} \right) + \frac{\sqrt{3}h_{\alpha}^2}{12} \left( \mathrm{ad}_{\bm{\xi}_{z1}}{\bm{\Phi}_{\xi}}_{z2} - \mathrm{ad}_{\bm{\xi}_{z2}}{\bm{\Phi}_{\xi}}_{z1} \right)
\end{split}
\end{equation}
\normalsize
Note that for a rigid joint, $\bm{\Omega}_\alpha = \bm{\xi}_\alpha $ and $\bm{\mathcal{Z}}_{\alpha} = {\bm{\Phi}_{\xi}}_{\alpha}$.

Using this we can rewrite \eqref{eq::Omegad_full} and \eqref{eq::Omegadd_full} as:

\small
\begin{subequations}
\begin{align}
\dot{\bm{\Omega}}_{\alpha} =& \bm{\mathcal{Z}}_{\alpha}\dot{\bm{q}} \mytagb \label{eq::Omegad} \\
\ddot{\bm{\Omega}}_{\alpha}=& \bm{\mathcal{Z}}_{\alpha}\ddot{\bm{q}} +\dot{\bm{\mathcal{Z}}}_{\alpha}\dot{\bm{q}}  \mytagb 
 \label{eq::Omegadd}
\end{align}
\end{subequations}
\normalsize
where,

\small
\begin{equation}
\mytagb \label{eq::Zd}
\begin{split}
\dot{\bm{\mathcal{Z}}}_{\alpha} =&  \frac{\sqrt{3}h_{\alpha}^2}{12} \left( \mathrm{ad}_{\dot{\bm{\xi}}_{z1}}{\bm{\Phi}_{\xi}}_{z2} - \mathrm{ad}_{\dot{\bm{\xi}}_{z2}}{\bm{\Phi}_{\xi}}_{z1} \right)
\end{split}
\end{equation}
\normalsize

Using \eqref{eq::Omegad} we derive the derivatives of $\dot{\bm{\Omega}}_{\alpha}$ with respect to $\bm{q}$ and $\dot{\bm{q}}$ as follows:

\small
\begin{subequations}
\begin{align}
\frac{\partial \dot{\bm{\Omega}}_{\alpha}}{\partial \bm{q}}  =& \frac{\partial \bm{\mathcal{Z}}_{\alpha}}{\partial \bm{q}}\dot{\bm{q}} = \dot{\bm{\mathcal{Z}}}_{\alpha} \; \mytagb  \label{eq::Omegad_qDerivative} \\
\frac{\partial \dot{\bm{\Omega}}_{\alpha}}{\partial \dot{\bm{q}}}  =& \bm{\mathcal{Z}}_{\alpha} \; \mytagb 
 \label{eq::Omegad_qdDerivative}
\end{align}
\end{subequations}
\normalsize

Similarly, the partial derivatives of $\ddot{\bm{\Omega}}_{\alpha}$ with respect to $\bm{q}$, $\dot{\bm{q}}$, and $\dot{\bm{q}}$ are given by:
\small
\begin{subequations}
\begin{align}
\frac{\partial \ddot{\bm{\Omega}}_{\alpha}}{\partial \bm{q}}  =&\frac{\partial \bm{\mathcal{Z}}_{\alpha}}{\partial \bm{q}}\ddot{\bm{q}}= \ddot{\bm{\mathcal{Z}}}_{\alpha} \; \mytagb \label{eq::Omegadd_qDerivative} \text{,}\\
\frac{\partial \ddot{\bm{\Omega}}_{\alpha}}{\partial \dot{\bm{q}}}  =&\frac{\partial \dot{\bm{\mathcal{Z}}}_{\alpha}}{\partial \dot{\bm{q}}}\dot{\bm{q}}+\dot{\bm{\mathcal{Z}}}_{\alpha}= 2\dot{\bm{\mathcal{Z}}}_{\alpha} \; \mytagb \label{eq::Omegadd_qdDerivative}\\
\frac{\partial \ddot{\bm{\Omega}}_{\alpha}}{\partial \ddot{\bm{q}}}  =& \bm{\mathcal{Z}}_{\alpha} \; \mytagb \label{eq::Omegadd_qddDerivative}
\end{align}
\end{subequations}
\normalsize
where,
\small
\begin{equation}
\mytagb \label{eq::Zdd}
\begin{split}
\ddot{\bm{\mathcal{Z}}}_{\alpha} =&  \frac{\sqrt{3}h_{\alpha}^2}{12} \left( \mathrm{ad}_{\ddot{\bm{\xi}}_{z1}}{\bm{\Phi}_{\xi}}_{z2} - \mathrm{ad}_{\ddot{\bm{\xi}}_{z2}}{\bm{\Phi}_{\xi}}_{z1} \right)
\end{split}
\end{equation}
\normalsize

\subsection*{Joint Motion Subspace and its Derivatives}
The joint motion subspace of the virtual rigid joint $\alpha$ is given by:

\small
\begin{equation}
\mytagb \label{eq::MotionSubspace}
\bm{S}_{\alpha} = \bm{T}_\alpha\bm{\mathcal{Z}}_{\alpha}
\end{equation}
\normalsize
where $\bm{T}_\alpha$ is the tangent operator (Appendix \ref{app::A}).

The relative velocity of the joint expressed in the frame at $X=0$ is given by $\bm{S}_{\alpha}\dot{\bm{q}}$. The partial derivative of this term with respect to $\bm{q}$ is given by:

\small
\begin{equation}
\mytagb \label{eq::dSdq_qd_start}
\begin{split}
\frac{\partial \bm{S}_{\alpha}}{\partial \bm{q}}\dot{\bm{q}} =&\frac{\partial \bm{T}_{\alpha}}{\partial \bm{q}}\bm{\mathcal{Z}}_{\alpha}\dot{\bm{q}}+\bm{T}_{\alpha}\frac{\partial \bm{\mathcal{Z}}_{\alpha}}{\partial \bm{q}}\dot{\bm{q}}\\
=&\frac{\partial \bm{T}_{\alpha}}{\partial \bm{q}}\dot{\bm{\Omega}}_\alpha+\bm{T}_{\alpha}\dot{\bm{\mathcal{Z}}}_{\alpha}
\end{split}
\end{equation}
\normalsize

To derive partial derivatives of $\bm{T}_{\alpha}$, we need to compute $\frac{\partial \theta_{\alpha}}{\partial \bm{q}}$ and $\frac{\partial \mathrm{ad}_{\bm{\Omega_{\alpha}}}^r}{\partial \bm{q}}\dot{\bm{\Omega}}_\alpha$. From the definition of $\theta$ (Appendix \ref{app::A}) we derive,

\small
\begin{equation}
\mytagb \label{eq::dthetadq}
\frac{\partial \theta_{\alpha}}{\partial \bm{q}} = \frac{1}{\theta}\bm{\Omega}_\alpha^T \bm{I}_{\theta}\bm{\mathcal{Z}}_{\alpha}
\end{equation}
\normalsize

With appropriate mathematical techniques, we can see that

\small
\begin{equation}
\mytagb \label{eq::dadOmega^rq}
\frac{\partial \mathrm{ad}_{\bm{\Omega_{\alpha}}}^r}{\partial \bm{q}}\dot{\bm{\Omega}}_\alpha = -\sum_{u=1}^{r}\mathrm{ad}_{\bm{\Omega_{\alpha}}}^{u-1}\mathrm{ad}_{\mathrm{ad}_{\bm{\Omega_{\alpha}}}^{r-u}\dot{\bm{\Omega}}_\alpha}\bm{\mathcal{Z}}_{\alpha}
\end{equation}
\normalsize

Putting these together in \eqref{eq::dSdq_qd_start} we get:

\small
\begin{equation}
\mytagb \label{eq::dSdq_qd_end}
\begin{split}
\frac{\partial \bm{S}_{\alpha}}{\partial \bm{q}}\dot{\bm{q}} =& \frac{1}{\theta}\sum_{r=1}^{4}f_r^{'} (\theta)\mathrm{ad}_{\bm{\Omega}_\alpha}^r\dot{\bm{\Omega}}_\alpha\bm{\Omega}_\alpha^T\bm{I}_{\theta}\bm{\mathcal{Z}}_{\alpha} \\
-& \sum_{r=1}^{4}f_r (\theta)\sum_{u=1}^{r}\mathrm{ad}_{\bm{\Omega_{\alpha}}}^{u-1}\mathrm{ad}_{\mathrm{ad}_{\bm{\Omega_{\alpha}}}^{r-u}\dot{\bm{\Omega}}_\alpha}\bm{\mathcal{Z}}_{\alpha}+ \bm{T}_\alpha \dot{\bm{\mathcal{Z}}}_{\alpha}
\end{split}
\end{equation}
\normalsize

Note that due to the order of tensor multiplication, here $\frac{\partial \bm{S}_{\alpha}}{\partial \bm{q}}\dot{\bm{q}} \neq \dot{\bm{S}}_{\alpha}$. In the element notation, the former is given by $\frac{\partial {S_{\alpha}}_{ij}}{\partial q_k}\dot{q_j}$, while the latter is given by $\frac{\partial {S_{\alpha}}_{ij}}{\partial q_k}\dot{q_k}$.

Similarly, we obtain:

\small
\begin{equation}
\mytagb \label{eq::dSdq_qdd_end}
\begin{split}
\frac{\partial \bm{S}_{\alpha}}{\partial \bm{q}}\ddot{\bm{q}} =& \frac{1}{\theta}\sum_{r=1}^{4}f_r^{'} (\theta)\mathrm{ad}_{\bm{\Omega}_\alpha}^r\bm{\mathcal{Z}}_{\alpha}\ddot{\bm{q}}\bm{\Omega}_\alpha^T\bm{I}_{\theta}\bm{\mathcal{Z}}_{\alpha} \\
-& \sum_{r=1}^{4}f_r (\theta)\sum_{u=1}^{r}\mathrm{ad}_{\bm{\Omega_{\alpha}}}^{u-1}\mathrm{ad}_{\mathrm{ad}_{\bm{\Omega_{\alpha}}}^{r-u}\bm{\mathcal{Z}}_{\alpha}\ddot{\bm{q}}}\bm{\mathcal{Z}}_{\alpha}+\bm{T}_{\alpha}\ddot{\bm{\mathcal{Z}}}_{\alpha}
\end{split}
\end{equation}
\normalsize

Moving to the time derivative of joint motion subspace $\dot{\bm{S}}_{\alpha}$, it is obtained by taking a time derivative of \eqref{eq::MotionSubspace},

\small
\begin{equation}
\mytagb \label{eq::MotionSubspace_derivative}
\dot{\bm{S}}_{\alpha} = \dot{\bm{T}}_\alpha\bm{\mathcal{Z}}_{\alpha}+\bm{T}_\alpha \dot{\bm{\mathcal{Z}}}_{\alpha}
\end{equation}
\normalsize
where $\dot{\bm{T}}$ is provided in Appendix \ref{app::A}.
Then, the derivative of $\dot{\bm{S}}_\alpha\dot{\bm{q}}$ with respect to $\bm{q}$ is given by:
\small
\begin{equation}
\mytagb \label{eq::dSddq_dq_start}
\begin{split}
\frac{\partial \dot{\bm{S}}_{\alpha}}{\partial \bm{q}}\dot{\bm{q}} =&\frac{\partial \dot{\bm{T}}_\alpha}{\partial \bm{q}}\bm{\mathcal{Z}}_{\alpha}\dot{\bm{q}}+\dot{\bm{T}}_\alpha\frac{\partial \bm{\mathcal{Z}}_{\alpha}}{\partial \bm{q}}\dot{\bm{q}}+\frac{\partial \bm{T}_\alpha}{\partial \bm{q}}\dot{\bm{\mathcal{Z}}}_{\alpha}\dot{\bm{q}}\\
=& \frac{\partial \dot{\bm{T}}_\alpha}{\partial \bm{q}}\dot{\bm{\Omega}}_{\alpha}+\dot{\bm{T}}_\alpha\dot{\bm{\mathcal{Z}}}_{\alpha}+\frac{\partial \bm{T}_\alpha}{\partial \bm{q}}\dot{\bm{\mathcal{Z}}}_{\alpha}\dot{\bm{q}}
\end{split}
\end{equation}
\normalsize

To evaluate the partial derivative of $\dot{\bm{T}_\alpha}$, we need to compute $\frac{\partial \dot{\theta}_{\alpha}}{\partial \bm{q}}$. Using the definition of $\dot{\theta}$ in Appendix \ref{app::A}, we get

\small
\begin{equation}
\mytagb \label{eq::dthetaddq}
\frac{\partial \dot{\theta}_{\alpha}}{\partial \bm{q}} = -\frac{\dot{\theta}}{\theta^2}\bm{\Omega}_\alpha^T \bm{I}_{\theta}\bm{\mathcal{Z}}_{\alpha}+\frac{1}{\theta}\dot{\bm{\Omega}}_\alpha^T \bm{I}_{\theta}\bm{\mathcal{Z}}_{\alpha}+\frac{1}{\theta}\bm{\Omega}_\alpha^T \bm{I}_{\theta}\dot{\bm{\mathcal{Z}}}_{\alpha}
\end{equation}
\normalsize

Substituting \eqref{eq::dthetadq}, \eqref{eq::dadOmega^rq}, and \eqref{eq::dthetaddq} in \eqref{eq::dSddq_dq_start} we get:

\small
\begin{equation}
\mytagb \label{eq::dSddq_dq_end}
\begin{split}
\frac{\partial \dot{\bm{S}}_{\alpha}}{\partial \bm{q}}\dot{\bm{q}} =& \frac{1}{\theta}\sum_{r=1}^{4} \left(f_r^{''}(\theta)\dot{\theta}\mathrm{ad}_{\bm{\Omega}}^r+f_r^{'}(\theta)\dot{\left(\mathrm{ad}^r_{\bm{\Omega}}\right)}\right)\dot{\bm{\Omega}}_\alpha\bm{\Omega}_\alpha^T\bm{I}_{\theta}\bm{\mathcal{Z}}_{\alpha} \\
+&\frac{1}{\theta}\sum_{r=1}^{4} f_r^{'}(\theta)\mathrm{ad}_{\bm{\Omega}}^r\dot{\bm{\Omega}}_\alpha\left((\dot{\bm{\Omega}}_\alpha^T-\frac{\dot{\theta}}{\theta}\bm{\Omega}_\alpha^T)\bm{I}_{\theta}\bm{\mathcal{Z}}_{\alpha}+\bm{\Omega}_\alpha^T\bm{I}_{\theta}\dot{\bm{\mathcal{Z}}}_{\alpha}\right)\\
-&\sum_{r=1}^{4} f_r^{'}(\theta)\dot{\theta}\sum_{u=1}^{r}\mathrm{ad}_{\bm{\Omega}_{\alpha}}^{u-1}\mathrm{ad}_{\mathrm{ad}_{\bm{\Omega}_{\alpha}}^{r-u}\dot{\bm{\Omega}}_\alpha}\bm{\mathcal{Z}}_{\alpha}\\
-& \sum_{r=1}^{4}f_r (\theta)\sum_{u=1}^{r}\biggl(\sum_{p=1}^{u-1}\mathrm{ad}_{\bm{\Omega}_{\alpha}}^{p-1}\mathrm{ad}_{\mathrm{ad}_{\bm{\Omega}_{\alpha}}^{u-p-1}\mathrm{ad}_{\bm{\dot{\Omega}}_{\alpha}}\mathrm{ad}_{\bm{\Omega}_{\alpha}}^{r-u}\dot{\bm{\Omega}}_\alpha}\bm{\mathcal{Z}}_{\alpha} \\
+& \mathrm{ad}_{\bm{\Omega}_{\alpha}}^{u-1}\mathrm{ad}_{\dot{\bm{\Omega}}_{\alpha}} \sum_{p=1}^{r-u}\mathrm{ad}_{\bm{\Omega}_{\alpha}}^{p-1}\mathrm{ad}_{\mathrm{ad}_{\bm{\Omega}_{\alpha}}^{r-u-p}\dot{\bm{\Omega}}_\alpha}\bm{\mathcal{Z}}_{\alpha} \\
+& \mathrm{ad}_{\bm{\Omega}_{\alpha}}^{u-1}\mathrm{ad}_{\mathrm{ad}_{\bm{\Omega}_{\alpha}}^{r-u}\dot{\bm{\Omega}}_\alpha}\dot{\bm{\mathcal{Z}}}_{\alpha} \biggr)\\
+&\frac{1}{\theta}\sum_{r=1}^{4}f_r^{'} (\theta)\mathrm{ad}_{\bm{\Omega}}^r\dot{\bm{\mathcal{Z}}}_{\alpha}\dot{\bm{q}}\bm{\Omega}_\alpha^T\bm{I}_{\theta}\bm{\mathcal{Z}}_{\alpha} \\
-& \sum_{r=1}^{4}f_r (\theta)\sum_{u=1}^{r}\mathrm{ad}_{\bm{\Omega_{\alpha}}}^{u-1}\mathrm{ad}_{\mathrm{ad}_{\bm{\Omega_{\alpha}}}^{r-u}\dot{\bm{\mathcal{Z}}}_{\alpha}\dot{\bm{q}}}\bm{\mathcal{Z}}_{\alpha}\\
+& \dot{\bm{T}}_\alpha \dot{\bm{\mathcal{Z}}}_{\alpha}
\end{split}
\end{equation}
\normalsize
where,
\small
\begin{subequations}
\begin{align}
f_1^{''}(\theta) =& \frac{1}{2\theta^4} \left(24-(24-6\theta^2)\cos(\theta)-(18\theta-\theta^3)\sin(\theta)\right) \nonumber \\
f_2^{''}(\theta) =& \frac{1}{2\theta^5} \left(24\theta-(60-9\theta^2)\sin(\theta)+(36\theta-\theta^3)\cos(\theta)\right)  \nonumber\\
f_3^{''}(\theta) =& \frac{1}{2\theta^6} \left(40-(40-8\theta^2)\cos(\theta)-(28\theta-\theta^3)\sin(\theta)\right) \nonumber \\
f_4^{''}(\theta) =& \frac{1}{2\theta^7} \left(40\theta-(90-11\theta^2)\sin(\theta) + (50\theta-\theta^3)\cos(\theta) \right)  \nonumber
\end{align}
\end{subequations}
\normalsize




Also, noticing that $\frac{\partial \dot{\theta}_{\alpha}}{\partial \dot{\bm{q}}}= \frac{\partial \theta_{\alpha}}{\partial \bm{q}}$, and $\frac{\partial \dot{\bm{\Omega}}_{\alpha}}{\partial \dot{\bm{q}}}= \frac{\partial \bm{\Omega}_{\alpha}}{\partial \bm{q}}$, it can be seen that,

\small
\begin{equation}
\mytagb \label{eq::dSdqd_qdd_end}
\frac{\partial \dot{\bm{S}}_{\alpha}}{\partial \dot{\bm{q}}}\dot{\bm{q}} = \frac{\partial \bm{S}_{\alpha}}{\partial \bm{q}}\dot{\bm{q}}
\end{equation}
\normalsize



Finally, the first term in the partial derivative of $ID$ with respect to $\bm{q}$ in \eqref{eq::dtau_dq_start} is $\frac{\partial \bm{S}_{\alpha}^T}{\partial \bm{q}}\bm{\mathcal{F}}^C$. From \eqref{eq::MotionSubspace} we get,

\small
\begin{equation}
\mytagb \label{eq::dSTrasposeF_dq_start}
\frac{\partial \bm{S}_{\alpha}^T}{\partial \bm{q}}\bm{\mathcal{F}}^C= \frac{\partial \bm{\mathcal{Z}}_{\alpha}^T}{\partial \bm{q}}\bm{T}_\alpha^T\bm{\mathcal{F}}^C+ \bm{\mathcal{Z}}_{\alpha}^T\frac{\partial \bm{T}_\alpha^T}{\partial \bm{q}}\bm{\mathcal{F}}^C
\end{equation}
\normalsize
where,

\small
\begin{equation}
\mytagb \label{eq::Z_traspose}
\begin{split}
\bm{\mathcal{Z}}_{\alpha}^T = \frac{h_{\alpha}}{2} \left( {\bm{\Phi}_{\xi}}_{z1}^T + {\bm{\Phi}_{\xi}}_{z2}^T \right)+ \frac{\sqrt{3}h_{\alpha}^2}{12} \left( {\bm{\Phi}_{\xi}}_{z1}^T\mathrm{ad}_{\bm{\xi}_{z2}}^* - {\bm{\Phi}_{\xi}}_{z2}^T\mathrm{ad}_{\bm{\xi}_{z1}}^* \right)
\end{split}
\end{equation}
and
\small
\begin{equation}
\bm{T}^T(\Omega) = \bm{I}_6 + \sum_{r=1}^{4}(-1)^rf_r(\theta)\mathrm{ad}_{\bm{\Omega}}^{*r}
\label{eq::T_traspose}
\mytagb \end{equation}
\normalsize

Based on previously derived results and the identity \eqref{I:ad* to ad_bar*} we obtain:

\small
\begin{equation}
\mytagb\label{eq::dSTrasposeF_dq_end}
\begin{split}
\frac{\partial \bm{S}_{\alpha}^T}{\partial \bm{q}}\bm{\mathcal{F}}^C =&\frac{\sqrt{3}h_{\alpha}^2}{12} \left( {\bm{\Phi}_{\xi}}_{z1}^T\overline{\mathrm{ad}}_{\bm{T}_\alpha^T\bm{\mathcal{F}}^C}^{*}{\bm{\Phi}_{\xi}}_{z2} - {\bm{\Phi}_{\xi}}_{z2}^T\overline{\mathrm{ad}}_{\bm{T}_\alpha^T\bm{\mathcal{F}}^C}^{*}{\bm{\Phi}_{\xi}}_{z1} \right)\\
+& \frac{1}{\theta}\bm{\mathcal{Z}}_{\alpha}^T\sum_{r=1}^{4} (-1)^r f_r^{'} (\theta)\mathrm{ad}_{\bm{\Omega}_\alpha}^{*r}\bm{\mathcal{F}}^C\bm{\Omega}_\alpha^T\bm{I}_{\theta}\bm{\mathcal{Z}}_{\alpha} \\
+& \bm{\mathcal{Z}}_{\alpha}^T\sum_{r=1}^{4}(-1)^rf_r (\theta)\sum_{u=1}^{r}\mathrm{ad}_{\bm{\Omega_{\alpha}}}^{*u-1}\overline{\mathrm{ad}}^*_{\mathrm{ad}_{\bm{\Omega_{\alpha}}}^{*r-u}\bm{\mathcal{F}}^C}\bm{\mathcal{Z}}_{\alpha}
\end{split}
\end{equation}
\normalsize
\subsection*{Derivatives of Adjoint Maps}
Using identities \eqref{I:Adg_dot} and \eqref{I:Ad_g*ad_v} we get,

\small
\begin{equation}
\frac{\partial \mathrm{Ad}_{\exp\left(\widehat{\bm{\Omega}}_\alpha\right)}}{\partial q_p}\bm{\mathcal{V}} = -\mathrm{ad}_{\mathrm{Ad}_{\exp\left(\widehat{\bm{\Omega}}_\alpha\right)}\bm{\mathcal{V}}}\bm{S}_{\alpha,p}
\mytagb \label{eq::Adg_dqp}
\end{equation}
\normalsize
where $q_p$ is the $p^{th}$ component of $\bm{q}$ and $\bm{S}_{\alpha,p}$ is the $p^{th}$ column of $\bm{S}_{\alpha}$. Combining all the columns, we get:

\small
\begin{equation}
\frac{\partial \mathrm{Ad}_{\exp\left(\widehat{\bm{\Omega}}_\alpha\right)}}{\partial \bm{q}}\bm{\mathcal{V}} = -\mathrm{ad}_{\mathrm{Ad}_{\exp\left(\widehat{\bm{\Omega}}_\alpha\right)}\bm{\mathcal{V}}}\bm{S}_{\alpha}
\mytagb \label{eq::Adg_dq}
\end{equation}
\normalsize

Similarly using identities \eqref{I:Adg*_dot}, \eqref{I:Adginv_dot}, and \eqref{I:Adginv*_dot} we derive:

\small
\begin{equation}
\frac{\partial \mathrm{Ad}_{\exp\left(\widehat{\bm{\Omega}}_\alpha\right)}^{*}}{\partial \bm{q}}\bm{\mathcal{F}} = \overline{\mathrm{ad}}_{\mathrm{Ad}_{\exp\left(\widehat{\bm{\Omega}}_\alpha\right)}^*\bm{\mathcal{F}}}^*\bm{S}_{\alpha}
\mytagb \label{eq::Adg*_dq}
\end{equation}
\begin{equation}
\frac{\partial \mathrm{Ad}_{\exp\left(\widehat{\bm{\Omega}}_\alpha\right)}^{-1}}{\partial \bm{q}}\bm{\mathcal{V}} = \mathrm{Ad}_{\exp\left(\widehat{\bm{\Omega}}_\alpha\right)}^{-1}\mathrm{ad}_{\bm{\mathcal{V}}}\bm{S}_{\alpha}
\mytagb \label{eq::Adginv_dq}
\end{equation}
\begin{equation}
\frac{\partial \mathrm{Ad}_{\exp\left(\widehat{\bm{\Omega}}_\alpha\right)}^{-*}}{\partial \bm{q}}\bm{\mathcal{F}} = -\mathrm{Ad}_{\exp\left(\widehat{\bm{\Omega}}_\alpha\right)}^{-*}\overline{\mathrm{ad}}_{\bm{\mathcal{F}}}^*\bm{S}_{\alpha}
\mytagb \label{eq::Adginv*_dq}
\end{equation}
\normalsize

\section{Section-wise Derivations}
\label{app::D}
\subsection*{Derivations in \ref{sec::dID_dq}}

Inverse dynamics of the soft body, enabled by the ``Pseudo rigid joint'' formulation of GVS is given by:

\small
\begin{equation}
\mytagd \label{eq::IDsum_app}
\begin{split}
ID = \sum_{\alpha=1}^{n_p-1} ID_\alpha =\sum_{\alpha=1}^{n_p-1} \bm{S}_\alpha^T\bm{\mathcal{F}}_\alpha^C
\end{split}
\end{equation}
\normalsize
where $\bm{\mathcal{F}}_\alpha^C$ is computed recursively according to:

\small
\begin{equation}
\bm{\mathcal{F}}_\alpha^C = \sum_{k=\alpha+1}^{n_p}\mathrm{Ad}_{\bm{g}_{\alpha k}}^*\bm{\mathcal{F}}_k=\mathrm{Ad}_{\bm{g}_{\alpha \alpha+1}}^*(\bm{\mathcal{F}}_{\alpha+1}+\bm{\mathcal{F}}_{\alpha+1}^C)
\mytagd \label{eq::FC_recursive_app}
\end{equation}
\normalsize

The partial derivative of $ID_\alpha$ with respect to $\bm{q}$ is given by:

\small
\begin{equation}
\begin{split}
\frac{\partial ID_\alpha}{\partial \bm{q}} =&\frac{\partial \bm{S}_\alpha^T}{\partial \bm{q}}\bm{\mathcal{F}}_\alpha^C+\bm{S}_\alpha^T\sum_{k=\alpha+1}^{n_p}\frac{\partial \mathrm{Ad}_{\bm{g}_{\alpha k}}^*}{\partial \bm{q}}\bm{\mathcal{F}}_k\\
+&\bm{S}_\alpha^T\sum_{k=\alpha+1}^{n_p}\mathrm{Ad}_{\bm{g}_{\alpha k}}^*\frac{\partial \bm{\mathcal{F}}_{k}}{\partial \bm{q}}\\
=&\frac{\partial \bm{S}_\alpha^T}{\partial \bm{q}}\bm{\mathcal{F}}_\alpha^C+\bm{S}_\alpha^T\left(\bm{\mathcal{N}}_\alpha^C\bm{R}_{\alpha}^B+\bm{\mathcal{M}}_\alpha^C\bm{Q}_{\alpha}^B+\bm{U}_\alpha^S+\bm{P}_\alpha^S\right)
\end{split}
\mytagd \label{eq::dtau_dq_end_app}
\end{equation}
\normalsize

The analytical derivative of the first term in \eqref{eq::dtau_dq_end_app} is given by \eqref{eq::dSTrasposeF_dq_end}. The term $\bm{P}_\alpha^S$ arises from the partial derivative of the coAdjoint maps. The rest of the terms come from the partial derivatives of $\bm{\mathcal{F}}_k$ transformed to the frame of $\alpha$.

\noindent
\rule{\columnwidth}{0.8pt}
\uline{Derivation of $\boldsymbol{P}_\alpha^S$ in \eqref{eq::dtau_dq_end_app}}

Using the identity, $\mathrm{Ad}_{\bm{g}_{\alpha k}}^* = \prod_{\beta=\alpha}^{k-1}\mathrm{Ad}_{\bm{g}_{\beta \beta+1}}^*$ and \eqref{eq::Adg*_dq} we get:

\small
\begin{equation}
\mytagd \label{eq::Adg*alphak_dqp_app}
\frac{\partial \mathrm{Ad}_{\bm{g}_{\alpha k}}^{*}}{\partial q_p}\bm{\mathcal{F}}_k = \sum_{\beta=\alpha}^{k-1}\mathrm{Ad}_{\bm{g}_{\alpha \beta}}^*\overline{\mathrm{ad}}_{\mathrm{Ad}_{\bm{g}_{\beta k}}^*\bm{\mathcal{F}}_k}^*\bm{S}_{\beta,p}
\end{equation}
\normalsize
where $q_p$ is the $p^{th}$ component of $\bm{q}$ and $\bm{S}_{\alpha,p}$ is the $p^{th}$ column of $\bm{S}_{\alpha}$. Combining all the columns, we get:

\small
\begin{equation}
\mytagd \label{eq::Adg*alphak_dq_app}
\frac{\partial \mathrm{Ad}_{\bm{g}_{\alpha k}}^{*}}{\partial \bm{q}}\bm{\mathcal{F}}_k = \sum_{\beta=\alpha}^{k-1}\mathrm{Ad}_{\bm{g}_{\alpha \beta}}^*\overline{\mathrm{ad}}_{\mathrm{Ad}_{\bm{g}_{\beta k}}^*\bm{\mathcal{F}}_k}^*\bm{S}_{\beta}
\end{equation}
\normalsize

Using this we derive,

\small
\begin{equation}
\mytagd \label{eq::dID_dq_II_start}
\sum_{k=\alpha+1}^{n_p}\frac{\partial \mathrm{Ad}_{\bm{g}_{\alpha k}}^*}{\partial \bm{q}}\bm{\mathcal{F}}_k=\sum_{k=\alpha+1}^{n_p}\sum_{\beta=\alpha}^{k-1}\mathrm{Ad}_{\bm{g}_{\alpha \beta}}^*\overline{\mathrm{ad}}_{\mathrm{Ad}_{\bm{g}_{\beta k}}^*\bm{\mathcal{F}}_k}^*\bm{S}_{\beta}
\end{equation}
\normalsize

Using summation identity \eqref{I:Summation 3}, and identities \eqref{I:Ad_g*ad_v} and \eqref{I:ad_v*Ad_g} we get:

\small
\begin{equation}
\mytagd \label{eq::dID_dq_II_app}
\sum_{k=\alpha+1}^{n_p}\frac{\partial \mathrm{Ad}_{\bm{g}_{\alpha k}}^*}{\partial \bm{q}}\bm{\mathcal{F}}_k=\sum_{\beta=\alpha}^{n_p-1}\mathrm{Ad}_{\bm{g}_{\alpha \beta}}^*\overline{\mathrm{ad}}_{\bm{\mathcal{F}}_\beta^C}^*\bm{S}_{\beta}=\bm{P}_\alpha^S
\end{equation}
\normalsize

Recursive computation of $\bm{P}_\alpha^S$ can be derived from \eqref{eq::dID_dq_II_app}:

\small
\begin{equation}
\mytagd \label{eq::PS_recursive_app}
\bm{P}_\alpha^S = \overline{\mathrm{ad}}_{\bm{\mathcal{F}}_\alpha^C}^*\bm{S}_{\alpha}+\mathrm{Ad}_{\bm{g}_{\alpha \alpha+1}}^*\bm{P}_{\alpha+1}^S
\end{equation}
\normalsize

\noindent
\rule{\columnwidth}{0.8pt}
\uline{Derivation of $\bm{\mathcal{N}}_\alpha^C\bm{R}_{\alpha}^B+\bm{\mathcal{M}}_\alpha^C\bm{Q}_{\alpha}^B+\bm{U}_\alpha^S$ in \eqref{eq::dtau_dq_end_app}}

The resultant point wrench in the local frame is given by,

\small
\begin{equation}
\mytagd \label{eq::TotalWrench_app}
\bm{\mathcal{F}}_k = \bm{\mathcal{M}}_k\dot{\bm{\eta}}_k+\mathrm{ad}_{\bm{\eta}_k}^*\bm{\mathcal{M}}_k\bm{\eta}_k-\bm{\mathcal{M}}_k\mathrm{Ad}_{\bm{g}_{k}}^{-1}\bm{\mathcal{G}}
\end{equation}
\normalsize

Using, $\dot{\bm{\eta}}_k = \bm{\gamma}_k+\bm{\varphi}_k$ (inertial and Coriolis components of acceleration), we rewrite $\bm{\mathcal{F}}_k$ as the resultant of inertial, Coriolis, and gravitational forces.

\small
\begin{equation}
\mytagd \label{eq::TotalWrench_app2}
\begin{split}
\bm{\mathcal{F}}_k =& \left(\bm{\mathcal{M}}_k\bm{\gamma}_k\right)+\left(\mathrm{ad}_{\bm{\eta}_k}^*\bm{\mathcal{M}}_k\bm{\eta}_k+\bm{\mathcal{M}}_k\bm{\varphi}_k\right)-\left(\bm{\mathcal{M}}_k\mathrm{Ad}_{\bm{g}_{k}}^{-1}\bm{\mathcal{G}}\right)\\
=&\bm{\mathcal{F}}_{Ik}+\bm{\mathcal{F}}_{Ck}+\bm{\mathcal{F}}_{Gk}
\end{split}
\end{equation}
\normalsize

The partial derivatives of $\bm{\mathcal{F}}_k$ involves the partial derivatives of $\bm{\mathcal{F}}_{Ik}$, $\bm{\mathcal{F}}_{Ck}$, and $\bm{\mathcal{F}}_{Gk}$.

\noindent
\rule{\columnwidth}{0.4pt}
\uline{Contribution of Gravity}

 We have,

\small
\begin{equation}
\mytagd \label{eq::Gravity}
\bm{\mathcal{F}}_{Gk} = -\bm{\mathcal{M}}_k\mathrm{Ad}_{\bm{g}_{k}}^{-1}\bm{\mathcal{G}}
\end{equation}
\normalsize

Using $\mathrm{Ad}_{\bm{g}_{k}}^{-1} = \prod_{\beta=k-1}^{1}\mathrm{Ad}_{\bm{g}_{\beta \beta+1}}^{-1}$ and \eqref{eq::Adginv_dq} we get:

\small
\begin{equation}
\mytagd \label{eq::dGravity_dq}
\boxed{\frac{\partial \bm{\mathcal{F}}_{Gk}}{\partial \bm{q}} = \bm{\mathcal{M}}_k\sum_{\beta=1}^{k-1}\mathrm{Ad}_{\bm{g}_{\beta k}}^{-1}\mathrm{ad}_{\mathrm{Ad}_{\bm{g}_{\beta}}^{-1}\bm{\mathcal{G}}}\bm{S}_{\beta}}
\end{equation}
\normalsize

\noindent
\rule{\columnwidth}{0.4pt}
\uline{Contribution of Coriolis Force}

The Coriolis component of $\bm{\mathcal{F}}_k$ is given by:

\small
\begin{equation}
\mytagd \label{eq::CoriolisForce}
\bm{\mathcal{F}}_{Ck} = \mathrm{ad}_{\bm{\eta}_k}^*\bm{\mathcal{M}}_k\bm{\eta}_k+ \bm{\mathcal{M}}_k\bm{\varphi}_k
\end{equation}
\normalsize
where,
\small
\begin{equation}
\mytagd \label{eq::VelocityTwist}
\bm{\eta}_k = \sum_{l=1}^{k-1}\mathrm{Ad}_{\bm{g}_{l k}}^{-1}\bm{S}_l\dot{\bm{q}}
\end{equation}
\normalsize
and
\small
\begin{equation}
\mytagd \label{eq::CoriolisForce_phi}
\bm{\varphi}_k = \sum_{l=1}^{k-1}\mathrm{Ad}_{\bm{g}_{l k}}^{-1}\left(\mathrm{ad}_{\bm{\eta}_l}\bm{S}_l\dot{\bm{q}}+\dot{\bm{S}}_l\dot{\bm{q}}\right)
\end{equation}
\normalsize

Using \eqref{I:ad* to ad_bar*} we write,

\small
\begin{equation}
\mytagd \label{eq::dCoriolis_dq_start}
\frac{\partial \bm{\mathcal{F}}_{Ck}}{\partial \bm{q}} = \left(\overline{\mathrm{ad}}_{\bm{\mathcal{M}}_k\bm{\eta}_k}^*+\mathrm{ad}_{\bm{\eta}_k}^*\bm{\mathcal{M}}_k\right)\frac{\partial \bm{\eta}_k}{\partial \bm{q}} + \bm{\mathcal{M}}_k\frac{\partial \bm{\varphi}_k}{\partial \bm{q}}
\end{equation}
\normalsize

To compute this, we need to derive the partial derivatives of $\bm{\eta}_k$ and $\bm{\varphi}_k$. From \eqref{eq::VelocityTwist} we get,

\small
\begin{equation}
\mytagd \label{eq::dVelocity_dq_start}
\frac{\partial \bm{\eta}_k}{\partial \bm{q}}  = \sum_{l=1}^{k-1}\frac{\partial \mathrm{Ad}_{\bm{g}_{l k}}^{-1}}{\partial \bm{q}}\bm{S}_l\dot{\bm{q}}+\mathrm{Ad}_{\bm{g}_{l k}}^{-1}\frac{\partial \bm{S}_l}{\partial \bm{q}}\dot{\bm{q}}
\end{equation}
\normalsize

The second term is given by \eqref{eq::dSdq_qd_end}, while the first can be derived similarly to \eqref{eq::dGravity_dq}. We get,
\small
\begin{equation}
\mytagd \label{eq::dAdginv_lk_dq_start}
\begin{split}
\sum_{l=1}^{k-1}\frac{\partial \mathrm{Ad}_{\bm{g}_{l k}}^{-1}}{\partial \bm{q}}\bm{S}_l\dot{\bm{q}}=&\sum_{l=1}^{k-1}\sum_{\beta=l}^{k-1}\mathrm{Ad}_{\bm{g}_{\beta k}}^{-1}\mathrm{ad}_{\mathrm{Ad}_{\bm{g}_{l \beta}}^{-1}\bm{S}_l\dot{\bm{q}}}\bm{S}_{\beta}\\
=&\sum_{l=1}^{k-1}\sum_{\beta=l}^{k-1}\mathrm{ad}_{\mathrm{Ad}_{\bm{g}_{l k}}^{-1}\bm{S}_l\dot{\bm{q}}}\mathrm{Ad}_{\bm{g}_{\beta k}}^{-1}\bm{S}_{\beta}
\end{split}
\end{equation}
\normalsize

Based on the summation identity \eqref{I:Summation 1}, \eqref{I:ad_v*Ad_g} we can rewrite this as:

\small
\begin{equation}
\mytagd \label{eq::dAdginv_lk_dq_end}
\sum_{l=1}^{k-1}\frac{\partial \mathrm{Ad}_{\bm{g}_{l k}}^{-1}}{\partial \bm{q}}\bm{S}_l\dot{\bm{q}}=\sum_{\beta=1}^{k-1}\mathrm{Ad}_{\bm{g}_{\beta k}}^{-1}\mathrm{ad}_{\bm{\eta}_\beta^+}\bm{S}_{\beta}
\end{equation}
\normalsize
where, $\bm{\eta}_\beta^+ = \sum_{l=1}^{\beta}\mathrm{Ad}_{\bm{g}_{l \beta}}^{-1}\bm{S}_l\dot{\bm{q}} = \bm{\eta}_\beta+\bm{S}_\beta\dot{\bm{q}}$.

Substituting this into \eqref{eq::dVelocity_dq_start} we get:

\small
\begin{equation}
\mytagd \label{eq::dVelocity_dq_end}
\frac{\partial \bm{\eta}_k}{\partial \bm{q}}  =  \sum_{\beta=1}^{k-1}\mathrm{Ad}_{\bm{g}_{\beta k}}^{-1}\bm{R}_\beta
\end{equation}
\normalsize
where,
\small
\begin{equation}
\mytagd \label{eq::Rbeta_app}
\bm{R}_\beta=\mathrm{ad}_{\bm{\eta}_\beta^+}\bm{S}_{\beta}+\frac{\partial \bm{S}_\beta}{\partial \bm{q}}\dot{\bm{q}}
\end{equation}
\normalsize

From \eqref{eq::CoriolisForce_phi} we get,

\small
\begin{equation}
\mytagd \label{eq::dCentripetal_dq}
\begin{split}
\frac{\partial \bm{\varphi}_k}{\partial \bm{q}} =& \sum_{l=1}^{k-1}\Biggl(\frac{\partial \mathrm{Ad}_{\bm{g}_{l k}}^{-1}}{\partial \bm{q}}\left(\mathrm{ad}_{\bm{\eta}_l}\bm{S}_l\dot{\bm{q}}+\dot{\bm{S}}_l\dot{\bm{q}}\right)-\mathrm{Ad}_{\bm{g}_{l k}}^{-1}\mathrm{ad}_{\bm{S}_l\dot{\bm{q}}}\frac{\partial \bm{\eta}_l}{\partial \bm{q}}\\
+&\mathrm{Ad}_{\bm{g}_{l k}}^{-1}\left(\mathrm{ad}_{\bm{\eta}_l}\frac{\partial \bm{S}_l}{\partial \bm{q}}\dot{\bm{q}}+\frac{\partial \dot{\bm{S}}_l}{\partial \bm{q}}\dot{\bm{q}}\right)\Biggr)
\end{split}
\end{equation}
\normalsize

The first term has the same form of \eqref{eq::dAdginv_lk_dq_end}. Hence, we get:

\small
\begin{equation}
\mytagd \label{eq::dCentripetal_dq_1}
\sum_{l=1}^{k-1}\frac{\partial \mathrm{Ad}_{\bm{g}_{l k}}^{-1}}{\partial \bm{q}}\left(\mathrm{ad}_{\bm{\eta}_l}\bm{S}_l\dot{\bm{q}}+\dot{\bm{S}}_l\dot{\bm{q}}\right)=\sum_{\beta=1}^{k-1}\mathrm{Ad}_{\bm{g}_{\beta k}}^{-1}\mathrm{ad}_{\bm{\varphi}_\beta^+}\bm{S}_{\beta}
\end{equation}
\normalsize
where, $\bm{\varphi}_\beta^+ = \bm{\varphi}_\beta+\mathrm{ad}_{\bm{\eta}_\beta}\bm{S}_\beta\dot{\bm{q}}+\dot{\bm{S}}_\beta\dot{\bm{q}}$.

The second term of \eqref{eq::dCentripetal_dq} can be expanded using equation \eqref{eq::dVelocity_dq_end} and the identities \eqref{I:Ad_g*ad_v} and \eqref{I:Summation 2} as follows:

\small
\begin{equation}
\mytagd \label{eq::dCentripetal_dq_2}
\begin{split}    
-\sum_{l=1}^{k-1}\mathrm{Ad}_{\bm{g}_{l k}}^{-1}\mathrm{ad}_{\bm{S}_l\dot{\bm{q}}}\frac{\partial \bm{\eta}_l}{\partial \bm{q}} &= -\sum_{l=1}^{k-1}\sum_{\beta=1}^{l-1} \mathrm{ad}_{\mathrm{Ad}_{\bm{g}_{l k}}^{-1}\bm{S}_l\dot{\bm{q}}}\mathrm{Ad}_{\bm{g}_{\beta k}}^{-1}\bm{R}_\beta\\
=&-\sum_{\beta=1}^{k-2} \left(\sum_{l=\beta+1}^{k-1}\mathrm{ad}_{\mathrm{Ad}_{\bm{g}_{l k}}^{-1}\bm{S}_l\dot{\bm{q}}}\right)\mathrm{Ad}_{\bm{g}_{\beta k}}^{-1}\bm{R}_\beta\\
=&-\sum_{\beta=1}^{k-2} \mathrm{ad}_{\bm{\eta}_k-\mathrm{Ad}_{\bm{g}_{\beta k}}^{-1}\bm{\eta}_\beta^+}\mathrm{Ad}_{\bm{g}_{\beta k}}^{-1}\bm{R}_\beta\\
=&\sum_{\beta=1}^{k-1} \mathrm{Ad}_{\bm{g}_{\beta k}}^{-1}\mathrm{ad}_{\bm{\eta}_\beta^+-\mathrm{Ad}_{\bm{g}_{\beta k}}\bm{\eta}_k}\bm{R}_\beta
\end{split}
\end{equation}
\normalsize
Note that we used $\bm{\eta}_{k-1}^+=\mathrm{Ad}_{\bm{g}_{k-1 k}}\bm{\eta}_k$ for changing the summation limit.

The analytical formulas of the last of \eqref{eq::dCentripetal_dq} term are provided by \eqref{eq::dSdq_qd_end} and \eqref{eq::dSddq_dq_end}. Substituting \eqref{eq::dCentripetal_dq_1} and \eqref{eq::dCentripetal_dq_2} in \eqref{eq::dCentripetal_dq} we get:


\small
\begin{equation}
\mytagd \label{eq::dCentripetal_dq_end}
\begin{split}
\frac{\partial \bm{\varphi}_k}{\partial \bm{q}} = \sum_{\beta=1}^{k-1}\mathrm{Ad}_{\bm{g}_{\beta k}}^{-1}\Biggl(\mathrm{ad}_{\bm{\varphi}_\beta^+}\bm{S}_{\beta}+\mathrm{ad}_{\bm{\eta}_\beta^+-\mathrm{Ad}_{\bm{g}_{\beta k}}\bm{\eta}_k}\bm{R}_\beta \\
+\mathrm{ad}_{\bm{\eta}_\beta}\frac{\partial \bm{S}_\beta}{\partial \bm{q}}\dot{\bm{q}}+\frac{\partial \dot{\bm{S}}_\beta}{\partial \bm{q}}\dot{\bm{q}}\Biggr)
\end{split}
\end{equation}
\normalsize

Substituting \eqref{eq::dCentripetal_dq_end} and \eqref{eq::dVelocity_dq_end} into \eqref{eq::dCoriolis_dq_start}, we get the final form of the derivative of Coriolis force:

\small
\begin{empheq}[box=\fbox]{gather}
\mytagd \label{eq::dCoriolis_dq_end}
\begin{gathered}
\frac{\partial \bm{\mathcal{F}}_{Ck}}{\partial \bm{q}} = \left(\overline{\mathrm{ad}}_{\bm{\mathcal{M}}_k\bm{\eta}_k}^*+\mathrm{ad}_{\bm{\eta}_k}^*\bm{\mathcal{M}}_k\right)\sum_{\beta=1}^{k-1}\mathrm{Ad}_{\bm{g}_{\beta k}}^{-1}\bm{R}_\beta \\
+ \bm{\mathcal{M}}_k\sum_{\beta=1}^{k-1}\mathrm{Ad}_{\bm{g}_{\beta k}}^{-1}\Biggl(\mathrm{ad}_{\bm{\varphi}_\beta^+}\bm{S}_{\beta}+\mathrm{ad}_{\bm{\eta}_\beta^+-\mathrm{Ad}_{\bm{g}_{\beta k}}\bm{\eta}_k}\bm{R}_\beta \\
+ \mathrm{ad}_{\bm{\eta}_\beta}\frac{\partial \bm{S}_\beta}{\partial \bm{q}}\dot{\bm{q}}+\frac{\partial \dot{\bm{S}}_\beta}{\partial \bm{q}}\dot{\bm{q}}\Biggr)
\end{gathered}
\end{empheq}
\normalsize

\noindent
\rule{\columnwidth}{0.4pt}
\uline{Contribution of Inertial Force}

The inertial component of $\bm{\mathcal{F}}_k$ is given by:

\small
\begin{equation}
\mytagd \label{eq::Inertialforce}
\bm{\mathcal{F}}_{Ik} = \bm{\mathcal{M}}_k\bm{\gamma}_k =\bm{\mathcal{M}}_k\sum_{l=1}^{k-1}\mathrm{Ad}_{\bm{g}_{l k}}^{-1}\bm{S}_l\ddot{\bm{q}}
\end{equation}
\normalsize

It is easy to see that the derivative of this term follows the same form of \eqref{eq::dVelocity_dq_end}. We get:

\small
\begin{equation}
\mytagd \label{eq::dInertial_dq}
\boxed{
\frac{\partial \bm{\mathcal{F}}_{Ik}}{\partial \bm{q}}  = \bm{\mathcal{M}}_k\sum_{\beta=1}^{k-1}\mathrm{Ad}_{\bm{g}_{\beta k}}^{-1}\left(\mathrm{ad}_{\bm{\gamma}_\beta^+}\bm{S}_{\beta}+\frac{\partial \bm{S}_\beta}{\partial \bm{q}}\ddot{\bm{q}}\right)}
\end{equation}
\normalsize
where, $\bm{\gamma}_\beta^+ = \sum_{l=1}^{\beta}\mathrm{Ad}_{\bm{g}_{l \beta}}^{-1}\bm{S}_l\ddot{\bm{q}} = \bm{\gamma}_\beta+\bm{S}_\beta\ddot{\bm{q}}$.

\noindent
\rule{\columnwidth}{0.4pt}
\uline{Combining all Contributions}

Combining partial derivatives of gravitational force \eqref{eq::dGravity_dq}, Coriolis force \eqref{eq::dCoriolis_dq_end}, and inertial force \eqref{eq::dInertial_dq}, we get:

\small
\begin{equation}
\mytagd \label{eq::dFk_dq_full_app}
\frac{\partial \bm{\mathcal{F}}_k}{\partial \bm{q}} =\bm{\mathcal{N}}_k\sum_{\beta=1}^{k-1}\mathrm{Ad}_{\bm{g}_{\beta k}}^{-1}\bm{R}_\beta+\bm{\mathcal{M}}_k\sum_{\beta=1}^{k-1}\mathrm{Ad}_{\bm{g}_{\beta k}}^{-1}\bm{Q}_\beta
\end{equation}
\normalsize
where,
\small
\begin{subequations}
\begin{align}
\bm{\mathcal{N}}_k =& \overline{\mathrm{ad}}_{\bm{\mathcal{M}}_k\bm{\eta}_k}^*+\mathrm{ad}_{\bm{\eta}_k}^*\bm{\mathcal{M}}_k -\bm{\mathcal{M}}_k\mathrm{ad}_{\bm{\eta}_k} \mytagd \label{eq::N_k_app} \\
\bm{Q}_\beta=&\mathrm{ad}_{\dot{\bm{\eta}}_\beta^+}\bm{S}_{\beta}+\mathrm{ad}_{\bm{\eta}_\beta^+}\bm{R}_\beta \mytagd \label{eq::Q_beta_app}\\
&+\mathrm{ad}_{\bm{\eta}_\beta}\frac{\partial \bm{S}_\beta}{\partial \bm{q}}\dot{\bm{q}}+\frac{\partial \dot{\bm{S}}_\beta}{\partial \bm{q}}\dot{\bm{q}}+\frac{\partial \bm{S}_\beta}{\partial \bm{q}}\ddot{\bm{q}}-\mathrm{ad}_{\mathrm{Ad}_{\bm{g}_{\beta}}^{-1}\bm{\mathcal{G}}}\bm{S}_{\beta} \nonumber
\end{align}
\end{subequations}
\normalsize

In the backward pass, the partial derivatives of $\bm{\mathcal{F}}_k$ are transformed to the frames of each ``virtual joint'' using coAdjoint maps. Using \eqref{eq::dFk_dq_full_app}, $\mathrm{Ad}_{\bm{g}_{\beta k}}^{-1} = \mathrm{Ad}_{\bm{g}_{\alpha k}}^{-1}\mathrm{Ad}_{\bm{g}_{\alpha \beta}}$, and summation identity \eqref{I:Summation 4} we get:

\small
\begin{equation}
\mytagd \label{eq::dID_dq_III_app}
\begin{split}
\sum_{k=\alpha+1}^{n_p} & \mathrm{Ad}_{\bm{g}_{\alpha k}}^*\frac{\partial \bm{\mathcal{F}}_k}{\partial \bm{q}}=\sum_{k=\alpha+1}^{n_p}\sum_{\beta=1}^{k-1}\mathrm{Ad}_{\bm{g}_{\alpha k}}^*\Biggl(\bm{\mathcal{N}}_k\mathrm{Ad}_{\bm{g}_{\alpha k}}^{-1}\mathrm{Ad}_{\bm{g}_{\alpha \beta}}\bm{R}_\beta\\
+&\bm{\mathcal{M}}_k\mathrm{Ad}_{\bm{g}_{\alpha k}}^{-1}\mathrm{Ad}_{\bm{g}_{\alpha \beta}}\bm{Q}_\beta\Biggr)\\
=&\bm{\mathcal{N}}_\alpha^C\sum_{\beta=1}^{\alpha-1}\mathrm{Ad}_{\bm{g}_{\alpha \beta}}\bm{R}_\beta+\bm{\mathcal{M}}_\alpha^C\sum_{\beta=1}^{\alpha-1}\mathrm{Ad}_{\bm{g}_{\alpha \beta}}\bm{Q}_\beta\\
+&\sum_{\beta=\alpha}^{n_p-1}\mathrm{Ad}_{\bm{g}_{\alpha \beta}}^*\left(\bm{\mathcal{N}}_\beta^C\bm{R}_\beta+\bm{\mathcal{M}}_\beta^C\bm{Q}_\beta\right)
\end{split}
\end{equation}
\normalsize

We can simplify \eqref{eq::dID_dq_III_app} into:
\small
\begin{equation}
\mytagd \label{eq::dID_dq_III_app2}
\boxed{
\sum_{k=\alpha+1}^{n_p} \mathrm{Ad}_{\bm{g}_{\alpha k}}^*\frac{\partial \bm{\mathcal{F}}_k}{\partial \bm{q}}
=\bm{\mathcal{N}}_\alpha^C\bm{R}_{\alpha}^B+\bm{\mathcal{M}}_\alpha^C\bm{Q}_{\alpha}^B+\bm{U}_\alpha^S}
\end{equation}
\normalsize
where, 
\small
\begin{subequations}
\begin{align}
\bm{R}_\alpha^B =& \sum_{\beta=1}^{\alpha-1}\mathrm{Ad}_{\bm{g}_{\alpha \beta}}\bm{R}_\beta \mytagd \label{eq::RB_app1} \\
\bm{Q}_\alpha^B=& \sum_{\beta=1}^{\alpha-1}\mathrm{Ad}_{\bm{g}_{\alpha \beta}}\bm{Q}_\beta  \mytagd \label{eq::QB_app1} \\
\bm{\mathcal{N}}_\alpha^C =& \sum_{k=\alpha+1}^{n_p}\mathrm{Ad}_{\bm{g}_{\alpha k}}^*\bm{\mathcal{N}}_k\mathrm{Ad}_{\bm{g}_{\alpha k}}^{-1} \mytagd \label{eq::NC_app1} \\
\bm{\mathcal{M}}_\alpha^C=& \sum_{k=\alpha+1}^{n_p}\mathrm{Ad}_{\bm{g}_{\alpha k}}^*\bm{\mathcal{M}}_k\mathrm{Ad}_{\bm{g}_{\alpha k}}^{-1} \mytagd \label{eq::MC_app1} \\
\bm{U}_\alpha^S=& \sum_{\beta=\alpha}^{n_p-1}\mathrm{Ad}_{\bm{g}_{\alpha \beta}}^*\left(\bm{\mathcal{N}}_\beta^C\bm{R}_\beta+\bm{\mathcal{M}}_\beta^C\bm{Q}_\beta\right)  \mytagd \label{eq::OS_app1}
\end{align}
\end{subequations}
\normalsize

$\bm{R}_{\alpha}^B$ and $\bm{Q}_{\alpha}^B$ can be recursively computed during the forward pass, while $\bm{\mathcal{N}}_\alpha^C$, $\bm{\mathcal{M}}_\alpha^C$, and $\bm{U}_\alpha^S$ can be computed during the backward pass as follows:

\small
\begin{subequations}
\begin{align}
\bm{R}_\alpha^B =& \mathrm{Ad}_{\bm{g}_{\alpha-1 \alpha }}^{-1}(\bm{R}_{\alpha-1}+\bm{R}_{\alpha-1}^B) \mytagd \label{eq::RB_app2} \\
\bm{Q}_\alpha^B=& \mathrm{Ad}_{\bm{g}_{\alpha-1 \alpha}}^{-1}(\bm{Q}_{\alpha-1}+\bm{Q}_{\alpha-1}^B)  \mytagd \label{eq::QB_app2} \\
\bm{\mathcal{N}}_\alpha^C =& \mathrm{Ad}_{\bm{g}_{\alpha \alpha+1}}^*(\bm{\mathcal{N}}_{\alpha+1}+\bm{\mathcal{N}}_{\alpha+1}^C)\mathrm{Ad}_{\bm{g}_{\alpha \alpha+1}}^{-1} \mytagd \label{eq::NC_app2} \\
\bm{\mathcal{M}}_\alpha^C=& \mathrm{Ad}_{\bm{g}_{\alpha \alpha+1}}^*(\bm{\mathcal{M}}_{\alpha+1}+\bm{\mathcal{M}}_{\alpha+1}^C)\mathrm{Ad}_{\bm{g}_{\alpha \alpha+1}}^{-1} \mytagd \label{eq::MC_app2} \\
\bm{U}_\alpha^S=& \bm{\mathcal{N}}_\alpha^C\bm{R}_\alpha+\bm{\mathcal{M}}_\alpha^C\bm{Q}_\alpha+\mathrm{Ad}_{\bm{g}_{\alpha \alpha+1}}^*\bm{U}_{\alpha+1}^S  \mytagd \label{eq::US_app2}
\end{align}
\end{subequations}
\normalsize

\subsection*{Derivations in \ref{sec::dID_dqd}}

The partial derivative of $ID_\alpha$ with respect to $\dot{\bm{q}}$ is given by:

\small
\begin{equation}
\mytagd \label{eq::dtau_dqd_end_app}
\frac{\partial ID_\alpha}{\partial \dot{\bm{q}}}=\bm{S}_\alpha^T\left(\bm{\mathcal{N}}_\alpha^C\bm{S}_{\alpha}^B+\bm{\mathcal{M}}_\alpha^C\bm{Y}_{\alpha}^B+\bm{V}_\alpha^S\right)
\end{equation}
\normalsize

Only the Coriolis force is a function of $\dot{\bm{q}}$. Therefore, $\bm{\mathcal{N}}_\alpha^C\bm{S}_{\alpha}^B+\bm{\mathcal{M}}_\alpha^C\bm{Y}_{\alpha}^B+\bm{V}_\alpha^S$ in \eqref{eq::dtau_dqd_end_app} originates from the partial derivative of $\bm{\mathcal{F}}_{Ck}$ with respect to $\dot{\bm{q}}$ transformed to the frame of $\alpha$. From \eqref{eq::CoriolisForce} we get:

\small
\begin{equation}
\mytagd \label{eq::dCoriolis_dqd_start}
\frac{\partial \bm{\mathcal{F}}_k}{\partial \dot{\bm{q}}} = \left(\overline{\mathrm{ad}}_{\bm{\mathcal{M}}_k\bm{\eta}_k}^*+\mathrm{ad}_{\bm{\eta}_k}^*\bm{\mathcal{M}}_k\right)\frac{\partial \bm{\eta}_k}{\partial \dot{\bm{q}}} + \bm{\mathcal{M}}_k\frac{\partial \bm{\varphi}_k}{\partial \dot{\bm{q}}}
\end{equation}
\normalsize

From \eqref{eq::VelocityTwist} we get:

\small
\begin{equation}
\mytagd \label{eq::dVelocityTwist_dqd}
\frac{\partial \bm{\eta}_k}{\partial \dot{\bm{q}}} =\sum_{l=1}^{k-1}\mathrm{Ad}_{\bm{g}_{l k}}^{-1}\bm{S}_l
\end{equation}
\normalsize

From \eqref{eq::CoriolisForce_phi} we get:

\small
\begin{equation}
\mytagd \label{eq::dCoriolis_phi_dqd_start}
\frac{\partial \bm{\varphi}_k}{\partial \dot{\bm{q}}} = \sum_{l=1}^{k-1}\mathrm{Ad}_{\bm{g}_{l k}}^{-1}\left(-\mathrm{ad}_{\bm{S}_l\dot{\bm{q}}}\frac{\partial \bm{\eta}_l}{\partial \dot{\bm{q}}}+\mathrm{ad}_{\bm{\eta}_l}\bm{S}_l+\frac{\partial \dot{\bm{S}}_l}{\partial \bm{q}}\dot{\bm{q}}+\dot{\bm{S}}_l\right)
\end{equation}
\normalsize

Using similar mathematical operations like that of \eqref{eq::dCentripetal_dq_2} we get:

\small
\begin{equation}
\mytagd \label{eq::dCoriolis_phi_dqd_end}
\begin{split}
\frac{\partial \bm{\varphi}_k}{\partial \dot{\bm{q}}} =& \sum_{\beta=1}^{k-1}\mathrm{Ad}_{\bm{g}_{\beta k}}^{-1}\Biggl(\mathrm{ad}_{\bm{\eta}_\beta^+-\mathrm{Ad}_{\bm{g}_{\beta k}}\bm{\eta}_k}\bm{S}_\beta+\mathrm{ad}_{\bm{\eta}_\beta}\bm{S}_\beta \\
+& \frac{\partial \bm{S}_\beta}{\partial \bm{q}}\dot{\bm{q}}+\dot{\bm{S}}_\beta\Biggr)
\end{split}
\end{equation}
\normalsize

Substituting \eqref{eq::dCoriolis_phi_dqd_end} into \eqref{eq::dCoriolis_dqd_start} we get:

\small
\begin{equation}
\mytagd \label{eq::dCoriolis_dqd_end}
\begin{split}
\frac{\partial \bm{\mathcal{F}}_k}{\partial \dot{\bm{q}}} =&\left(\overline{\mathrm{ad}}_{\bm{\mathcal{M}}_k\bm{\eta}_k}^*+\mathrm{ad}_{\bm{\eta}_k}^*\bm{\mathcal{M}}_k\right)\sum_{\beta=1}^{k-1}\mathrm{Ad}_{\bm{g}_{\beta k}}^{-1}\bm{S}_\beta\\ 
+& \bm{\mathcal{M}}_k\sum_{\beta=1}^{k-1}\mathrm{Ad}_{\bm{g}_{\beta k}}^{-1}\Biggl(\mathrm{ad}_{\bm{\eta}_\beta^+-\mathrm{Ad}_{\bm{g}_{\beta k}}\bm{\eta}_k}\bm{S}_\beta+\mathrm{ad}_{\bm{\eta}_\beta}\bm{S}_\beta\\
+&\frac{\partial \bm{S}_\beta}{\partial \bm{q}}\dot{\bm{q}}+\dot{\bm{S}}_\beta\Biggr)
\end{split}
\end{equation}
\normalsize

By rearranging this, we get:

\small
\begin{equation}
\mytagd \label{eq::dFk_dqd_full_app}
\frac{\partial \bm{\mathcal{F}}_k}{\partial \dot{\bm{q}}} =\bm{\mathcal{N}}_k\sum_{\beta=1}^{k-1}\mathrm{Ad}_{\bm{g}_{\beta k}}^{-1}\bm{S}_\beta+\bm{\mathcal{M}}_k\sum_{\beta=1}^{k-1}\mathrm{Ad}_{\bm{g}_{\beta k}}^{-1}\bm{Y}_\beta
\end{equation}
\normalsize
where,

\small
\begin{equation}
\mytagd \label{eq::Y_beta_app}
\bm{Y}_\beta =  \mathrm{ad}_{\bm{\eta}_\beta^+}\bm{S}_\beta+\mathrm{ad}_{\bm{\eta}_\beta}\bm{S}_\beta+\frac{\partial \bm{S}_\beta}{\partial \bm{q}}\dot{\bm{q}}+\dot{\bm{S}}_\beta 
\end{equation}
\normalsize

Transforming the partial derivatives of $\bm{\mathcal{F}}_k$ at all points $k>\alpha$ to the frame of $\alpha$, similar to \eqref{eq::dID_dq_III_app}, we get:
\small
\begin{equation}
\mytagd \label{eq::dID_dqd_III_app}
\boxed{
\sum_{k=\alpha+1}^{n_p} \mathrm{Ad}_{\bm{g}_{\alpha k}}^*\frac{\partial \bm{\mathcal{F}}_k}{\partial \dot{\bm{q}}}
=\bm{\mathcal{N}}_\alpha^C\bm{S}_{\alpha}^B+\bm{\mathcal{M}}_\alpha^C\bm{Y}_{\alpha}^B+\bm{V}_\alpha^S}
\end{equation}
\normalsize

$\bm{S}_{\alpha}^B$ and $\bm{Y}_{\alpha}^B$ can be recursively computed during the forward pass, while $\bm{V}_\alpha^S$ can be computed during the backward pass as follows:

\small
\begin{subequations}
\begin{align}
\bm{S}_\alpha^B =& \mathrm{Ad}_{\bm{g}_{\alpha-1 \alpha }}^{-1}(\bm{S}_{\alpha-1}+\bm{S}_{\alpha-1}^B) \mytagd \label{eq::SB_app2} \\
\bm{Y}_\alpha^B=& \mathrm{Ad}_{\bm{g}_{\alpha-1 \alpha}}^{-1}(\bm{Y}_{\alpha-1}+\bm{Y}_{\alpha-1}^B)  \mytagd \label{eq::YB_app2} \\
\bm{V}_\alpha^S=& \bm{\mathcal{N}}_\alpha^C\bm{S}_\alpha+\bm{\mathcal{M}}_\alpha^C\bm{Y}_\alpha+\mathrm{Ad}_{\bm{g}_{\alpha \alpha+1}}^*\bm{V}_{\alpha+1}^S  \mytagd \label{eq::VS_app2}
\end{align}
\end{subequations}

\normalsize
\subsection*{Derivations in \ref{sec::dID_dqdd}}
The partial derivatives of $ID_\alpha$ with respect to $\ddot{\bm{q}}$ is given by:

\small
\begin{equation}
\mytagd \label{eq::dtau_dqdd_end_app}
\frac{\partial ID_\alpha}{\partial \ddot{\bm{q}}} =\bm{S}_\alpha^T\left(\bm{\mathcal{M}}_\alpha^C\bm{S}_{\alpha}^B+\bm{W}_\alpha^S\right)
\end{equation}
\normalsize

Since, only the inertial force is a function of $\ddot{\bm{q}}$, $\bm{\mathcal{M}}_\alpha^C\bm{S}_{\alpha}^B+\bm{W}_\alpha^S$ in \eqref{eq::dtau_dqdd_end_app} originates from the partial derivative of $\bm{\mathcal{F}}_{Ik}$ with respect to $\ddot{\bm{q}}$ transformed to the frame of $\alpha$. From \eqref{eq::Inertialforce} we get:

\small
\begin{equation}
\mytagd \label{eq::dInertial_dqdd_start}
\frac{\partial \bm{\mathcal{F}}_k}{\partial \ddot{\bm{q}}} = \bm{\mathcal{M}}_k\sum_{\beta=1}^{k-1}\mathrm{Ad}_{\bm{g}_{\beta k}}^{-1}\bm{S}_\beta
\end{equation}
\normalsize

Transforming the partial derivatives of $\bm{\mathcal{F}}_k$ at all points $k>\alpha$ to the frame of $\alpha$, similar to \eqref{eq::dID_dq_III_app}, we get:
\small
\begin{equation}
\mytagd \label{eq::dID_dqdd_III_app}
\boxed{
\sum_{k=\alpha+1}^{n_p} \mathrm{Ad}_{\bm{g}_{\alpha k}}^*\frac{\partial \bm{\mathcal{F}}_k}{\partial \ddot{\bm{q}}}
=\bm{\mathcal{M}}_\alpha^C\bm{S}_{\alpha}^B+\bm{W}_\alpha^S}
\end{equation}
\normalsize
where, $\bm{W}_\alpha^S$ is computed recursively during the backward pass.
\begin{equation}
\mytagd \label{eq::WS_app2}
    \bm{W}_\alpha^S = \bm{\mathcal{M}}_\alpha^C\bm{S}_\alpha+\mathrm{Ad}_{\bm{g}_{\alpha \alpha+1}}^*\bm{W}_{\alpha+1}^S
\end{equation}

\subsection*{Derivations in \ref{sec::dtau_dq}}

The generalized actuation force for tendon-like actuators is given by:
\small
\begin{equation}
\mytagd \label{eq::TendonActuator}
\bm{B}\bm{u} = \sum_{i=1}^{n_p}w_i{\bm{\phi}_\xi}_i^T\sum_{k=1}^{n_{a}}\begin{pmatrix} \tilde{\bm{d}}_{ik}\bm{t}_{ik} \\ \bm{t}_{ik} \end{pmatrix}u_k
\end{equation}
\normalsize
where, $\bm{d}_{ik}(X_i)=[0\;y_{ik}\;z_{ik}]^T$ is the local coordinates of the $k^{th}$ tendon and $\bm{t}_{ik}$ is the the unit tangent vector. We have,

\small
\begin{subequations}
\begin{align}
\bm{t}_{ik} =&\frac{\bm{\mathcal{T}}_{ik}}{\|\bm{\mathcal{T}}_{ik}\|} \mytagd \label{eq::unittangent} \\
\bm{\mathcal{T}}_{ik} =& \widehat{\bm{\xi}}_i\begin{pmatrix}
\bm{d}_{ik}\\
1
\end{pmatrix}+\begin{pmatrix}
\bm{d}_{ik}' \\
1
\end{pmatrix} \mytagd \label{eq::tangent}
\end{align}
\end{subequations}
\normalsize

Using the equation of strain parameterization \eqref{eq::strainparameterization} and \eqref{eq::unittangent} we get:
\small
\begin{equation}
\mytagd  \label{eq::dB_dq_u}
\frac{\partial \bm{B}}{\partial \bm{q}}\bm{u}=\sum_{i=1}^{n_p}w_i{\bm{\phi}_\xi}_i^T\sum_{k=1}^{n_{a}}\frac{u_k}{\|\bm{\mathcal{T}}_{ik}\|}\begin{pmatrix} \tilde{\bm{d}}_{ik}{\tilde{\bm{t}}_{ik}}^2\tilde{\bm{d}}_{ik} &  -\tilde{\bm{d}}_{ik}{\tilde{\bm{t}}_{ik}}^2 \\ {\tilde{\bm{t}}_{ik}}^2\tilde{\bm{d}}_{ik} &  -{\tilde{\bm{t}}_{ik}}^2 \end{pmatrix}{\bm{\phi}_\xi}_i
\end{equation}
\normalsize

Generalized stiffness and damping matrices of a soft body are given by:
\small
\begin{subequations}
\begin{align}
&\bm{K} =   \sum_{i=1}^{n_p}w_i{\bm{\phi}_\xi}_i^T \bm{\Sigma}(X_i) {\bm{\phi}_\xi}_i \mytagd \label{eq::Stiffness}  \\
&\bm{D} =   \sum_{i=1}^{n_p}w_i{\bm{\phi}_\xi}_i^T \bm{\Upsilon}(X_i) {\bm{\phi}_\xi}_i \mytagd \label{eq::Damping}
\end{align}
\end{subequations}
\normalsize
where $ \bm{\Sigma}(X_i) = \text{diag}(G J_{x_i},\; E J_{y_i},\; E J_{z_i},\; E A_i, \; G A_i,\\ G A_i)  \in \mathbb{R}^{6 \times 6} $ is the screw elasticity matrix, $ E $ being the Young modulus and $ G $ the shear modulus and $ \bm{\Upsilon}(X_i) =\upsilon \text{diag}(J_{x_i}, 3 J_{y_i}, 3 J_{z_i}, 3 A_i, A_i, A_i)   \in \mathbb{R}^{6 \times 6} $ is the screw damping matrix, $ \upsilon $ being the  material damping.
\subsection*{Derivations in \ref{sec::JointCoordinateControlled}}
For a rigid joint, the generalized actuation force is given by:
\small
\begin{equation}
\mytagd  \label{eq::RigidJoint_Actuation}
\bm{B}\bm{u}=\sum_{k}\bm{\Phi}_k^T \bm{T}_k^T\mathrm{Ad}_{\exp\left(\widehat{\bm{\xi}_k}\right)}^*\bm{\Phi}_k\bm{u}_k
\end{equation}
\normalsize
where $k$ is the index of actuated rigid joints. 

The partial derivative of \eqref{eq::RigidJoint_Actuation} term with respect to $\bm{q}$ is given by:

\small
\begin{equation}
\mytagd \label{eq::dJointB_dq}
\begin{split}
\frac{\partial \bm{B}}{\partial \bm{q}}\bm{u} =& \sum_{k}\Biggl(\frac{1}{\theta}\bm{\Phi}_{k}^T\sum_{r=1}^{4} (-1)^r f_r^{'} (\theta)\mathrm{ad}_{\bm{\Omega}_k}^{*r}\bm{\mathcal{F}}_{k0}\bm{\Omega}_k^T\bm{I}_{\theta}\bm{\Phi}_{k} \\
+& \bm{\Phi}_{k}^T\sum_{r=1}^{4}(-1)^rf_r (\theta)\sum_{u=1}^{r}\mathrm{ad}_{\bm{\Omega_{k}}}^{*u-1}\overline{\mathrm{ad}}^*_{\mathrm{ad}_{\bm{\Omega_{k}}}^{*r-u}\bm{\mathcal{F}}_{k0}}\bm{\Phi}_{k}\\
+&\bm{S}_k^T\overline{\mathrm{ad}}^*_{\bm{\mathcal{F}}_{k0}}\bm{S}_k \Biggr)
\end{split}
\end{equation}
\normalsize
where, $\bm{\mathcal{F}}_{k0} = \mathrm{Ad}_{\exp\left(\widehat{\bm{\xi}_k}\right)}^*\bm{\Phi}_k\bm{u}_k$. Note that, for 1 DoF joints, $\bm{T} = \mathrm{Ad}_{\exp\left(\widehat{\bm{\xi}_k}\right)}$. Hence, for such joints, $\bm{B}$ is a constant, and its derivative is $\bm{0}$.
\subsection*{Derivations in \ref{sec::ClosedChainSystems}} 



\uline{Jacobian of Constraint Wrench}

Using identity \eqref{I:Adginv*_dot} and \eqref{I:ad* to ad_bar*} we derive partial derivative of the constraint force acting on $A$ with respect to $q_p$ as follows:
\small
\begin{equation}
\mytagd \label{eq::dFCL_dq_start}
\begin{split}
\frac{\partial \bm{\mathcal{F}}_{kA}}{\partial q_p} =& \frac{\partial \mathrm{Ad}_{\bm{g}_{kBA}}^{-*}}{\partial q_p}\bm{\Phi}_{\perp k}\bm{\lambda}_k\\ =& -\mathrm{ad}_{\bm{J}_{kBA,p}}^*\mathrm{Ad}_{\bm{g}_{kBA}}^{-*}\bm{\Phi}_{\perp k}\bm{\lambda}_k\\ =& -\overline{\mathrm{ad}}_{\mathrm{Ad}_{\bm{g}_{kBA}}^{-*}\bm{\Phi}_{\perp k}\bm{\lambda}_k}^*\left(\bm{J}_{kA,p}-\mathrm{Ad}_{\bm{g}_{kBA}}^{-1}\bm{J}_{kB,p}\right)
\end{split}
\end{equation}
\normalsize

Collecting all columns and using \eqref{I:ad_v*bar Ad_g}, we get:

\small
\begin{equation}
\mytagd \label{eq::dFCL_dq}
\begin{split}
\frac{\partial \bm{\mathcal{F}}_{kA}}{\partial \bm{q}} =& -\overline{\mathrm{ad}}_{\mathrm{Ad}_{\bm{g}_{kBA}}^{-*}\bm{\Phi}_{\perp k}\bm{\lambda}_k}^*\left(\bm{J}_{kA}-\mathrm{Ad}_{\bm{g}_{kBA}}^{-1}\bm{J}_{kB}\right)\\
=& -\mathrm{Ad}_{\bm{g}_{kBA}}^{-*}\overline{\mathrm{ad}}_{\bm{\Phi}_{\perp k}\bm{\lambda}_k}^*\left(\mathrm{Ad}_{\bm{g}_{kBA}}\bm{J}_{kA}-\bm{J}_{kB}\right)
\end{split}
\end{equation}
\normalsize

\noindent
\rule{\columnwidth}{0.8pt}
\uline{Constraint Jacobian with respect to $\bm{q}$}

The closed-chain constraint equation is given by:

\small
\begin{equation}
    \mytagd \label{eq::constrainteq_app}
    \bm{C}(\ddot{\bm{q}},\dot{\bm{q}},\bm{q}) = \bm{A}\ddot{\bm{q}} + \dot{\bm{A}}\dot{\bm{q}} + \frac{2}{T_B}\bm{A}\dot{\bm{q}} + \frac{1}{T_B^2}\bm{e} = 0
\end{equation}
\normalsize

The Jacobian of this equation with respect to $\bm{q}$ is given by:

\small
\begin{equation}
\mytagd \label{eq::dconstraint_dq_app}
    \frac{\partial \bm{C}}{\partial \bm{q}} = \frac{\partial \bm{A}}{\partial \bm{q}}\ddot{\bm{q}}+\frac{\partial \dot{\bm{A}}}{\partial \bm{q}}\dot{\bm{q}} + \frac{2}{T_B}\frac{\partial \bm{A}}{\partial \bm{q}}\dot{\bm{q}} + \frac{1}{T_B^2}\frac{\partial \bm{e}}{\partial \bm{q}}
\end{equation}
\normalsize

From the definition of the Pfaffian matrix and its derivative in equations \eqref{eq::PfaffianMatrix} and \eqref{eq::PfaffianMatrix_derivative}, we can rewrite the first three terms of equation \eqref{eq::dconstraint_dq_app} as follows:
\small
\begin{equation}
\mytagd \label{eq::DAdq}
\begin{aligned}
\frac{\partial \bm{A}}{\partial \bm{q}}\ddot{\bm{q}} + \frac{\partial \dot{\bm{A}}}{\partial \bm{q}}\dot{\bm{q}} + \frac{2}{T_B}\frac{\partial \bm{A}}{\partial \bm{q}}\dot{\bm{q}} =& \bigg\Vert_{k=1}^{n_{CL}}\bm{\Phi}_{\perp k}^T\Biggl(\frac{\partial \mathrm{Ad}_{\bm{g}_{kBA}}\dot{\bm{\eta}}_{kA}}{\partial \bm{q}} \\ -\frac{\partial \dot{\bm{\eta}}_{kB}}{\partial \bm{q}} 
+& \frac{2}{T_B}\Bigl(\frac{\partial \mathrm{Ad}_{\bm{g}_{kBA}}\bm{\eta}_{kA}}{\partial \bm{q}}-\frac{\partial \bm{\eta}_{kB}}{\partial \bm{q}}\Bigr)\Biggr)
\end{aligned}
\end{equation}
\normalsize

Equation \eqref{eq::dVelocity_dq_end} provides the Jacobian of velocity twist. From \eqref{eq::dInertial_dq} and \eqref{eq::dCentripetal_dq_end} we can infer the expression for the Jacobian of the acceleration twist.

\small
\begin{subequations}
\begin{align}
\frac{\partial \bm{\eta}_{k}}{\partial \bm{q}}=& \bm{R}_k^B \mytagd \label{eq::dVelocity_dq_CL}\\
\frac{\partial \dot{\bm{\eta}}_{k}}{\partial \bm{q}}=& \bm{L}_k^B-\mathrm{ad}_{\bm{\eta}_k}\bm{R}_k^B
\mytagd \label{eq::dAcceleration_dq_CL}
\end{align}
\end{subequations}
\normalsize
where $\bm{L}_k^B$ is $\bm{Q}_k^B$ \eqref{eq::QB} without the gravity component.

Using \eqref{I:Adg_dot}, the derivative of the Adjoint terms in \eqref{eq::DAdq} can be derived as:

\small
\begin{subequations}
\begin{align}
\frac{\partial \mathrm{Ad}_{\bm{g}_{kBA}}}{\partial \bm{q}}\dot{\bm{\eta}}_{kA}=& -\mathrm{ad}_{\mathrm{Ad}_{\bm{g}_{kBA}}\dot{\bm{\eta}}_{kA}}\left(\mathrm{Ad}_{\bm{g}_{kBA}}\bm{J}_{kA}-\bm{J}_{kB}\right) \mytagd \label{eq::dAdjetadot_dq_CL}\\
\frac{\partial \mathrm{Ad}_{\bm{g}_{kBA}}}{\partial \bm{q}}\bm{\eta}_{kA}=& -\mathrm{ad}_{\mathrm{Ad}_{\bm{g}_{kBA}}\bm{\eta}_{kA}}\left(\mathrm{Ad}_{\bm{g}_{kBA}}\bm{J}_{kA}-\bm{J}_{kB}\right)
\mytagd \label{eq::dAdjeta_dq_CL}
\end{align}
\end{subequations}
\normalsize

The final form of \eqref{eq::DAdq} can be obtained by combining \eqref{eq::dVelocity_dq_CL}, \eqref{eq::dAcceleration_dq_CL}, \eqref{eq::dAdjetadot_dq_CL}, and \eqref{eq::dAdjeta_dq_CL}.

\noindent
\rule{\columnwidth}{0.4pt}

The last term in \eqref{eq::dconstraint_dq_app} includes the Jacobian of the closed-chain kinematic error. Let $\bm{\epsilon} = \left(\log(\bm{g}_{kBA})\right)^\vee$. We have,

\small
\begin{equation}
\mytagd \label{eq:exp_eps}
\exp\left(\widehat{\bm{\epsilon}}\right) = \bm{g}_{kBA}
\end{equation}
\normalsize

Using identities \eqref{dexpdt} and \eqref{I:gdot} we get:
\small
\begin{equation}
\mytagd \label{eq:deps_dq_start}
\exp\left({\widehat{\bm{\epsilon}}}\right)\widehat{\left(\mathrm{Ad}_{\exp\left({\widehat{\bm{\epsilon}}}\right)}^{-1}\bm{T}(\bm{\epsilon})\frac{\partial \bm{\epsilon}}{\partial q_p}\right)} = \bm{g}_{kBA}\widehat{\bm{J}}_{kBA,p}
\end{equation}
\normalsize
where $\bm{T}$ is the tangent operator provided in Appendix \ref{app::A}.

Rearranging the terms and considering all columns we get:

\small
\begin{equation}
\mytagd \label{eq:deps_dq_end}
\frac{\partial \bm{\epsilon}}{\partial \bm{q}} = \bm{T}^{-1}\left(\log(\bm{g}_{kBA})^\vee\right)(\mathrm{Ad}_{\bm{g}_{kBA}}\bm{J}_{kA}-\bm{J}_{kB})
\end{equation}
\normalsize

Using this we obtain the Jacobian of kinematic constraints.

\small
\begin{equation}
\mytagd \label{eq:de_dq}
\frac{\partial \bm{e}}{\partial \bm{q}}= \big\Vert_{k=1}^{n_{CL}}\bm{\Phi}_{\perp k}\bm{T}^{-1}\left(\log(\bm{g}_{kBA})^\vee\right)(\mathrm{Ad}_{\bm{g}_{kBA}}\bm{J}_{kA}-\bm{J}_{kB})
\end{equation}
\normalsize

\noindent
\rule{\columnwidth}{0.8pt}
\uline{Constraint Jacobian with respect to $\dot{\bm{q}}$}

The Jacobian of the equation \eqref{eq::constrainteq_app} with respect to $\dot{\bm{q}}$ is given by:
\small
\begin{equation}
\mytagd \label{eq:dconstraint_dqd_app}
  \frac{\partial \bm{C}}{\partial \dot{\bm{q}}}  = \frac{\partial \dot{\bm{A}}}{\partial \dot{\bm{q}}}\dot{\bm{q}}+\dot{\bm{A}} + \frac{2}{T_B}\bm{A}
\end{equation}
\normalsize
From the time derivative of the Pfaffian matrix, we can see that:
\small
\begin{equation}
\mytagd \label{eq:DAdqd}
\frac{\partial \dot{\bm{A}}}{\partial \dot{\bm{q}}}\dot{\bm{q}}+\dot{\bm{A}} = \bigg\Vert_{k=1}^{n_{CL}}\bm{\Phi}_{\perp k}\Biggl(\frac{\partial \bm{\varphi}_{kA}}{\partial \dot{\bm{q}}}-\frac{\partial \bm{\varphi}_{kB}}{\partial \dot{\bm{q}}} \Biggr)
\end{equation}
\normalsize
where, the partial derivative of $\bm{\varphi}_{k}$ is given by equation \eqref{eq::dCoriolis_phi_dqd_end}. We can rewrite it as:
\small
\begin{equation}
\frac{\partial \bm{\varphi}_{k}}{\partial \dot{\bm{q}}}= \bm{Y}_k^B-\mathrm{ad}_{\bm{\eta}_k}\bm{S}_k^B
\mytagd \label{eq::dCoriolis_dq_CL}
\end{equation}
\normalsize

\subsection*{Derivations in \ref{sec::Contact}}

The contact force in the global frame is given by:

\small
\begin{equation}
\mytagd \label{eq::contactforce_app}
 \bm{f}_c = k_c\delta_k^p\bm{u}_{\perp k} 
\end{equation}
\normalsize

The partial derivative of $\bm{f}_c$ with respect to $\bm{q}$ is given by

\small
\begin{equation}
\frac{\partial \bm{f}_c}{\partial \bm{q}}= k_c \bm{u}_{\perp k} p \delta_k^{p-1}\frac{\partial \delta}{\partial \bm{q}}+k_c \delta_k^{p} \frac{\partial \bm{u}_{\perp k}}{\partial \bm{q}}
\mytagd \label{eq::contactforce_start}
\end{equation}
\normalsize

From the definition of penetration \eqref{eq::penetration} and unit normal, we get

\small
\begin{equation}
\frac{\partial \delta}{\partial \bm{q}}=\bm{u}_{\perp k}^T \bm{C}_1 \frac{\partial \bm{r}_{g k}}{\partial \bm{q}}
\mytagd \label{eq::dpenetration_dq}
\end{equation}
\normalsize
and
\small
\begin{equation}
\mytagd \label{eq::dunitnormal_dq}
\begin{split}
\frac{\partial \bm{u}_{\perp k}}{\partial \bm{q}}=&\frac{1}{\|\bm{n}_{\perp k}\|}\bm{C}_1 \frac{\partial \bm{r}_{g k}}{\partial \bm{q}}-\bm{n}_{\perp k}\frac{1}{\| \bm{n}_{\perp k} \|^3}\bm{n}_{\perp k}^T\bm{C}_1 \frac{\partial \bm{r}_{g k}}{\partial \bm{q}}\\
=&\frac{1}{\|\bm{n}_{\perp k}\|}\left(\bm{I}_3-\bm{u}_{\perp k}\bm{u}_{\perp k}^T\right)\bm{C}_1 \frac{\partial \bm{r}_{g k}}{\partial \bm{q}}
\end{split}
\end{equation}
\normalsize

Linear velocity in the global frame is given by $\bm{v}_{gk} = \frac{\partial \bm{r}_{g k}}{\partial \bm{q}} \dot{\bm{q}} = [\bm{0} \; \bm{I}_3] \mathrm{Ad}_{\bm{g}_{\theta k}}\bm{S}_k^B \dot{\bm{q}}$. Using this we get,
\small
\begin{equation}
\frac{\partial \bm{r}_{g k}}{\partial \bm{q}}=[\bm{0} \; \bm{I}_3] \mathrm{Ad}_{\bm{g}_{\theta k}}\bm{S}_k^B
\mytagd \label{eq::dr_dq}
\end{equation}
\normalsize

Substituting \eqref{eq::dpenetration_dq}, \eqref{eq::dunitnormal_dq}, and \eqref{eq::dr_dq} in \eqref{eq::contactforce_start}  and using $\bm{F}_{gk} = [\bm{0}^{1 \times 3} \bm{f}_c^T]^T$ we get,

\small
\begin{equation}
\frac{\partial \bm{F}_{g k}}{\partial \bm{q}}=\begin{pmatrix} 
\bm{0} & \bm{0}\\ 
\bm{0} & \bm{k}^*\bm{C}_1
\end{pmatrix} \mathrm{Ad}_{\bm{g}_{\theta k}}\bm{S}_k^B
\mytagd \label{eq::contactforce_end}
\end{equation}
\normalsize
where, $\bm{k}^*=k_c p\delta_k^{p-1}\bm{u}_{\perp k}\bm{u}_{\perp k}^T+\frac{k_c\delta_k^{p}}{\|\bm{n}_{\perp k}\|}(\bm{I}_3-\bm{u}_{\perp k}\bm{u}_{\perp k}^T)$.

\subsection*{Derivations in \ref{sec::Hydrodynamic}}

The drag-lift force is given by:

\small
\begin{equation}
\mytagd  \label{eq::DragLift_app}
\bm{\mathcal{F}}_{Dk} = \bm{\mathcal{D}}_k  \|\bm{v}_k\|  \bm{\eta}_k
\end{equation}
\normalsize

The partial derivative of $\bm{\mathcal{F}}_{Dk}$ with respect to $\bm{q}$ is,

\small
\begin{equation}
\mytagd \label{eq::dDragLift_dq_start_app}
\frac{\partial \bm{\mathcal{F}}_{Dk}}{\partial \bm{q}}  = \bm{\mathcal{D}}_k  \bm{\eta}_k \frac{\partial \|\bm{v}_k\|}{\partial \bm{q}}+\bm{\mathcal{D}}_k\|\bm{v}_k\| \frac{\partial \bm{\eta}_k}{\partial \bm{q}}
\end{equation}
\normalsize

Using $\|\bm{v}_k\| = \sqrt{\bm{\eta}_k^T \bm{I}_v \bm{\eta}_k}$ we get,

\small
\begin{equation}
\mytagd \label{eq::dnormv_dq_app}
\frac{\partial \|\bm{v}_k\|}{\partial \bm{q}}  = \frac{1}{\|\bm{v}_k\|}\bm{\eta}_k^T\bm{I}_v \frac{\partial \bm{\eta}_k}{\partial \bm{q}}
\end{equation}
\normalsize

Substituting \eqref{eq::dnormv_dq_app} in \eqref{eq::dDragLift_dq_start_app} we get,

\small
\begin{equation}
\mytagd \label{eq::dDragLift_dq_end_app}
\frac{\partial \bm{\mathcal{F}}_{Dk}}{\partial \bm{q}}  = \bm{\mathcal{D}}_k^*  \frac{\partial \bm{\eta}_k}{\partial \bm{q}}
\end{equation}
\normalsize
where, $\bm{\mathcal{D}}_k^* = \bm{\mathcal{D}}_k\left(\frac{1}{\|\bm{v}_k\|}\bm{\eta}_k \bm{\eta}_k^T\bm{I}_{v} + \|\bm{v}_k\|  \bm{I}_6 \right)$.

It can be seen that the partial derivative of $\bm{\mathcal{F}}_{Dk}$ with respect to $\dot{\bm{q}}$ takes a similar form.

\small
\begin{equation}
\mytagd \label{eq::dDragLift_dqd_end_app}
\frac{\partial \bm{\mathcal{F}}_{Dk}}{\partial \dot{\bm{q}}}  = \bm{\mathcal{D}}_k^*  \frac{\partial \bm{\eta}_k}{\partial \dot{\bm{q}}}
\end{equation}
\normalsize

\bibliographystyle{agsm}
\bibliography{Manuscript}
\end{document}